\documentclass[10pt,journal,compsoc]{IEEEtran}
\usepackage{color}

\ifCLASSOPTIONcompsoc
\usepackage[nocompress]{cite}
\else
\usepackage{cite}
\fi

\usepackage[utf8]{inputenc}
\usepackage{longtable}
\usepackage{graphicx}
\usepackage{color}

\usepackage[cmex10]{amsmath}

\usepackage{array}

\usepackage{mdwmath}
\usepackage{mdwtab}
\usepackage{subfig}
\usepackage{url}

\usepackage{multirow}
\usepackage{booktabs}
\usepackage{amssymb}
\usepackage{lscape}

\usepackage{rotating}

\usepackage{rotating}

\usepackage{tikz}
\usepackage{pifont}

\newcommand{\image}{I}              		
\newcommand{\images}{\textbf{I}}     	
\newcommand{\nimages}{n}            		

\newcommand{\leftimage}{\image_{left}}              
\newcommand{\referenceimage}{\image_{ref}}     

\newcommand{\depthmap}{D}			
\newcommand{\estimateddepthmap}{\hat{\depthmap}}  
\newcommand{\depth}{d}				
\newcommand{\estimateddepth}{\hat\depth}				

\newcommand{\optimal}[1]{\tilde{#1}}

\newcommand{\disparitymap}{D}		
\newcommand{\estimateddisparitymap}{\hat{\disparitymap}}  
\newcommand{\disparity}{d}			
\newcommand{\estimateddisparity}{\hat{\disparity}}			

\newcommand{\costvolume}{C}              		

\newcommand{\pixel}{x}				

\newcommand{\rightpixel}{y}
\newcommand{\npixels}{N}

\newcommand{\viewpoint}{v}			

\newcommand{\recofunc}{f}               	
\newcommand{\encodingfunc}{h}      		
\newcommand{\decodingfunc}{g}      		
\newcommand{\confidence}{c} 		
  
\newcommand{\params}{\theta}			

\newcommand{\featurevector}{\textbf{x}}   
\newcommand{\latentspace}{\mathcal{X}}	
\newcommand{\featuredim}{c}			
\newcommand{\featuremap}{\textbf{f}}	

\newcommand{\objectivefunc}{\mathcal{L}}
\newcommand{\loss}{\objectivefunc}			
\newcommand{\weights}{W}				

\newcommand{\ie}{\emph{i.e., }}
\newcommand{\eg}{\emph{e.g., }}
\newcommand{\etal}{\emph{et al.}}
\newcommand{\noi}{\noindent}

\newcommand{\ltwo}{L_2}						
\newcommand{\lone}{L_1}						

\newcommand{\width}{W}						
\newcommand{\height}{H}						

\newcommand{\ndisparities}{n_d}				
\newcommand{\neighborhood}{\mathcal{N}}		

\newcommand{\distance}{d}

\newcommand{\energy}{E}			
\newcommand{\smoothnessenergy}{\energy_s}	

\newcommand{\argmin}{\arg\min}

\newcommand{\threshold}{\epsilon}

\newcommand{\xmark}{\small{\ding{53}}}%

\newcommand{\lossltwo}{\loss_{1}^{0}}
\newcommand{\lossabsdisparitydiff}{\loss_{1}^{1}}
\newcommand{\confidenceguidedloss}{\loss_{1}^{2} }
\newcommand{\lossweightedmeanabs}{\loss_{1}^{3}}
\newcommand{\lossapproximatesmoothlone}{\loss_{1}^{4}}
\newcommand{\losssmoothlone}{\loss_{1}^{5}}
\newcommand{\losscrossentropy}{\loss_{1}^{6}}
\newcommand{\losssubpixelcrossentropy}{\loss_{1}^{7}}
\newcommand{\losshinge}{\loss_{1}^{8} }
\newcommand{\inversewarploss}{\loss_{1}^{9}}		
\newcommand{\inversefeaturewarploss}{\loss_{1}^{10}}	
\newcommand{\lossgradient}{\loss_{1}^{11} }
\newcommand{\lossstructuraldisspatch}{\loss_{1}^{12}}
\newcommand{\lossltwomatching}{\loss_{1}^{13}}
\newcommand{\lossedgesemantic}{\loss_{1}^{14}}
\newcommand{\lossimprovedranking}{\loss_{1}^{15}}

\newcommand{\losslonegradient}{\loss_{2}^{1}}
\newcommand{\ltwosmoothnessgradient}{\loss_{2}^{2}}
\newcommand{\lossweightsecondgradient}{\loss_{2}^{3}}
\newcommand{\lossavgbhblone}{\loss_{2}^{4}}
\newcommand{\lossloopconsistency}{\loss_{2}^{5}}
\newcommand{\lossleftrightconsistency}{\loss_{2}^{6}}
\newcommand{\lossmdh}{\loss_{2}^{7}}
\newcommand{\lossscaleinvariantgradient}{\loss_{2}^{8}}

\graphicspath{{figures/}}
\DeclareGraphicsExtensions{.pdf}

\begin{document}
\bstctlcite{IEEEexample:BSTcontrol}


\title{A Survey on Deep Learning Architectures for Image-based Depth Reconstruction}

\author{Hamid Laga}
\date{September 2018}

\author{Hamid Laga  
	\IEEEcompsocitemizethanks{
		\IEEEcompsocthanksitem Hamid Laga  is with  Murdoch University (Australia), and with the Phenomics and Bioinformatics Research Centre, University of South Australia. E-mail: H.Laga@murdoch.edu.au
		}
	\thanks{Manuscript received April 19, 2005; revised December 27, 2012.}}

\markboth{A Survey on Depth Estimation Using Deep Learning}%
{Laga \MakeLowercase{\etalnospace}: A Survey on Depth Estimation Using Deep Learning}

\IEEEcompsoctitleabstractindextext{
	\begin{abstract}
		Estimating depth from RGB images  is a long-standing ill-posed problem, which has been explored for decades by the computer vision,  graphics, and machine learning communities.  In this article, we provide a comprehension survey of the recent developments in this field. We will  focus on the works which use deep learning techniques to estimate depth from one or multiple images.  Deep learning, coupled with the availability of large training datasets, have revolutionized the way the depth reconstruction problem is being approached  by the research community. In this article, we survey  more than $100$ key  contributions that appeared  in the past five years, summarize the most commonly used pipelines, and discuss their benefits and limitations. In retrospect of what has been achieved so far, we also conjecture what the future may hold for learning-based depth reconstruction research. 
			\end{abstract}
			
}

\maketitle


\IEEEdisplaynotcompsoctitleabstractindextext

%
\IEEEpeerreviewmaketitle

\section{Introduction}
\label{sec:introduction}

The goal of image-based 3D reconstruction is to infer the 3D  geometry and  structure of  real objects and scenes from one or multiple RGB images. This long standing ill-posed problem is fundamental to many applications such as robot navigation, object recognition and scene understanding, 3D modeling and animation,  industrial control, and medical diagnosis. 

Recovering the lost dimension from  2D images has been the goal of multiview stereo and  Structure-from-X methods, which have  been extensively investigated for many decades. The first generation of methods  has  focused on  understanding and formalizing the 3D to 2D projection process, with the aim to devise solutions to the ill-posed inverse problem. Effective solutions typically  require multiple images, captured using accurately calibrated cameras. Although these techniques can achieve remarkable results, they are still limited in many aspects. For instance, they are not suitable when dealing with occlusions, featureless regions, or highly textured regions with repetitive features. 




Interestingly, we, as humans, are good at solving such ill-posed inverse problems by leveraging prior knowledge. For example, we can easily infer the approximate size and rough geometry of objects using only one eye. We can even guess what it would look like from another view. We can do this because all the previously seen objects and scenes have enabled us to build a prior knowledge and develop mental models of what objects, and the 3D world in general, look like. The second generation of depth reconstruction  methods try to leverage this prior knowledge by formulating the problem as a recognition task. The avenue of deep learning techniques, and more importantly, the increasing availability of large training data sets, have lead to a new generation of methods that are able to recover the lost dimension even from a single image.  Despite being recent, these methods have demonstrated  exciting and promising results on various tasks related to computer vision and graphics.


In this article,  we provide a comprehensive and structured review of the recent advances in image-based depth reconstruction using deep learning  techniques. We have gathered more than $100$ papers, which appeared from $2014$  to December $2018$ in leading computer vision, computer graphics, and machine learning conferences and journals dealing specifically with this problem\footnote{This number is continuously increasing even at the time we are writing this article and during the review process.}.  The goal is to help the reader navigate in this emerging field, which gained a significant momentum in the past few years. Compared to the existing literature, the main contributions of this article are as follows;
\begin{itemize}
	\item To the best of our knowledge, this is the first survey 	paper in the literature which focuses on image-based depth reconstruction using deep learning techniques.
	
	\item We adequately cover the contemporary literature with respect to this area. We present a comprehensive review of  more than $100$ articles, which appeared from $2014$ to December $2018$.
	
	\item This article also provides a comprehensive review and an insightful analysis on all aspects of depth reconstruction using deep learning, including the training data, the choice of network architectures and their effect on the  reconstruction results, the training strategies, and the application scenarios. 
	
	\item We provide a comparative summary of the properties and performances of the reviewed methods for different scenarios including depth reconstruction from stereo pairs, depth reconstruction from  multiple images, and depth reconstruction from a single RGB image. 
	
\end{itemize} 

\noi The rest of this article is organized as follows; Section~\ref{sec:problemstatement} fomulates the problem and lays down the taxonomy. Section~\ref{sec:depth_by_stereo_matching}  focuses on the recent papers that use deep learning architectures for stereo matching. Section~\ref{sec:reco_depth_regression} reviews the methods that directly regress depth maps from one or multiple images without explicitly matching features across the input images.  Section~\ref{sec:training} focuses on the training procedures including the choice of training datasets and loss functions.   Section~\ref{sec:discussion_and_comparison} discuss the performance of some key methods. Finally, Sections~\ref{sec:future} and~\ref{sec:conclusion} discuss  potential future research directions, and  summarize the paper. 


\section{Scope and taxonomy}
\label{sec:problemstatement}

Let $\images = \{\image_k, k=1, \dots, \nimages \}$ be a set of $\nimages \ge 1$ RGB images of the same 3D scene, captured using cameras whose intrinsic and extrinsic parameters can be   \emph{known} or \emph{unknown}.   The images can be captured by multiple cameras placed around the 3D scene,  and thus they are spatially correlated, or with a single camera moving around the scene producing images that are  temporally correlated. The goal is to estimate one or multiple depth maps, which can be from the same viewpoint as the input~\cite{li2015depth,eigen2015predicting,garg2016unsupervised,liu2016learning},   or from a new arbitrary viewpoint~\cite{yang2015weakly,kulkarni2015deep,tatarchenko2016multi,zhou2016view,park2017transformation}. In this article focuses on methods that estimate depth from one or multiple images with known or unknown camera parameters. Structure-from-Motion (SfM) and Simultaneous Localization and Mapping (SLAM) techniques, which  estimate at the same time depth  and (relative) camera pose from  multiple images or  a video stream, are beyond the scope of this article and require a separate survey.


Learning-based depth reconstruction can be summarized as the process of learning a predictor $\recofunc_{\params}$ that can infer a depth map $\estimateddepthmap$ that is as close as possible to the unknown depth map $\depthmap$.  In other words, we seek to find a function $\recofunc_{\params}$ such that $ \objectivefunc(\images) = \distance\left(\recofunc_{\theta}(\images), \depthmap  \right)$ is minimized. Here, $\theta$ is a set of parameters, and  $\distance(\cdot, \cdot)$ is a certain measure of distance between the real depth map $\depthmap$ and the reconstructed depth map $\recofunc_\params(\images)$.  The reconstruction objective  $\objectivefunc$ is also known as the  \emph{loss function} in the deep learning jargon.  

We can distinguish two main categories of methods. Methods in the first class (Section~\ref{sec:depth_by_stereo_matching}) mimic the traditional stereo-matching techniques by explicitly learning how to match, or put in correspondence, pixels across the input images. Such correspondences can then be converted into an optical flow or a disparity map, which in turn can be converted into depth at each  pixel in the input image.   The predictor $\recofunc$ is composed of three modules: a feature extraction module, a feature matching and cost aggregation module, and a disparity/depth estimation module. 

The second class of methods (Section~\ref{sec:reco_depth_regression}) do not explicitly learn the matching function. Instead, they  learn a function that directly predicts depth (or disparity)  at each pixel in the input image(s).    These methods are very general and have been  used to estimate depth from a single image as well as  from multiple images taken from arbitrary view points. The predicted depth map $\depthmap$ can be from  the same viewpoint  as the input, or from a  new arbitrary viewpoint  $\viewpoint$.   We refer to these methods as \emph{regression-based} depth estimation.   


In all methods, the estimated depth maps can  be further refined using refinement modules~\cite{eigen2014depth,li2015depth,wang2015designing,eigen2015predicting} and/or  progressive reconstruction strategies where the reconstruction is  refined every time new images become available (Section~\ref{sec:disp_refinement}).  

The subsequent sections will review the state-of-the-art techniques. Within each class of methods, we will first review how the different modules within the common pipeline have been implemented using deep learning techniques. We will then discuss the different methods based on their input and output, the network architecture,  the training procedures including the loss functions they use and the degree of supervision they require, and   their performances on  standard benchmarks.



\section{Depth by stereo matching}
\label{sec:depth_by_stereo_matching}
Stereo-based depth reconstruction methods take $\nimages>1$ RGB images  and produce a depth map, a disparity map, or  an optical flow~\cite{dosovitskiy2015flownet,mayer2016large}, by matching features across the images.    The input images may be captured with calibrated~\cite{flynn2016deepstereo} or uncalibrated~\cite{ummenhofer2017demon} cameras. 

This section focuses on deep learning-based methods that mimic the traditional stereo-matching pipeline, \ie methods that learn  how to explicitly match patches across  stereo images for disparity/depth map estimation. We will first review how  individual blocks of the stereo-matching pipeline have been implemented using deep learning (Section~\ref{sec:stereo_pipeline}), and then discuss how these blocks are put together and trained for depth reconstruction (Section~\ref{sec:cnns_for_stereomatching}).

\subsection{The pipeline}
\label{sec:stereo_pipeline}

The stereo-based depth reconstruction process can be formulated as the problem of estimating a map $\depthmap$ ($\depthmap$ can be a depth/disparity map, or an optical flow) which minimizes an energy function of the form:
\begin{equation}
	\energy(\depthmap) = \sum_{\pixel} \costvolume(\pixel, \depth_\pixel) + \sum_\pixel\sum_{y\in \neighborhood_\pixel} \smoothnessenergy(\depth_\pixel, \depth_y).
	\label{eq:stereomatching_energy}
\end{equation}

\noi Here, $\pixel$ and $y$ are image pixels, $\depth_\pixel =  \depthmap(\pixel)$ is the depth / disparity at $\pixel$, $\costvolume$ is a 3D cost volume where $\costvolume(\pixel, \depth_\pixel)$ is the cost of pixel $x$ having depth or disparity equal to $\depth_\pixel$, $\neighborhood_\pixel$ is the set of pixels that are within the neighborhood of  $\pixel$, and $\smoothnessenergy$ is a regularization term, which is used to impose various constraints, \eg  smoothness and left-right  depth/disparity consistency, to the final solution. The first term of Equation~\eqref{eq:stereomatching_energy} is the matching cost. In the case of rectified stereo pairs, it measures the cost of matching the pixel $\pixel =(i, j)$ of the left image with the pixel $y=(i, j-\depth_x)$ of the right image. In the more general multiview stereo case, it measures the inverse likelihood of $\pixel$ on the reference image having depth $\depth_\pixel$.  

In general, this problem is solved with a pipeline of  four building blocks~\cite{scharstein2002taxonomy}:  (1) feature extraction, (2) matching cost calculation and aggregation,  (3) disparity/depth  calculation, and (4) disparity/depth refinement.  The first two blocks construct the cost volume $\costvolume$.  The third and fourth blocks define the regularization term and find the depth/disparity map $\tilde{D}$ that minimizes Equation~\eqref{eq:stereomatching_energy}. In this section, we review the recent methods that implement these individual blocks using deep learning techniques.

\begin{table*}[t]
	\caption{\label{tab:taxonomy_stereo_matching}Taxonomy of deep learning-based stereo matching algorithms. "Arch." refers to architecture. "CV" refers to cost volume. "corr." refers to correlation. }
	\resizebox{\linewidth}{!}{%
	
	\begin{tabular}{|c|c|c|c|c|c|c|c|c|c|c|}
		\noalign{\hrule height 1.3pt}
			\multirow{4}{*}{\textbf{Method}}&
			\multicolumn{2}{c|}{\textbf{(1) Feature extraction}}&
			\multicolumn{4}{c|}{\textbf{(2) Matching cost computation}}&
			\multicolumn{3}{c|}{\textbf{(3) Cost volume regularization}} &
			\multirow{4}{*}{\textbf{Depth estimation}}\\
		\cline{2-10}
			&
			\multirow{3}{*}{Scale} & \multirow{3}{*}{Arch.} & 
			\multirow{3}{*}{Hand-crafted} & \multicolumn{3}{c|}{Learned similarity} &
			\multirow{3}{*}{Input} & \multirow{3}{*}{Approach / Net arch.} & \multirow{3}{*}{Output} & 
			\\
			\cline{5-7}
			&
			& &
			& Feature  & \multicolumn{2}{c|}{ Similarity learning} & 
			& & & \\
			\cline{6-7}
			&
			& & 
			&aggregation & Network & Output & 
			& & &\\
		\cline{2-11}
		&
		Fixed vs.  & ConvNet vs.   & $\ltwo$ & Pooling,  & FC, CNN & Matching score,   & Cost volume (CV) & Standard stereo & Regularized CV  & argmin, argmax  \\
		& multiscale &ResNet  & correlation &  Concatenation &  & matching features  & CV+ features (CVF) &encoder & Disparity/depth  & soft argmin, soft argmax  \\
		& &  & &   &  &  & CVF + segmentation (CVFS) &   encoder + decoder &   & subpixel MAP  \\
		 &  &  &  &   &  &  & CVF + edge map (CVFE) &     &   & \ \\
		
		
		\noalign{\hrule height 1.1pt}
		
		MC-CNN Accr~\cite{zbontar2015computing,zbontar2016stereo}&  fixed & CNN  &  $-$ & concatenation&  4 FC  layers & matching score & cost volume & Standard stereo  &  Regularized CV & argmin \\ 
		\hline
		Luo \etal~\cite{luo2016efficient}&  fixed &  ConvNet &  correlation  &  $-$ &$-$& matching score & cost volume & Standard stereo  &   Regularized CV & argmin \\ 
		\hline
		Chen \etal~\cite{chen2015deep} & multiscale &  ConvNet & corr. + voting & $-$ & $-$ &matching score  & cost volume & Standard stereo   & Regularized CV &  argmin \\ 
		\hline
		L-ResMatch~\cite{shaked2017improved}& fixed     & ResNet  & $-$ &  concatenation & 4 (FC+ReLu) + FC &  matching score & cost volume& standard stereo +  & Regularized CV& argmin \\ 
									      &             &               &  &    &   &    &  &  4 Conv + 5 FC & & \\ 

		\hline
		Han \etal \cite{han2015matchnet} &  fixed&  ConvNet & $-$ & concatenation & 2 (FC + Relu), FC & matching score& $-$&$-$ &$-$ &  softmax\\ 
		\hline
		DispNetCorr \cite{mayer2016large} & fixed  &   ConvNet &  1D correlation& $-$ & $-$ & matching score & CV + features &  encoder + decoder& disparity & $-$ \\ 
		\hline  
		Pang \etal \cite{pang2017cascade} & fixed &   ConvNet  & 1D correlation & $-$ & $-$ & matching score  & CV + features & encoder + decoder & disparity& \\ 
		\hline  
		Yu \etal \cite{yu2018deep}& fixed & ResNet  &  $-$ & concatenation & encoder+decoder& matching scores & cost volume & 3D Conv & Regularized CV& softargmin \\ 
		\hline 
		Yang \etal~\cite{yang2018segstereo} (un)sup.&  fixed  &  ResNet & correlation & $-$ & $-$ & matching scores & CVF + segmentation & encoder-decoder & depth& \\ 
		\hline
		Liang \etal~\cite{liang2018learning} &  multiscale  & ConvNet & correlation &$-$ & $-$ & matching score & CV + features &encder-decoder &  depth& $-$ \\ 
		\hline
		Khamis \etal~\cite{khamis2018stereonet}&   fixed &  ResNet & $\ltwo$& $-$ & $-$& $-$  & cost volume & encoder & regularized volume & soft argmin/max \\ 
		\hline
		Chang \& Chen \cite{chang2018pyramid}  & multiscale & ResNet &$-$  &  concatenate & $-$&$-$ & cost volume & 12 conv layers, residual & regularized volume & regression \\ 
		  (basic)  &  &  & &   & & &  & blocks, upsampling &  & \\ 
		\hline
		Chang \& Chen \cite{chang2018pyramid}   & multiscale & ResNet & $-$&  concatenation & $-$ & $-$ & cost volume & stacked encoder-decorder  blocks, & regularized volume & regression \\ 
		 (stacked)  &  &  & &   & & &  & residual connections, upsampling &  & \\ 
		\hline
		
		Zhong \etal \cite{zhong2017self} &fixed &  ResNet &   $-$ & concatenation &  encoder-decoder &  matching scores &cost volume  & $-$ & regularized volume & soft argmin\\ 
		\hline
		 SGM-Net \cite{seki2017sgm}&   $-$ & $-$ &$-$ &$-$ &$-$ &$-$ & cost volume  & MRF + SGM-Net & regularized volume & $-$ \\ 
		\hline
		EdgeStereo \cite{song2018stereo}& fixed  &  VGG-16  & correlation & $-$ & &matching scores & CVF + edge map&  encoder-decoder (res. pyramid)& depth & $-$ \\ 
		\hline 
		Tulyakov \etal \cite{tulyakov2018practical}& fixed &   ConvNet  & $-$  &concatenation & encoder-decoder&matching signatures & matching signatures & encoder + decoder & regularized volume & subpixel MAP \\ 
		 &  &    && at each disparity  & & & &  & &  \\ 
		\hline
		Jie \etal \cite{jie2018left}&  $-$&  $-$& $-$ &$-$ &$-$ &$-$ & Left and right CVs & Recurrent ConvLSTM  & disparity  & $-$\\ 
							&  &  &  & & & &	  & with left-right consistency &  & \\ 

		\hline
		Zagoruyko \etal \cite{zagoruyko2015learning}&  multiscale&  ConvNet &  $-$ & concatenation &  FC & matching scores & $-$& $-$ &$-$ & $-$ \\ 
	
		\hline
		Hartmann \etal~\cite{hartmann2017learned} & fixed & ConvNet &  & Avg pooling & CNN  & matching score&$-$ & $-$& $-$ &  softmax\\
		
		\hline
		Huang \etal~\cite{huang2018deepmvs} & fixed & ConvNet & $-$  & Max pooling & CNN  & matching features & cost volume & encoder& regularized volume &  argmin \\
		
		\hline
		Yao \etal~\cite{yao2018mvsnet} & fixed & ConvNet  &  $-$ & Var. pooling &  $-$ & matching features & cost volume  &encoder-decoder &  regularized volume& softmax \\
		
		\hline
		Flynn \etal~\cite{flynn2016deepstereo}   & fixed &  2D Conv& $-$ & Conv across & CNN  &  matching score & cost volume &  encoder& regularized volume &  soft argmin/max\\ 
		&  &  &  & depth layers & & & & & & \\ 
		\hline
		Kar \etal~\cite{kar2017learning } & fixed & ConvNet  & $-$ & feature unprojection  + & CNN  &  matching score & cost volume &encoder-decoder & 3D occupancy  &  projection\\ 
		&  &  &  &  recurrent fusion & & & & &grid &  \\
		\hline
		Kendall \etal~\cite{kendall2017end} &fixed  & ConvNet & $-$ &  concatenation & CNN  &  matching features & cost volume & encoder-decoder& regularized volume&  soft argmin/max \\ 
%
%
		
		\noalign{\hrule height 1.3pt}
	\end{tabular}
	
	}	
\end{table*}

\subsubsection{Feature extraction}

\begin{figure}[t]
\centering{
	\includegraphics[width=0.45\textwidth]{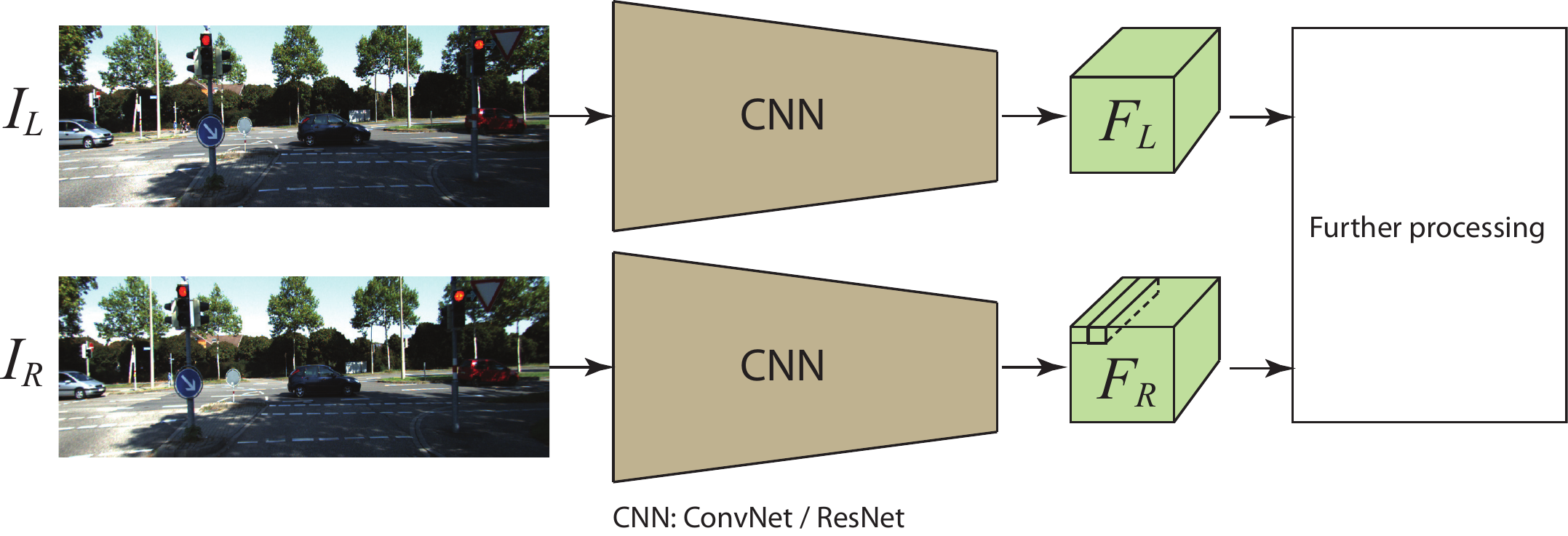}
	\caption{\label{fig:featureextraction} Feature extraction networks. }
}
\end{figure}

\begin{figure}[t]
\centering{
	\includegraphics[width=0.45\textwidth]{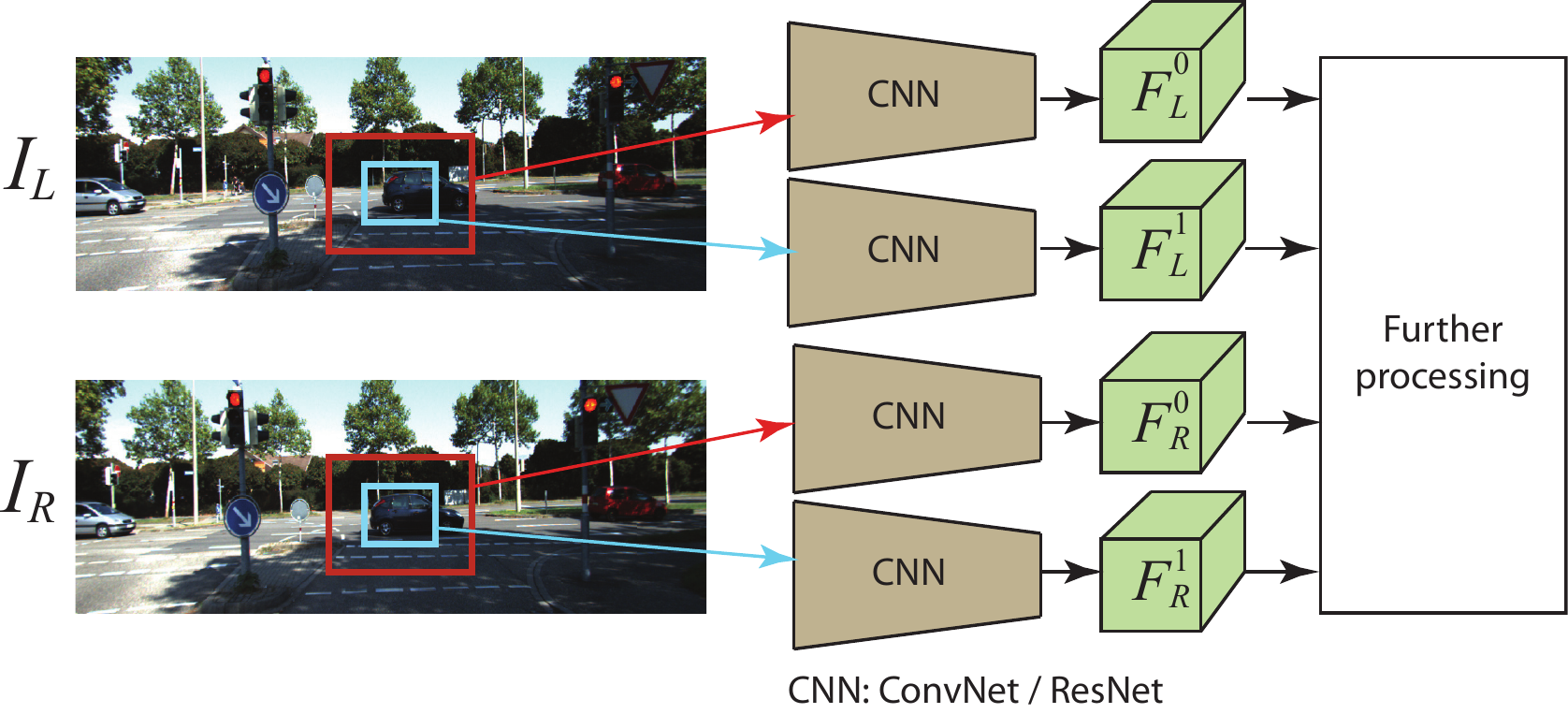}
	\caption{\label{fig:multiscalefeatureextraction} Multiscale feature extraction networks. }

}
\end{figure}

The first step  is to compute a good set of features to match across images.  This has been modelled using CNN architectures where the encoder takes  either patches around pixels of interests or entire images, and produces dense feature maps in the 2D image space. These features can be of fixed scale (Section~\ref{sec:fixedscale_features}) or multiscale (Section~\ref{sec:multiscalescale_features}).  

\vspace{6pt}
\paragraph{Fixed-scale features}  
\label{sec:fixedscale_features}

The main type of network architectures that have been used in the  literature is the multi-branch network with shared weights~\cite{zagoruyko2015learning,zbontar2015computing,dosovitskiy2015flownet,zbontar2016stereo,luo2016efficient,mayer2016large,kar2017learning,kendall2017end,pang2017cascade,liang2018learning}, see also Figure~\ref{fig:featureextraction}. It is composed of $\nimages \ge 2$  encoding branches, one for each input image, which act as descriptor computation modules.  Each branch is a Convolutional Neural Networks (CNN), which  takes a patch around a pixel $i$ and outputs  a feature vector that characterizes that patch~\cite{zagoruyko2015learning,zbontar2015computing,dosovitskiy2015flownet,simo2015discriminative,zbontar2016stereo,luo2016efficient,mayer2016large,flynn2016deepstereo,kar2017learning,kendall2017end,pang2017cascade,huang2018deepmvs,yao2018mvsnet,liang2018learning}.  It is generally composed of convolutional layers, spatial normalizations, pooling layers, and rectified linear units (ReLU). The  scale of the features that are extracted  is controlled by the size of the convolutional filters used in each layer as well as by the number of convolutional and pooling layers. Increasing the size of the filters and/or the number of layers increases the scale of the features that will be extracted. It has also the advantage of capturing more interactions between the image pixels. However, this comes with a high computation cost. To reduce the computational cost while increasing the field of view of the network, some techniques, \eg~\cite{Fu_2018_CVPR}, use dilated convolutions, \ie large convolutional filters but with holes and thus they are computationally efficient.


Instead of using fully convolutional networks, some techniques~\cite{yang2018segstereo} use residual networks, \eg ResNet~\cite{he2016deep}, \ie CNNs with residual blocks.  A residual block takes an input and estimates the residual that needs to be added to that input. They are used to ease the training of substantially deep networks since learning the residual of a signal is much easier than learning to predict the signal itself. Various types of residual blocks have been used in the literature. For example, Shaked and Wolf~\cite{shaked2017improved} proposed appending residual blocks with multilevel connections. Its particularity is that the network  learns by itself how to adjust the contribution of the added skip connections. 

Table~\ref{tab:feature_extractors} summarises the detailed architecture (number of layers, filter sizes, and stride at each layer) of various methods and the size of the features they produce.  Note that, one advantage of convolutional networks is that the convolutional operations within one level are independent from each other, and thus they are parallelizable. As such, all the features of an entire image can be computed with a single forward pass. 

\begin{table*}[t]
	\caption{\label{tab:feature_extractors} Network architectures for feature extraction.  Each layer of the network is described in the following format: (filter size, type, stride, output feature size, scaling). Scaling refers to upscaling or downscaling of the resolution of the output with respect to the input.  SPP refers to Spatial Pyramid Pooling. The last column refers to the feature size as produced by the last layer.  }
	
	\resizebox{\linewidth}{!}{%
	\begin{tabular}{@{}l l l l l l@{}}
	\toprule
	\textbf{Method} & \textbf{Input} & \textbf{Type} & \textbf{Architecture} & \textbf{Feature size} \\
	\midrule


	Dosovitskiy \etal~\cite{dosovitskiy2015flownet} & $512 \times 384$  & CNN & $(7\times 7, conv, 2, 64, -), (5\times 5, conv, 2, 128, -), (5\times 5, conv, 2, 256, -)$ &    $64\times 48 \times 256$\\
	
	\midrule
	Chen \etal~\cite{chen2015deep} & $13\times 13$ &  CNN &  $(3\times 3, conv, -, 32, -)_{1, 2}$, $(5\times 5, conv, -, 200, -)_{3, 4}$ & $1\times 200$ \\
	
	\midrule
	Zagoruyko~\cite{zagoruyko2015learning} &patches of varying sizes & CNN + SPP  &  $(-, conv + ReLu, -, -, -)_{1, 2}, (-, conv + SPP, -, -, -) $&   $1\times\featuredim$\\
	
	\midrule	
	Zbontar \& LeCun~\cite{zbontar2016stereo} (fast) &   patches $9 \times 9$ &     CNN & $(3\times 3, conv+ReLu, -, 64, -)_{1, 2, 3}, (3\times 3, conv, -, 64, -)$ &  $1\times 64$\\
	\midrule	
	Zbontar \& LeCun~\cite{zbontar2016stereo} (accr) & patches $9 \times 9$ & CNN & $(3\times 3, conv+ReLu, -, 112, -)_{1, 2, 3}, (3\times 3, conv, -, 112, -)$&     $1\times 112$\\
	
	
	\midrule	
	Luo \etal~\cite{luo2016efficient} & Small patch &  CNN &  $(3\times 3 \text{ or  } 5\times 5, conv+ReLu, -, 32 \text{ or } 64, -)_{1, 2, 3},  \text{or} $&    $1 \times 32$ or $1\times 64$\\
							& 	& & $(5\times 5, conv, -, 32 \text{ or } 64, -) $ & \\
	\midrule	
	DispNetC~\cite{mayer2016large}, Pang  \etal~\cite{pang2017cascade}& $768\times 364$& CNN &  $(7\times 7, conv, 2, 64, -), (5\times 5, conv, 2, 128, -)$  & $192 \times 96 \times 128$  \\
	
		\midrule	
	Kendall \etal~\cite{kendall2017end}& $\height \times \width$  & CNN (2D conv) +&  $(5\times 5, conv,  2, 32, /2), [(3\times 3, conv,  -, 32, -), (3\times 3, res,  -, 32, -)]_{1, \dots, 7}, $ &  $\frac{1}{2}\height\times \frac{1}{2}\width \times 32$\\
					&  &   Residual connections & $(3\times 3, conv,  -, 32, -)$, No RLu or BN on the last layer & \\
	
	
	\midrule	
	Liang  \etal~\cite{liang2018learning} & $\height \times \width$ & CNN & $(7\times 7, conv, 2, 64, /2), (5\times 5, conv, 2, 128, /2) $. & $\frac{1}{4}\height\times \frac{1}{4}\width \times 128$ \\
	
	\midrule	
	Kar \etal~\cite{kar2017learning}& $224\times 224$ & CNN& $(3\times 3, conv, -, 64, -), (3\times 3, conv, -, 64, -), (2\times 2, \text{maxpool}, -, 64, -)$   &  $32\times 32 \times 1024$ \\
		& & & $(3\times 3, conv, -, 128, -), (3\times 3, conv, -, 128, -), (2\times 2, \text{maxpool}, -, 128, -)$   & \\
		& & & $(3\times 3, conv, -, 512, -), (3\times 3, conv, -, 512, -), (2\times 2, \text{maxpool}, -, 512, -)$   & \\
		& & & $(3\times 3, conv, -, 1024, -), (3\times 3, conv, -, 1024, -)$   & \\
	
	\midrule 
	Yang \etal~\cite{yang2018segstereo} & $\height \times \width$ &  CNN + Residual blocks&  $ (3\times 3, conv, 2,  64, /2),  (3\times 3, conv, 1, 64, /1), (3\times 3, conv, 1, 128, /1)$   \\
								&				      & & $(3\times 3, maxpool, 2, 128, /2), (3\times 3, res\_block, 1, 256, /2)$,  & \\
								&				     &  &$(3\times 3, res\_block, 1, 256, /1)_{1, 2}$,  $(3\times 3, res\_blcok, 1,512, /2)$ & $-$ \\ 
	
	\midrule
	Shaked and Wolf~\cite{shaked2017improved} &  $11\times 11$ & CNN + & $conv_1, ReLU,$  Outer $\lambda-$ residual block, $ReLU, conv_{2\cdots5}, ReLU,$&  $1\times 1\times 112$ \\
									     & & Outer $\lambda-$residual blocks &  Outer $\lambda-$ residual block& \\ 
	\bottomrule
	\end{tabular}
	}
\end{table*}

\vspace{6pt}
\paragraph{Multiscale features}  
\label{sec:multiscalescale_features}

The methods described in Section~\ref{sec:fixedscale_features} can be extended to extract features at multiple scales, see Figure~\ref{fig:multiscalefeatureextraction}. This is done either by feeding the network with patches of different sizes centered at the same pixel~\cite{zagoruyko2015learning,chen2015deep,chang2018pyramid}, or  by using the features computed by the intermediate layers~\cite{liang2018learning}. Note that the deeper is a layer in the network, the larger is the scale of the features it computes.

Liang \etal~\cite{liang2018learning} compute multiscale features using a two-layer convolutional network.  The output of the two layers are then concatenated and fused,  using a convolutional layer, which results in \emph{multi-scale fusion features}.  Zagoruyko and Komodakis~\cite{zagoruyko2015learning} proposed a central-surround two-stream network which is essentially a network composed of two siamese networks combined at the output by a top network. The first siamese network, called central high-resolution stream, receives as input two $32 \times 32$ patches that are generated by cropping (at the original resolution) the central $32\times 32$ part of each  $64 \times 64$-input patch.  The second network, called surround low-resolution stream,  receives as input two $32 \times 32$ patches generated by downsampling at half the original input.  Chen \etal~\cite{chen2015deep} also used a similar approach but each network processes patches of size $13\times 13$.   The main advantage of this architecture is that it can compute the features at two different resolutions in a single forward pass. It, however,  requires one stream by scale, which  is not practical if more than two scales are needed. 

Chang and Chen~\cite{chang2018pyramid} used Spatial Pyramid Pooling (SPP) module to aggregate context in different scales and location. More precisely, the feature extraction module is composed of a CNN of seven layers, and an SPP module followed by convolutional layers. The CNN produces a feature map of size $\frac{1}{4} \height \times \frac{1}{4} \width \times 128$. The SPP module then takes a patch around each pixel but at four different sizes ($8\times8\times 128$, $16\times16\times 128$, $32\times32\times 128$, and $64\times64\times 128$), and converts them into one-channel by mean pooling followed by a $1\times 1$ convolution. These are then upsampled to the desired size and concatenated with features from different layers of the CNN, and further processed with additional convolutional layers to produce the features that will be fed to the subsequent modules for matching and disparity computation.  Chang and Chen~\cite{chang2018pyramid}  showed that the SPP module enables estimating disparity for inherently ill-posed regions.

In general, Spatial Pyramid Pooling (SPP) are convenient for  processing patches of arbitrary sizes. For instance, Zaogoruyko and Komodakis~\cite{zagoruyko2015learning} append an SPP layer at the end of the feature computation network.   Such a layer aggregates the features of the last convolutional layer through spatial pooling, where the size of the pooling regions dependents on the size of the input. By doing so, one will be able to feed the network with patches of arbitrary sizes and compute feature vectors of the same dimension. 

\subsubsection{Matching cost computation} 
\label{sec:cost_computation}

\begin{figure}[t]
\centering{
	\includegraphics[width=.48\textwidth]{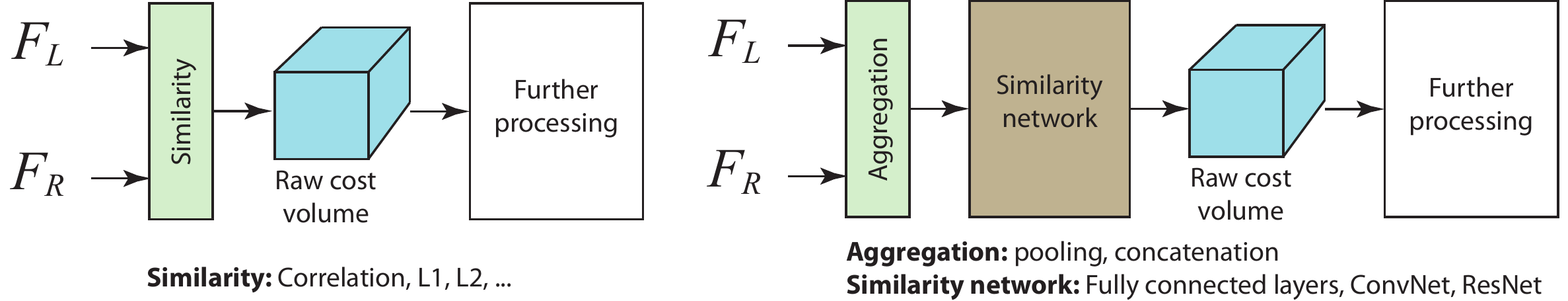} 
	\caption{\label{fig:similarity_computation} Taxonomy of similarity computation networks. }
}
\end{figure}

This module takes  the features computed on each of the input images,  and computes the matching scores of Equation~\eqref{eq:stereomatching_energy}.  The matching scores form  a 3D volume, called \emph{Disparity Space Image (DSI)}~\cite{scharstein2002taxonomy}, of the form $\costvolume(\pixel, \disparity_{\pixel})$ where $\pixel=(i, j)$  is the image coordinates of  pixel $\pixel$ and $\disparity_{\pixel}\in [0, \ndisparities]$ is the candidate disparity/depth value.  It is of size  $\tilde{\width} \times \tilde{\height} \times (\ndisparities+1)$, where   $\tilde{\width} \times \tilde{\height} $ is the resolution at which we want to compute the depth map and $\ndisparities$ is the number of depth/disparity values.   In stereo matching,  if the left and right images have been rectified so that the epipolar lines are horizontal then  $\costvolume(\pixel, \disparity_{\pixel})$ is the similarity between the pixel $\pixel = (i, j)$ on the rectified left image and the pixel $\rightpixel = (i, j - \disparity_{\pixel})$ on the rectified right image.   Otherwise, $\costvolume(\pixel, \disparity_{\pixel})$ indicates the likelihood, or probability, of the pixel $\pixel$ having depth $\disparity_{\pixel}$.

Similar to traditional stereo-matching methods~\cite{scharstein2002taxonomy},   the cost volume is computing by comparing the deep features of  the input images using standard metrics such as the $\ltwo$  distance, the cosine distance, and  the (normalized)  correlation distance (Section~\ref{sec:using_distance_measures}). With the avenue of deep neural networks, several new mechanisms have been proposed (Section~\ref{sec:similarity_networks}). Figure~\ref{fig:similarity_computation} shows the main similarity computation architectures. Below, we discuss them in details. 

\vspace{6pt}
\paragraph{Using distance measures}  
\label{sec:using_distance_measures}
The simplest way to form a cost volume is by taking the distance between the feature vector of a pixel and the feature vectors of the matching candidates, \ie the pixels on the other image that are within a pre-defined disparity range. There are several distance measures that can be used. Khamis \etal~\cite{khamis2018stereonet}, for example,  used the $\ltwo$ distance.  Other techniques, \eg \cite{dosovitskiy2015flownet,chen2015deep,zbontar2016stereo,luo2016efficient,mayer2016large,liang2018learning},  used correlation, \ie the inner product between feature vectors. The main advantage of correlation  over the $\ltwo$ distance is that it can be implemented using a layer of 2D~\cite{dosovitskiy2015flownet}  or 1D~\cite{mayer2016large} convolutional operations, called  correlation layer.  1D correlations are  computationally more efficient than their 2D counterpart. They, however, require  rectified images so that the search for correspondences is restricted to pixels within the same raw.  

Compared to the two other methods that will be described below, the main advantage of the correlation layer is that it does not require training since the filters are in fact the features computed by the second branch of the network.

\vspace{6pt}
\paragraph{Using similarity-learning networks} 
\label{sec:similarity_networks}

These methods aggregate the features produced by the different branches, and process them with a top network, which produces a matching score. The rational is to let the network learn from data the appropriate similarity measure. 

\vspace{6pt}
\noi \textbf{(1) Feature aggregation. } Some stereo reconstruction methods first aggregate the features computed by the different branches of the network before passing them through  further processing layers.   The aggregation can be done in two different ways:

\vspace{6pt}
\noi \textit{Aggregation by concatenation. }  The simplest way is to just concatenate  the  learned features computed by the different branches of the network and feed them  to the  similarity computation network~\cite{zagoruyko2015learning,zbontar2015computing,zbontar2016stereo,kendall2017end}.  Kendall \etal~\cite{kendall2017end} concatenate each feature with their corresponding feature from the opposite stereo image across each disparity level, and pack these into a 4D volume of dimensionality $\height \times \width \times (\ndisparities +1) \times \featuredim$ ($\featuredim$ here is the dimension of the features).  Huang \etal~\cite{huang2018deepmvs}, on the other hand, concatenate  the   $64^3$ feature volume, computed for the $64\times 64\times3$ reference image, and another volume of the same size  from the plane-sweep volume plane that corresponds to the $n-$th  input  image at the $d-$th disparity level, to form a  $64\times64\times 128$  volume.   Zhong \etal~\cite{zhong2017self} followed the same approach but  concatenate the features in an interleaved manner. That is, if $\featuremap_L$ is the feature map of the left image and $\featuremap_R$ the feature map of the right image then  the final feature volume is assembled in such a way that its $2i-$th slice holds the left feature map while the $(2i+1)-$th slice holds the right feature map but at disparity $\disparity = i$. That is,
    \begin{equation}
    	\featuremap_{LR}(u, v, \disparity) = \featuremap_L (u, v) \| \featuremap_R(u - \disparity, v),  
    \end{equation}
    where $\|$ denotes the vector concatenation. 
    
\vspace{6pt}
\noi \textit{Aggregation by pooling. } Another approach is to use pooling layers to aggregate the feature maps. For instance, Hartmann \etal~\cite{hartmann2017learned} used average pooling. Huang \etal~\cite{huang2018deepmvs} used max-pooling, while  Yao \etal~\cite{yao2018mvsnet} take their variance, which is equivalent to first computing the average feature vector and then taking the average distance  of the other features to the mean. 

%

The main advantage of pooling  over  concatenation is three-fold; First, it does not increase the dimensionality of the data that is fed to the top similarity computation network, which facilitates the training.  Second, it  makes it possible to input a varying number of views without retraining the network.  This is particularly suitable for multiview stereo (MVS) approaches, especially when dealing with an arbitrary number of input images and when the number of images at runtime may be different from the number of images at training. Finally,   pooling ensures that the results are invariant with respect to the order in which the images are fed to the network.

\vspace{6pt} 
\noi\textbf{(2) Similarity computation. } There are two types of networks that have been used in the literature: fully connected networks and convolutional networks.  

\vspace{6pt}
\noi\textit{Using fully-connected networks. }  In these methods,  the similarity computation network is composed of fully connected layers~\cite{zagoruyko2015learning,zbontar2015computing,zbontar2016stereo}. The last layer produces the  probability of  the input feature vectors being a good or a bad match. Zagoruyko and  Komodakis~\cite{zagoruyko2015learning}, for example,  used a  network composed of two fully connected layers (each with $512$ hidden units) that are separated by a ReLU activation layer.  Zbontar and LeCun~\cite{zbontar2015computing,zbontar2016stereo} used five fully connected layers with $300$ neurones each except for the last layer, which projects the output to two real numbers that are fed through a softmax function,  which in turn produces the  probability of the two input feature vectors being a good match.

\vspace{6pt} 
\noi\textit{Using convolutional networks.}  Another approach is to aggregate the features and further post-process them using convolutional networks, which output  either matching scores~\cite{flynn2016deepstereo,kar2017learning,hartmann2017learned} (similar to correlation layers),  or  matching features~\cite{kendall2017end, huang2018deepmvs}.   The most commonly used CNNs include max-pooling layers, which provide invariance in spatial transformation. Pooling layers also widen the receptive field area of a CNN without increasing the number of parameters. The drawback is that the network loses fine details. To overcome this limitation, Park and Lee~\cite{park2017look} introduced a pixel-wise pyramid pooling layer to enlarge the receptive field during the comparison of two input patches. This method produced more accurate matching cost than~\cite{zbontar2015computing}.

One limitation of correlation layers and convolutional networks that produce a single cost value is that they decimate the feature dimension. Thus, they  restrict the network to only learning relative representations between features, and cannot carry absolute feature representations.  Instead, a matching feature can be seen as a descriptor, or a feature vector, that characterizes the similarity between two given patches. The simplest way of computing matching features is by  aggregating the feature maps produced by the descriptor computation branches of the network~\cite{zhou2016learning,kendall2017end}, or by using an encoder that takes the concatenated features and produces another volume of matching features~\cite{huang2018deepmvs}.   For instance,  Huang \etal~\cite{huang2018deepmvs} take the $64\times64\times 128$  volume, formed by features, and process it using three convolutional layers to produce a $64\times64\times 4$ volume of matching features. Since the approach  computes $\ndisparities+1$ matching features, one for each disparity level ($\ndisparities=100$ in~\cite{huang2018deepmvs}),  these need to be aggregated into a single matching feature. This is  done using another encoder-decoder network with  skip connections~\cite{huang2018deepmvs}. Each level of the encoder is formed by a stride-2 convolution layer followed by an ordinary convolution layer. Each level of the decoder is formed by two convolution layers followed by a bilinear upsampling layer. It produces a volume of matching features of size $64\times64\times 800$.

\vspace{6pt} 
\noi\textbf{(3) Cost volume aggregation. } In general, multiview stereo methods, which take $\nimages$ input images,  compute $\nimages-1$ cost or feature matching volumes, one for each pair $(\image_0, \image_i)$, where $\image_0$ is the reference image. These need to be aggregated into a single cost/feature matching volume before feeding it into the disparity/depth calculation module. This has been done either by using (max, average) pooling or pooling followed by an encoder~\cite{huang2018deepmvs,yao2018mvsnet}, which produces the final cost/feature matching volume $\costvolume$.

\subsubsection{Disparity and depth computation}
\label{sec:depth_disparity_calculation}
We have seen so far the various deep learning techniques that have been used to estimate the cost volume $\costvolume$, \ie the first term of Equation~\eqref{eq:stereomatching_energy}. The goal now is to estimate the depth/disparity map $\optimal{\depthmap}$ that minimizes the energy function $\energy(\depthmap) $ of Equation~\eqref{eq:stereomatching_energy}.  This is  done in two steps; (1) cost volume regularization, and (2) disparity/depth estimation from the regularized cost volume. 

\begin{figure}[t]
\centering{
	\includegraphics[width=.48\textwidth]{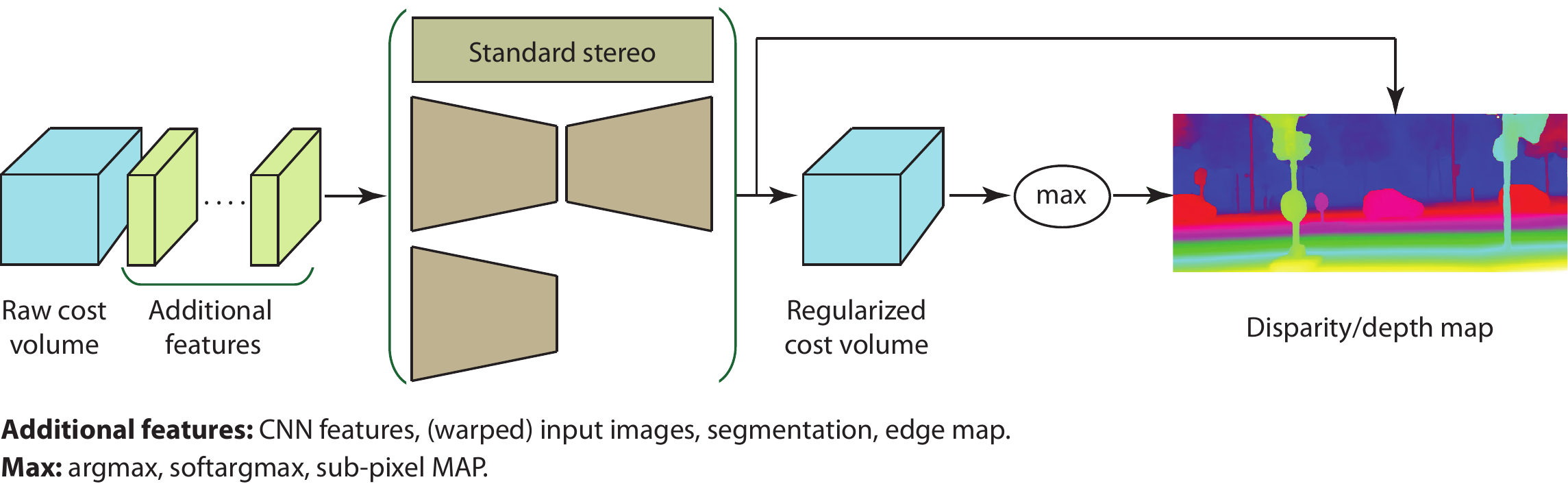} 
	\caption{\label{fig:regularization} Cost volume regularization and disparity/depth map estimation. }
}
\end{figure}

\vspace{6pt}
\paragraph{Cost volume regularization}

Once ta raw  cost volume is estimated, one can estimate disparity/depth by  dropping the smoothness term of Equation~\eqref{eq:stereomatching_energy} and taking the argmin, the softargmin, or the subpixel MAP approximation (see Section~\ref{sec:final_depth_estimation}).   In general, however, the raw cost volume computed from image features could be noise-contaminated (\eg due to the existence of non-Lambertian surfaces, object occlusions, and repetitive patterns). Thus, the estimated depth maps can be noisy.  

Several deep learning-based regularization techniques have been proposed to estimate accurate depth maps from the cost volume, see Figure~\ref{fig:regularization} for an illustration of the taxonomy.  Their input can be the cost volume~\cite{flynn2016deepstereo,khamis2018stereonet,kendall2017end,zhong2017self,yao2018mvsnet,kar2017learning,chang2018pyramid,huang2018deepmvs}, the cost volume concatenated with the features of the reference image~\cite{dosovitskiy2015flownet,liang2018learning} and/or with  semantic features such as the segmentation mask~\cite{yang2018segstereo} or the edge map~\cite{song2018stereo}. The produced volume is then processed with either an encoder-decoder network with skip connections~\cite{kendall2017end,zhong2017self,yao2018mvsnet,kar2017learning,chang2018pyramid,dosovitskiy2015flownet,liang2018learning,yang2018segstereo,song2018stereo}, or just an encoder~\cite{flynn2016deepstereo,khamis2018stereonet,huang2018deepmvs}, to produce either a regularized cost volume~\cite{kendall2017end,zhong2017self,yao2018mvsnet,kar2017learning,chang2018pyramid,flynn2016deepstereo,khamis2018stereonet,huang2018deepmvs}, or directly the disparity/depth map~\cite{dosovitskiy2015flownet,yang2018segstereo,song2018stereo,liang2018learning}. In the former case, the regularized volume is  processed using  argmin, softargmin, or subpixel MAP approximation (Section~\ref{sec:final_depth_estimation}) to produce the final disparity/depth map.


Note that some methods adopt an MRF-based stereo framework for cost volume regularization~\cite{chen2015deep,zbontar2015computing,luo2016efficient}. In these methods, the initial cost volume $\costvolume$  is fed to a  global~\cite{scharstein2002taxonomy} or a semi-global~\cite{hirschmuller2008stereo} matcher to compute the disparity map.  Semi-global methods define the smoothness term as
\begin{equation}
\small{
	\smoothnessenergy(\depth_\pixel, \depth_y) = \alpha_1 \delta(|\depth_\pixel - \depth_y| = 1) + \alpha_2 \delta(|\depth_\pixel - \depth_y| > 1),
	}
	\label{eq:semiglobal_matching}
\end{equation}

\noi where $\alpha_1$ and $\alpha_2$ are positive weights chosen such that $\alpha_2 > \alpha_1$.   Instead of manually setting these two parameters,  Seki \etal~\cite{seki2017sgm} proposed  SGM-Net, a neural network trained to provide these parameters at each image pixel.  They obtained better penalties than  hand-tuned methods as in~\cite{zbontar2015computing}.

\vspace{6pt}
\paragraph{Disparity/depth estimation} 
\label{sec:final_depth_estimation}

The simplest way to estimate  disparity/depth  from the (regularized) cost volume  $\costvolume$ is by using the pixel-wise  argmin, \ie  $\depth_{\pixel}  = \arg\min_{\depth} \costvolume(\pixel, \depth)$ (or equivalently $\arg\max$ if the volume $\costvolume$ encodes likelihood)~\cite{huang2018deepmvs}. However, the agrmin/argmax operatir is unable to produce sub-pixel accuracy and cannot be trained with back-propagation due to its non-differentiability.  Another approach is to process the cost volume using a layer of per-pixel softmin, also called soft argmin (or equivalently softmax), over disparity/depth~\cite{flynn2016deepstereo,kendall2017end,khamis2018stereonet}:
\begin{equation}
	\disparity^* =\frac{1}{ \sum_{j=0}^{\ndisparities}e^{-\costvolume(x, j)}}  \sum_{d=0}^{\ndisparities} \disparity\times  e^{-\costvolume(x, \disparity)}.    
\end{equation}






\noi It approximates sub-pixel MAP solution when the distribution is unimodal and symmetric~\cite{tulyakov2018practical}. When this assumption is not fulfilled, the softargmin blends the modes and may produce a solution that is far from all the modes. Also, the network only learns for the disparity range used during training. If the disparity range changes at runtime, then the network needs to be re-trained. To address these issues, Tulyakov \etal~\cite{tulyakov2018practical} introduced the sub-pixel MAP approximation that computes a weighted mean around the disparity with maximum posterior probability as:
\begin{equation}
	\disparity^* = \sum_{\disparity:  |\hat\disparity - \disparity | \le \delta}\disparity \cdot \sigma( \costvolume(x, \disparity) ),
\end{equation}

\noi where $\delta$ is a meta parameter set to $4$ in~\cite{tulyakov2018practical}, and $\displaystyle \hat\disparity = \arg\max_{\disparity} \costvolume(x, \disparity)$. Note that, in Tulyakov \etal~\cite{tulyakov2018practical}, the sub-pixel MAP is only used for inference.


\subsubsection{Refinement}
\label{sec:disp_refinement}

\begin{table*}[t]
	\caption{\label{tab:taxonomy_refinement}Taxonomy of disparity/depth refinement techniques. "reco error": reconstruction error. "CSPN": Convolutional Spatial Propagation Networks. }
	\resizebox{\linewidth}{!}{%
	
	\begin{tabular}{|>{\raggedright}p{3cm} |>{\raggedright}p{3cm}|>{\raggedright}p{3cm} |>{\raggedright}p{3cm}  |>{\raggedright}p{3cm}|>{\raggedright}p{3cm}|}
		\noalign{\hrule height 1.1pt}
			\textbf{Traditional methods}  &\multicolumn{5}{c|}{ \textbf{Deep learning-based methods}}  \\
			\cline{2-6}
				& \textbf{Input} &\multicolumn{3}{c|}{ \textbf{Approach}} & \textbf{Other cues}\\
			\cline{3-5}
			
			& &  \textbf{Bottom-up} &  \textbf{Top-down} &  \textbf{Guided} & \\
		\noalign{\hrule height 0.8pt}
			
			Variational~\cite{dosovitskiy2015flownet}   &  Raw depth & Split and merge~\cite{Lee_2018_CVPR} &  Decoder~\cite{eigen2014depth}&  Detect - Replace - Refine~\cite{gidaris2017detect} &  Joint depth and normal~\cite{Qi_2018_CVPR}\\ 
			 
			 Fully-connected CRF~\cite{huang2018deepmvs}  & Depth + Ref. Image~\cite{yao2018mvsnet}  & Sliding window~\cite{wang2015designing} &  Encoder + decoder~\cite{ummenhofer2017demon,khamis2018stereonet,zhang2018activestereonet}& Depth-balanced loss~\cite{Lee_2018_CVPR}  & Left-Right consistency~\cite{jie2018left}\\ 
			 
			Hierarchical CRF~\cite{li2015depth}         				& Depth + CV + Ref. Image + rec. error~\cite{liang2018learning}  & Diffusion using CSPN~\cite{liu2017learningaffinity} &   Encoder + decoder with residual learning \cite{pang2017cascade,yao2018mvsnet,Jeon_2018_ECCV,liang2018learning,zhang2018activestereonet} & & \\ 
			
			 Depth propagation~\cite{chen2015deep} 				& Depth + Rewarped right image~\cite{Zhang_2018_ECCV}        & Diffusion using recurrent convolutional operation~\cite{cheng2018learning}  & Progressive upsampling~\cite{Jeon_2018_ECCV} & & \\ 
			 CRF energy minimized with CNN~\cite{liu2016learning}   & Depth + Learned features~\cite{eigen2014depth}.   		   & & & & \\ 
			\hline


		\noalign{\hrule height 1.1pt}
		
		
	\end{tabular}
	}
\end{table*}

In general, the predicted disparity/depth maps are of low resolution,  miss fine details, and may suffer from over-smoothing especially at object boundaries.  Some methods also output incomplete and/or sparse maps.    Deep-learning networks that directly predict high resolution and high quality maps would require a large number of parameters and thus are usually difficult to train.  Instead, an additional refinement block is added to the pipeline. Its goal  is to (1) improve the resolution of the estimated disparity/depth map,  (2) refine the reconstruction of the fine details, and (3) perform depth/disparity completion.  Such refinement block can be implemented using  traditional approaches.  For instance, Dosovitskiy \etal~\cite{dosovitskiy2015flownet} use the variational approach from~\cite{brox2011large}. Huang \etal~\cite{huang2018deepmvs}   apply the  Fully-Connected Conditional Random Field (DenseCRF) of~\cite{krahenbuhl2011efficient} to the predicted raw disparities.   Li \etal~\cite{li2015depth} refine the predicted depth (or surface normals) from the super-pixel level to pixel level using a hierarchical Conditional Random Field (CRF).   The use of DenseCRF or hierarchical CRF encourages the pixels that are spatially close and with similar colors to have closer disparity predictions.   Also, this step removes unreliable matches via left-right check. Chen \etal~\cite{chen2015deep} compute the final disparity map from the raw one, after removing unreliable matches,  by propagating reliable disparities to non-reliable areas~\cite{sun2014real}.  Note that the use of CRF for depth estimation has been also explored by Liu \etal~\cite{liu2016learning}. However, unlike Li \etal~\cite{li2015depth},  Liu \etal~\cite{liu2016learning}  used a CNN to minimize the CRF energy.

In this section, we will look at how the refinement block has been implemented using deep learning, see Table~\ref{tab:taxonomy_refinement} for a taxonomy of these methods. In general, the input to the refinement module can be: (1) the estimated depth/disparity map, (2) the estimated depth/disparity map concatenated with the reference image scaled to the resolution of the estimated depth/disparity map~\cite{yao2018mvsnet},   (3)  the initially-estimated disparity map, the  cost volume, and the reconstruction error, which is  calculated as  the absolute difference between the multi-scale fusion features of the left image and the  multi-scale fusion features of the right image but back-warped using the initial disparity map to the left image~\cite{liang2018learning}, (4)  the raw disparity/depth map, and the right image but warped into the view of the left image using the estimated initial disparity map~\cite{Zhang_2018_ECCV}, and (5) the estimated depth/disparity map concatenated  with the  feature map of the reference image, \eg the output of a first convolutional layer~\cite{eigen2014depth}.    

Note that  refinement can be  hierarchical by cascading several refinement modules~\cite{pang2017cascade,zhang2018activestereonet}.



\vspace{6pt}
\paragraph{Bottom-up approaches}  A bottom-up network   operates in a sliding window-like approach. It takes small patches and estimates the refined depth at the center of the patch~\cite{wang2015designing}.  Lee \etal~\cite{Lee_2018_CVPR} follow a split-and-merge approach. The input image is split  into regions, and a depth is estimated for each region. The estimates are then merged using a fusion network, which operates in   the Fourier domain so that depth maps with different cropping ratios can be handled. The rational  is that inferring accurate depth at the desired resolution would require large networks with a large number of parameters to estimate. By making the network focus on small regions, fine details can be recovered with less parameters. However, obtaining the entire refined map will require multiple forward passes, which is not suitable for realtime applications.

Another bottom-up refinement  strategy is based on diffusion processes. The idea is to start with an incomplete depth map and use anisotropic diffusion to propagate the known depth to the regions where depth is missing.  Convolutional Spatial Propagation Networks (CSPN)~\cite{liu2017learningaffinity}, which implement an anisotropic diffusion process,  are particularly suitable this task. They take as input the original image and a sparse depth map, which can be the output of a depth estimation network, and predict, using a deep CNN, the diffusion tensor. This is then applied to the initial  map to obtain the refined one.  Cheng \etal~\cite{cheng2018learning} used this approach in their proposed refinement module. It   takes an initial depth estimate and performs linear propagation, in which the propagation is performed with a manner of recurrent convolutional operation, and the affinity among neighboring pixels is learned through a deep CNN. 


\vspace{6pt}
\paragraph{Top-down approaches} Another approach is to use a top-down network that processes the entire raw disparity/depth map. It can be implemented as (1) a decoder, which  consists  of unpooling units to extend the resolution of its input, as opposed to pooling, and convolution layers~\cite{eigen2014depth},  or with  (2)  a encoder-decoder network~\cite{ummenhofer2017demon}. In the latter case,  the encoder is to map the input into a latent space. The decoder then predicts the high resolution   map from the latent variable. These networks also use skip connections from the contracting part to the expanding part so that fine details can be preserved.  To avoid the   checkboard artifacts produced by the deconvolutions and  upconvolutions~\cite{odena2016deconvolution,khamis2018stereonet,zhang2018activestereonet},  several papers first upsample the initial map, \eg using bilinear upsampling, and then apply convolutions~\cite{khamis2018stereonet,zhang2018activestereonet}. 

These architectures can be used to directly predict the high resolution maps but also to predict the residuals~\cite{yao2018mvsnet,liang2018learning,zhang2018activestereonet}. As opposed to directly learning the refined disparity map,  residual learning provides a more effective refinement.  In this approach, the estimated  map and the resized reference image are concatenated  and used as a 4-channel input to a refinement network,  which  learns the disparity/depth residual. The  estimated residual is then added to the originally estimated  map to generate the refined map. 

Pang \etal~\cite{pang2017cascade} refine the raW disparity map using a  cascade of two CNNs. The first stage advances the  DispNet of~\cite{mayer2016large}  by adding extra up-convolution modules, leading to disparity images with more details. The second stage, initialized by the output of the first stage,  explicitly rectifies the disparity; it couples with the first-stage and generates residual signals across multiple scales. The summation of the outputs from the two stages gives the final disparity.

Jeon and Lee~\cite{Jeon_2018_ECCV} proposed a  deep Laplacian Pyramid Network to spatially varying noise and holes. By considering local and global contexts, the network progressively reduces the noise and fills the holes from coarse to fine scales. It first predicts, using residual learning,  a clean complete depth image at a coarse scale (quarter of the original resolution).  The prediction is then progressively upsampled through the pyramid to predict the half and original sized clean depth image. The network is trained with 3D supervision using  a loss that is a combination of a data loss and a structure-preserving loss. The data loss is a weighted sum of $\lone$ distance between the ground-truth depth and the estimated depth, the $\lone$ distance between the gradient of the ground-truth depth and the estimated depth, and the $\lone$ distance between the normal vectors of the estimated and ground-truth depths. The structure-preserving loss is gradient-based to preserve the original structures and discontinuities.  It is defined as the $\ltwo$ distance between the maximum gradient around a pixel in the ground-truth depth map and the maximum gradient around that pixel in the estimate depth map.



\vspace{6pt}
\paragraph{Guided refinement}

Gidaris and Komodakis~\cite{gidaris2017detect} argue that the approaches that  predict either new  depth estimates or residual corrections  are sub-optimal. Instead, they propose a generic CNN architecture that decomposes the refinement task into three steps: (1) detecting the incorrect initial estimates, (2) replacing the incorrect labels with new ones, and  (3) refining the renewed labels by predicting residual corrections with respect to them. Since the approach is generic, it can be used to refine the raw depth map produced by any other method, \eg\cite{luo2016efficient}.

In general, the predictions of the baseline backbone, which is composed of an encoder-decoder,   are coarse and smooth due to the lack of depth details. To overcome this, Zhang \etal~\cite{zhang2018deep} introduced a hierarchical guidance strategy, which guides the estimation process to predict fine-grained details. They perform this by attaching refinement networks (composed of 5 conv-residual blocks and several following $1\times 1$ convolution layers) to the last three layers of the encoder (one per layer). Its role is to predict predict depth maps at these levels. The features learned by these refinement networks are used as input to their corresponding layers on the decoder part of the backbone network. This is similar to using skip connection. However, instead of feeding directly the features of the encoder, these are further processed to predict depth map at that level. 	

Finally, to handle equally close and far depths, Li \etal~\cite{Lee_2018_CVPR} introduced depth-balanced Euclidean loss to reliably train the network on a wide range  of depths.

\vspace{6pt}
\paragraph{Leveraging other cues} 

Qi \etal~\cite{Qi_2018_CVPR} proposed a mechanism that uses the depth map to refine the quality of the normal estimates, and the normal map to refine the quality of the depth estimates. This is done using a two-stream CNN, one for estimating an initial depth map and another for estimating an initial normal map. Then,  it uses another two-stream networks: a depth-to-normal network and a normal-to-depth network. The former is used to refine the normal map using the initial depth map. The latter is used to refine the depth map using the estimated normal map.  				
			\begin{itemize}
				\item The depth-to-normal network first takes the initial depth map and generates a rough normal map using   PCA analysis. This is then fed into a 3-layer CNN, which estimates the residual. The residual is then added to the rough normal map, concatenated with the initial raw normal map, and further processed with one convolutional layer to output the refined normal map.
				
				\item The normal-to-depth network uses kernel regression process, which takes the initial normal and depth maps, and regresses the refined depth map.
			\end{itemize}

\noi Instead of estimating a single depth map from the reference  image, one can estimate multiple depth maps, one per input image, check the consistency of the estimates, and use the consistency maps to (recursively) refine  the  estimates. In the case of stereo matching, this process is referred to as the left-right consistency check, which traditionally was an isolated post-processing step and heavily hand-crafted. 

The standard approach for implementing the left-right consistency check is as follows;
\begin{itemize}
	\item Compute two disparity maps $\disparitymap_l$ and $\disparitymap_r$, one for the left image and another for the right image.
	\item Reproject the right disparity map onto the coordinates of the left image, obtaining  $\tilde\disparitymap_r$.
	\item Compute the error or confidence map indicating whether the estimated disparity  is correct or not.
	\item Finally, use the computed confidence map to refine the disparity estimate.
\end{itemize}

\noi A simple way  of computing the confidence map is by taking pixel-wise difference. Seki \etal~\cite{seki2016patch}, on the other hand,  used a CNN trained in a classifier manner. It outputs a label per pixel indicating whether the estimated disparity  is correct or not. This confidence map is then incorporated into a Semi-Global Matching (SGM) for dense disparity estimation. 

Jie \etal~\cite{jie2018left} perform left-right consistency check jointly with disparity estimation, using a  Left-Right Comparative Recurrent (LRCR) model. It   consists of two parallely stacked convolutional LSTM networks. The left network takes the cost volume and  generates a disparity map for the left image. Similarly, the right network generates, independently of the left network,  a disparity map for the right image.   The two maps are  converted to the opposite coordinates (using the known camera parameters) for comparison with each other. Such comparison produces two error maps, one for the left disparity and another for the right disparity.  Finally, the error map for each image is concatenated with its associated cost volume and used as input at the next step to the convolutional LSTM. This will allow the LRCR model to selectively focus on the left-right mismatched regions at the next step.

\subsection{Stereo matching networks}
\label{sec:cnns_for_stereomatching}\label{sec:architecture_stereo_matching}

In the previous section, we have discussed how the different blocks of the stereo matching pipeline have been implemented using deep learning. This section discusses how different state-of-the-art techniques used these blocks and put them together to solve the pairwise stereo matching-based depth reconstruction problem.   


\subsubsection{Early methods}  
Early methods, \eg \cite{zbontar2015computing,chen2015deep,zagoruyko2015learning,han2015matchnet,luo2016efficient,tulyakov2017weakly}, replace the hand-crafted features and similarity computation with deep learning architectures. The basic architecture is composed of a stack of the modules described in Section~\ref{sec:stereo_pipeline}.  The feature extraction module is implemented as a multi-branch network, with shared weights. Each branch computes features from its input. These are then matched using:
\begin{itemize}
	\item a fixed correlation layer (implemented as a convolutional layer)~\cite{zbontar2015computing,luo2016efficient}, 
	\item a fully connected neural network~\cite{han2015matchnet,zagoruyko2015learning,shaked2017improved,zbontar2016stereo}, which takes as input the concatenated features of the patches from the left and right images and produces a matching score. 
	\item convolutional networks composed of convolutional layers followed by ReLU~\cite{hartmann2017learned}.
\end{itemize}

\noi  Using convolutional and/or fully-connected layers  enables the network to learn from data the appropriate similarity measure, instead of imposing  one at the outset.  It is more accurate than using a correlation layer but is significantly slower. 

Note that while Zbontar \etal~\cite{zbontar2015computing,zbontar2016stereo} and Han \etal~\cite{han2015matchnet} use standard convolutional layers in the feature extraction block,   Shaked and Wolf~\cite{shaked2017improved} add residual blocks with multilevel weighted residual connections to facilitate the training of very deep networks.    It was demonstrated that this architecture outperformed  the base network of Zbontar \etal~\cite{zbontar2015computing}.   To enable multiscale features, Chen \etal~\cite{chen2015deep} replicate twice the feature extraction module and the correlation layer. The two instances take patches around the same pixel but of different sizes, and produce two matching scores. These are then merged using voting.  Chen \etal's approach shares some similarities with the central-surround two-stream network of~\cite{zagoruyko2015learning}. The main difference is that in~\cite{zagoruyko2015learning}, the output of the four branches of the descriptor computation module is given as input to a top decision network for fusion and similarity computation, instead of using voting.   Zagoruyko and Komodakis~\cite{zagoruyko2015learning} add at the end of each feature computation branch a Spatial Pyramid Pooling so that patches of arbitrary sizes can be compared. 

Using these approaches, inferring the raw cost volume  from a pair of stereo images is performed using a moving window-like approach, which would require multiple forward passes ($\ndisparities$ forward passes per pixel).   However, since correlations are highly parallelizable,  the number of forward passes can be significantly reduced. For instance,  Luo \etal~\cite{luo2016efficient} reduce the number of forward passes to one pass per pixel by using a siamese network where the first branch takes a patch around a pixel while the second branch takes a larger patch that expands over all possible disparities.    The output is  a single 64D representation for the left branch, and $\ndisparities \times 64$ for the right branch.  A correlation layer then computes a vector of length $\ndisparities$ where its $\depth-$th element is the cost of matching the pixel $\pixel$ on the left image with the pixel $\pixel - \depth$ on the rectified right image. Other papers, \eg \cite{zhong2017self,kendall2017end,shaked2017improved,hartmann2017learned,khamis2018stereonet,zhang2018activestereonet,yang2018segstereo}, compute the feature maps of the left and right images in a single forward pass. These, however, have a high memory footprint at runtime and thus the feature map is  usually computed at a resolution that is lower than the resolution of the input images. 

These early methods produce matching scores that can be aggregated into a cost volume, which corresponds to  the data term of Equation~\eqref{eq:stereomatching_energy}. They then extensively rely  on hand-engineered post-processing steps, which are not jointly trained with the feature computation and feature matching networks, to regularize the cost volume and refine the disparity/depth estimation~\cite{zbontar2016stereo,luo2016efficient,chen2015deep,seki2017sgm}.


\subsubsection{End-to-end methods}
Recent works solve the stereo matching problem using a pipeline that is trained end-to-end without post-processing. For instance, Kn{\"o}belreiter \etal~\cite{knobelreiter2017end} proposed  a hybrid CNN-CRF. The CNN part  computes the matching term of Equation~\eqref{eq:stereomatching_energy}. This then becomes the unary term of a Conditional Random Field (CRF) module, which performs the regularization. The pairwise term of the CRF is parameterized by edge weights and is computed using another CNN. Using the learned unary and pairwise costs, the CRF tries to find a joint solution optimizing the total sum of all unary and pairwise costs in a 4-connected graph.  The whole CNN-CRF  hybrid  pipeline, which  is trained end-to-end, could achieve a competitive performance using much fewer parameters (and thus a better utilization of the training data) than the earlier methods. 

%
Others papers \cite{dosovitskiy2015flownet,mayer2016large,zhong2017self,kendall2017end,shaked2017improved,hartmann2017learned,khamis2018stereonet,zhang2018activestereonet,yang2018segstereo} implement the entire pipeline  using convolutional networks.  In these approaches, the cost volume is computed in a single  forward pass, which results in a high memory footprint. To reduce the memory footprint, some methods such as~\cite{dosovitskiy2015flownet,mayer2016large} compute a lower resolution raw cost volume, \eg one half or one fourth of the size of the input images. Some methods, \eg~\cite{zhong2017self,kendall2017end,shaked2017improved,chang2018pyramid}, ommit the matching module. The  left-right features, concatenated across the disparity range, are directly fed to the regularization and depth computation module. This, however,  results in even larger memory footprint.  Tulyakov \etal~\cite{tulyakov2018practical} reduce the memory use, without sacrificing accuracy,  by introducing a matching module that compresses the concatenated features into compact matching signatures. The approach uses mean pooling instead of feature concatenation. This also reduces the memory footprint. More importantly, it allows the network to handle arbitrary number of multiview images, and to vary the number of input at runtime without re-training the network. Note that pooling layers have been  used to aggregate features of different scales~\cite{chang2018pyramid}.

The regularization module takes the cost volume, the concatenated features, or the cost volume concatenated with the reference image~\cite{dosovitskiy2015flownet}, with the features of the reference image~\cite{dosovitskiy2015flownet,liang2018learning}, and/or with semantic features such as the segmentation mask~\cite{yang2018segstereo} or the edge map~\cite{song2018stereo}, which serve as semantic priors. It then regularizes it and outputs either a depth/disparity map~\cite{dosovitskiy2015flownet,mayer2016large,pang2017cascade,song2018stereo,liang2018learning} or a distribution over depth/disparities~\cite{kendall2017end,zhong2017self,chang2018pyramid,jie2018left}. Both the segmentation mask~\cite{yang2018segstereo} and the edge map~\cite{song2018stereo} can be computed using deep networks that are trained jointly and end-to-end with the disparity/depth estimation networks. Appending  semantic features to the  cost volume  improves the reconstruction of fine details, especially near object boundaries. 

The regularization module is usually implemented as  convolution-deconvolution (hourglass) deep network with skip connections between the contracting and expanding parts~\cite{mayer2016large,dosovitskiy2015flownet,pang2017cascade,kendall2017end,zhong2017self,chang2018pyramid,liang2018learning},  or as a convolutional network~\cite{flynn2016deepstereo,khamis2018stereonet}.  It can use 2D convolutions~\cite{mayer2016large,dosovitskiy2015flownet,pang2017cascade,liang2018learning} or 3D convolutions~\cite{kendall2017end,zhong2017self,jie2018left,chang2018pyramid}. The latter has less parameters. In both cases, their disparity range is fixed in advance and cannot be re-adjusted without re-training. Tulyakov \etal~\cite{tulyakov2018practical} introduced the sub-pixel MAP approximation for inference, which computes a weighted mean around the disparity with  MAP probability. They showed that it is more robust to erroneous modes in the distribution and allows to modify the disparity range without re-training. 




Depth can be computed from the regularized cost volume using (1)  the softargmin operator~\cite{flynn2016deepstereo,kendall2017end,zhang2018activestereonet,khamis2018stereonet}, which is differentiable and allows sub-pixel accuracy but limited to network outputs that are unimodal,  or (2) sub-pixel MAP approximation~\cite{tulyakov2018practical}, which can handle multi-modal distributions.

Some papers, \eg\cite{kendall2017end}, directly regress high-resolution   map without an explicit refinement module. This is done by adding a final upconvolutinal layer to the regression module in order to upscale the cost volume to the resolution of the input images. In general, however,  inferring high resolution depth maps would require large networks, which are expensive in terms of memory storage but also hard to train given the large number of free parameters.   As such, some methods first estimate a low-resolution depth map and then refine it using a refinement module~\cite{pang2017cascade,liang2018learning,jie2018left}.  The refinement module as well as the early modules are trained jointly and end-to-end.  

\subsection{Multiview stereo (MVS) matching networks}
\label{sec:mvs_architectures}

\begin{figure}[t]
\centering{
	\begin{tabular}{@{}cc@{}}
		\includegraphics[width=0.25\textwidth]{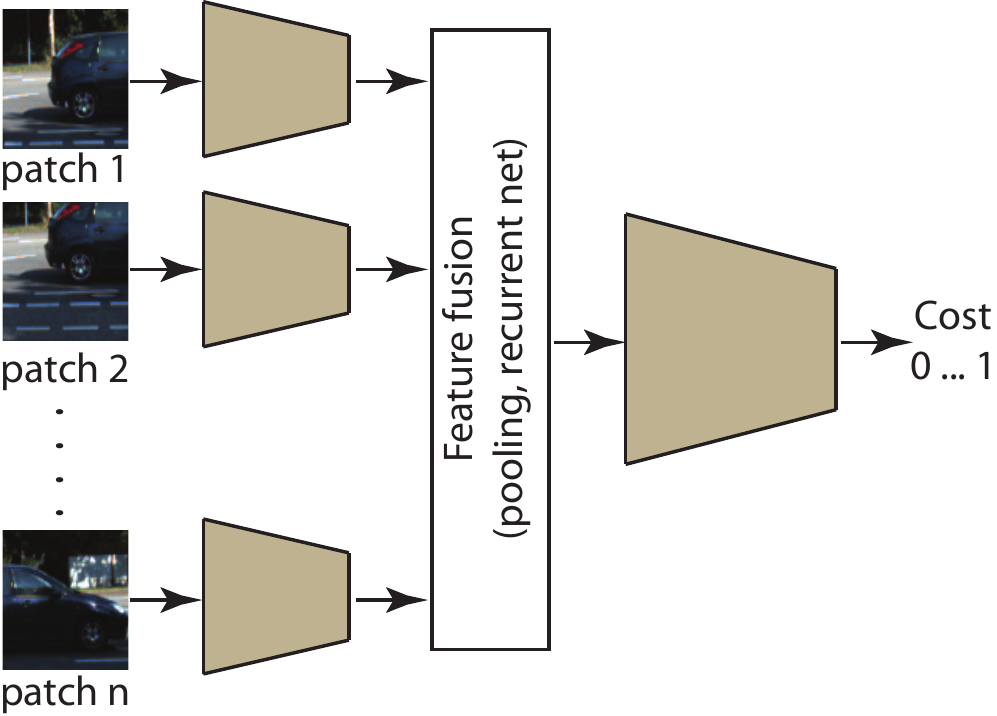} & \includegraphics[width=0.25\textwidth]{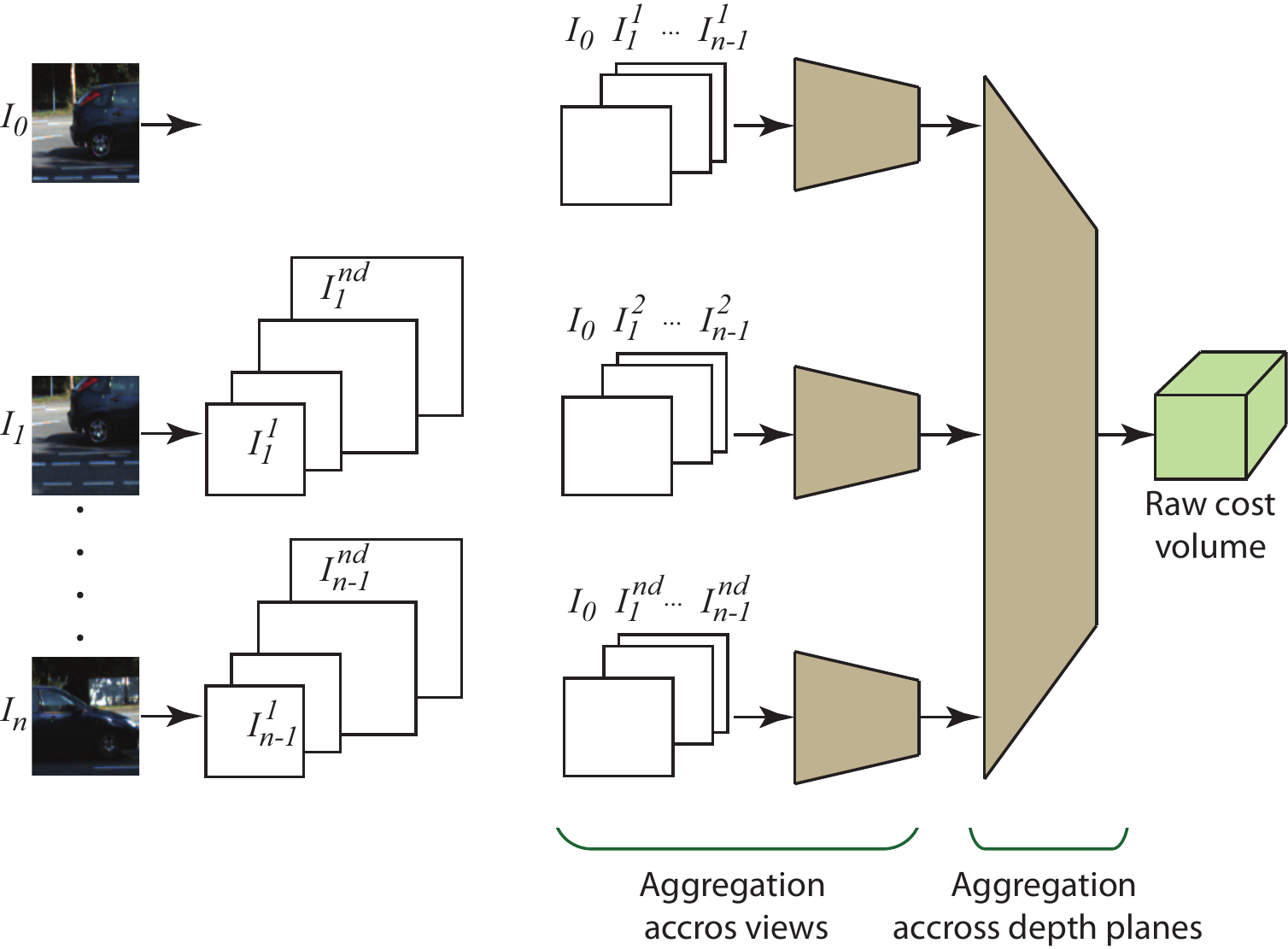} \\
		\small{(a) Hartmann \etal~\cite{hartmann2017learned}.} & \small{(b) Flynn \etal~\cite{flynn2016deepstereo}.} \\
		& 
	\end{tabular}
	\begin{tabular}{@{}c@{}}
		\includegraphics[width=0.5\textwidth]{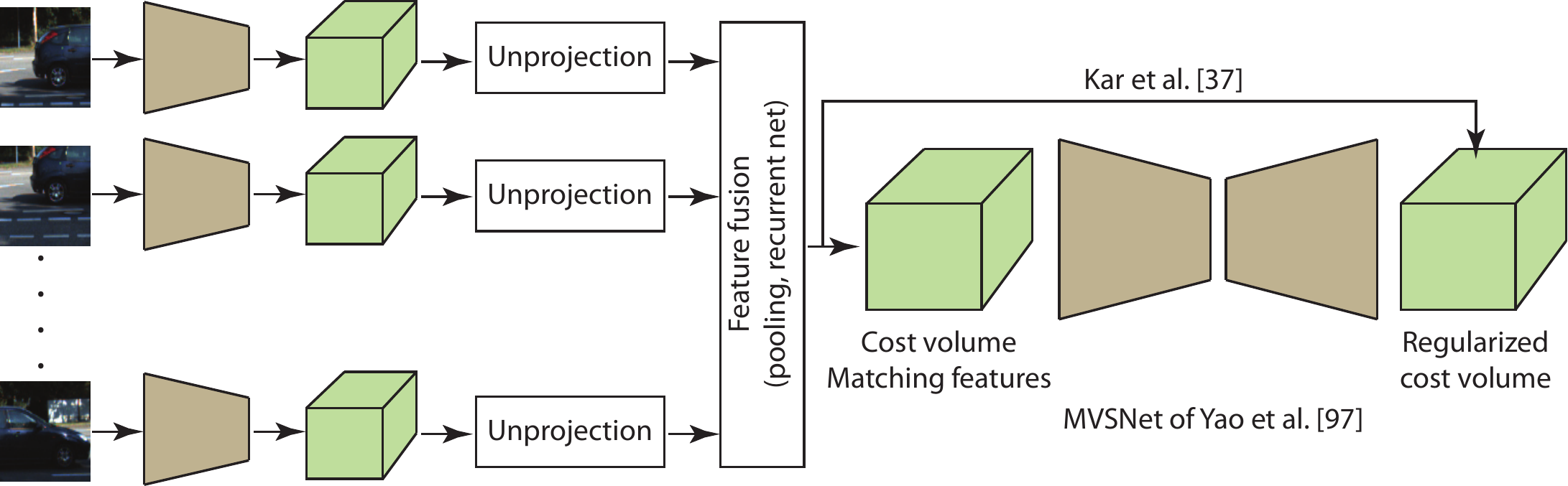}\\
		\small{(c) Kar \etal~\cite{kar2017learning} and Yao \etal~\cite{yao2018mvsnet}.}\\
		\\
		\includegraphics[width=0.5\textwidth]{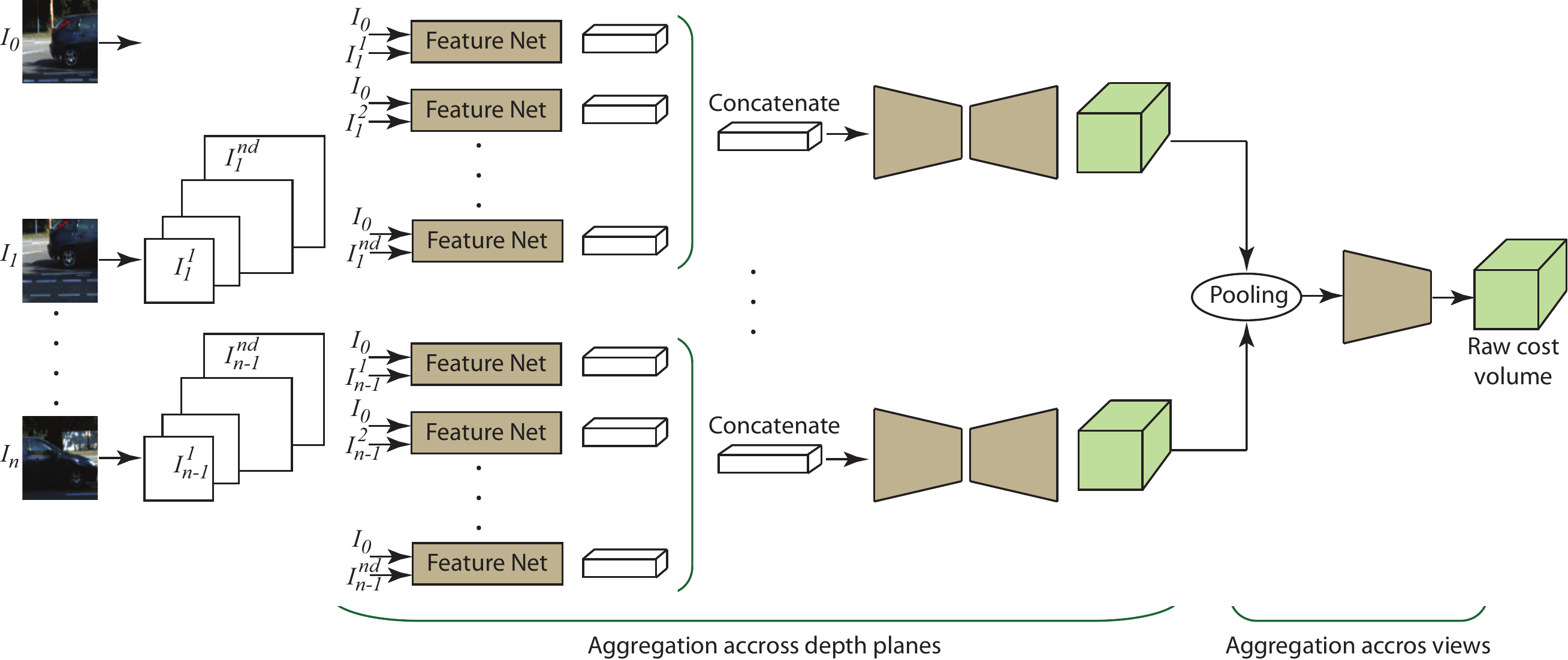}\\
		\small{(d) Huang \etal~\cite{huang2018deepmvs}.} 		
	\end{tabular}
	\caption{\label{fig:mvs} Taxonomy of multivewstereo methods. (a), (b), and (c) perform early fusion, while (d) performs  early fusion by aggregating features across depth plans, and late fusion by aggregating cost volumes across views. }

}
\end{figure}

The methods described in Section~\ref{sec:architecture_stereo_matching} have been  designed to reconstruct depth/disparity  maps from a pair of stereo images. These methods can be extended to the multiview stereo (MVS) case, \ie $\nimages > 2$, by   replicating the feature computation branch $\nimages$ times. The features computed by the different branches can then be aggregated using, for example,  pooling~\cite{hartmann2017learned,huang2018deepmvs,yao2018mvsnet} or a recurrent fusion unit~\cite{kar2017learning} before feeding the aggregated features into a top network, which regresses the depth map (Figures~\ref{fig:mvs}-(a), (b), and (c)). Alternatively, one can  sample pairs of views, estimate the cost volume from each pair, and then merge the cost volumes either by voting or pooling~\cite{huang2018deepmvs} (Figure~\ref{fig:mvs}-(d)). The former called \emph{early fusion} while the latter is called \emph{late fusion}.

The early work of Hartmann \etal~\cite{hartmann2017learned} introduced a mechanism to learn multi-patch  similarity, which replaces the correlation layer used in stereo matching.  The approach uses pooling to aggregate the features computed on the different patches before feeding them to the subsequent blocks of the standard stereo matching pipeline. Recent techniques use Plane-Sweep Volumes (PSV)~\cite{flynn2016deepstereo,huang2018deepmvs}, feature unprojection to the 3D space~\cite{kar2017learning,yao2018mvsnet}, and image unprojection to the 3D  space resulting in   the  Colored Voxel Cube (CVC)~\cite{ji2017surfacenet}. 

Flynn \etal~\cite{flynn2016deepstereo} and Huang \etal~\cite{huang2018deepmvs} use the camera parameters to unproject the input images into Plane-Sweep Volumes (PSV) and feed them into the subsequent feature extraction and feature matching networks. Flynn \etal~\cite{flynn2016deepstereo}'s network is composed of $\ndisparities$ branches, one for each depth plane (or depth value). The $\depth-$th  branch of the network takes as input the reference image and the planes of the Plane-Sweep Volumes of the other images and which are  located at  depth $\depth$. These are packed together and fed to a two-stages network. The first stage, which  consists of 2D convolutional rectified linear layers that share weights across all depth planes, computes matching features between the reference image and the PSV planes located at depth $\depth$. The second stage is composed of  convolutional layers that  are connected across depth planes in order to model  interactions between them.  The final layer of the network is a per-pixel softmax over depth, which returns the most probable depth value per pixel.  The approach, which  has been used for pairwise and multiview stereo matching, requires that the number of views and the camera parameters of each view to be known. It also requires setting in advance the disparity range.

Huang \etal~\cite{huang2018deepmvs}'s approach, which also operates on the plane-sweep volumes, uses a network composed of three parts: the patch matching part, the intra-volume feature aggregation part, and the inter-volume feature aggregation part:
\begin{itemize}
	\item The patch matching part is a siamese network. Its first branch extracts features from a patch in the reference image and the second one from the plane-sweep volume that corresponds to the $i-$th  input  image at the $\depth-$th disparity level. ($100$ disparity values have been used.)  The features are then concatenated and passed to the subsequent convolutional layers.   This process is  repeated for  all the plane-swept images. 
	
	\item The output from the patch matching module of the $\depth-$th  plane-sweep volume are concatenated and fed into another encoder-decoder  which produces a feature vector $F_\depth$ of size $64\times 64\times 800$. 
	
	\item	All the feature vectors $F_i, i=1, \dots, \nimages$ (one for each input image), are aggregated using a max-pooling layer followed by convolutional layers, which produce a depth map of size $64\times 64$.
\end{itemize}

\noi Unlike Flynn \etal~\cite{flynn2016deepstereo},  Huang \etal~\cite{huang2018deepmvs}'s approach does not require a fixed number of input views since  aggregation is performed using pooling.  In fact, the number of views at runtime can be different from the number of views used during training.  

The main advantage of using PSVs is that they eliminate the need to supply rectified images.  In other words, the camera parameters are implicitly encoded.  However, in order to compute the PSVs, the intrinsic and extrinsic camera parameters need to be either provided in advance or estimated using, for example, Structure-from-Motion techniques as in~\cite{huang2018deepmvs}.  Also, these methods require setting in advance the disparity range and its discretisation. 

Instead of using PSVs,  other methods use the camera parameters to unproject either the input images~\cite{ji2017surfacenet} or  the learned features~\cite{kar2017learning,yao2018mvsnet} into either a regular 3D feature grid,  by rasterizing the viewing rays with the known camera poses,  a 3D frustum of a reference camera~\cite{yao2018mvsnet}, or by warping  of  the features into different  parallel frontal planes of the reference camera, each one located at a specific depth.  This unprojection aligns the features along epipolar lines, enabling efficient local matching by using  either  some distance measures such as  the Euclidean  or cosine distances~\cite{kar2017learning},  using a recurrent network~\cite{kar2017learning},  or using an encoder composed of multiple convolutional layers producing the probability of each voxel being on the surface of the 3D shape.

Note that the approaches of Kar \etal~\cite{kar2017learning}  and Ji  \etal~\cite{ji2017surfacenet} perform volumetric reconstruction and use 3D convolutions. Thus, due to the memory requirements, only a coarse volume of size $32^3$ could be estimated. Huang \etal~\cite{huang2018deepmvs} overcome this limitation by directly regressing depth from different reference images.   Similarly, Yao \etal~\cite{yao2018mvsnet} focus on producing the depth map for one reference image at each time. Thus, it can directly reconstruct a large scene.

Table~\ref{tab:performance_mvs_ktti15} summarizes the performance of these techniques. Note that most of them do not achieve sub-pixel accuracy, require the depth range to be specified in advance and cannot vary it at runtime without re-adjusting the network architecture and retraining it. Also, these methods fail in reconstructing tiny features such as those present in vegetation. 
 


\section{Depth estimation by regression}
\label{sec:reco_depth_regression}

Instead of trying to match features across images, methods in this class directly regress disparity/depth from the input images  or their learned features~\cite{dosovitskiy2015flownet,mayer2016large,ummenhofer2017demon,godard2017unsupervised}.    These methods  have no direct notion of descriptor matching.  They consider a learned, view-based representation for  depth reconstruction from either $n$ predefined viewpoints $\{\viewpoint_1, \dots, \viewpoint_n\}$, or from any arbitrary viewpoint specified by the user.  Their goal is to learn a predictor $\recofunc$  (see Section~\ref{sec:problemstatement}), which predicts depth map from an input $\images$.

\subsection{Network architectures}

We classify the state-of-the-art into two  classes, based on the type of network architectures they use. In the first class of methods, the predictor $\recofunc$ is an encoder which directly regresses the depth map~\cite{garg2016unsupervised}.  

In the second class of methods, the predictor $\recofunc$ is composed of an encoder and a top network. The  encoder,  which learns, using a convolutional network,  a function $\encodingfunc$  that  maps the input  $\images$  into a compact latent representation $\featurevector = \encodingfunc(\images) \in \latentspace$. The space  $\latentspace$ is referred to as \emph{the latent space}. The encoder  can be designed following any of the architectures discussed in Section~\ref{sec:stereo_pipeline}.  The top network $\decodingfunc$ takes the compact representation, and eventually the target viewpoint $\viewpoint$,    and generates the estimated depth map  $\hat{\depthmap} = \decodingfunc\left( \encodingfunc(\images), \viewpoint\right) =  (\decodingfunc \circ \encodingfunc) (\images, \viewpoint)$.   Some methods use a top network composed of fully connected layers~\cite{eigen2014depth,li2015depth,wang2015designing,liu2016learning}. Others  use  a decoder composed of upconvolutional layers~\cite{eigen2015predicting,mayer2016large,zhou2016learning,laina2016deeper,godard2017unsupervised}. 

%


The advantage  of  fully-connected layers is that they aggregate  information coming from the entire image, and thus enable the network to infer depth at each pixel using global  information.  Convolutional operations, on the other hand, can only see  local regions. To capture larger spatial relations, one needs to  increase the number of convolution layers or use dilated convolutions, \ie large convolutional filters but with holes


\subsubsection{Input encoding networks}
In general, the encoder stage is composed of convolutional layers, which capture local interactions between image features, followed by a number of fully-connected layers, which capture global interactions. Some  layers are followed  by spatial pooling operations to reduce the resolution of the output. For instance, Eigen \etal~\cite{eigen2014depth}, one of the early works that tried to regress depth directly from a single input image, used an encoder composed of five feature extraction layers of convolution and max-pooling followed by one fully-connected layer. This maps the input image of size $304\times228$ or $576\times 172$, depending on the dataset, to a latent representation of dimension $1 \times 4096$. Liu \etal~\cite{liu2016learning}, on the other hand, used 7 convolutional layers to map the input into a low resolution feature map of dimension $512$.

Garg \etal~\cite{garg2016unsupervised}   used  this architecture to directly regress depth map from  an input RGB images of size $188\times 620$.  The encoder is composed of  $7$ convolutional layers. The second, third, and fifth layers are followed by pooling layers to reduce the size of the output and thus the number of parameters of the network. The output of the sixth layer is a feature vector, which can be seen as a latent representation of size $16\times 16 \times 2048$. The last convolutional layer  maps this latent representation into a depth map of size $17\times 17$, which is upsampled using two fully connected layers and three upconvolutional layers, into a depth map of size $176\times 608$. Since it  only relies on convolutional operation to regress depth, the approach does not capture global interactions. 

Li \etal~\cite{li2015depth} extended the approach of Eigen \etal~\cite{eigen2014depth}  to  operate on superpixels and at multiple scales. Given an image, super-pixels are obtained and multi-scale image patches  (at five different sizes) are extracted around the super-pixel centers.   All patches of a super-pixel are  resized to $227\times 227$ pixels to form a multiscale input to a pre-trained multi-branch deep network (AlexNet or VGGNet).  Each branch generates a latent representation of size $1\times 4096$. The latent representations from the different branches are concatenated together and fed to the top network. Since the networks process patches, obtaining the entire depth map requires multiple forward passes.

Since its introduction, this approach has been extended in many ways. For instance, Eigen and Fergus~\cite{eigen2015predicting} showed a substantial improvement by switching from AlexNet (used in~\cite{eigen2014depth}) to VGG, which has a higher disciminative power. Also, instead of using fully convolutional layers, Laina \etal~\cite{laina2016deeper}   incorporate residual blocks to ease the training.  The encoder is implemented following the same architecture as ResNet50 but without the fully connected layers. 

Using repeated spatial pooling  reduces the spatial resolution of the feature maps. Although high-resolution maps can be obtained using the refinement techniques of Section ~\ref{sec:disp_refinement}, this would require additional computational and memory costs. To overcome this problem, Fu \etal~\cite{Fu_2018_CVPR},   removed some pooling layers and replaced some convolutions with dilated convolutions. In fact, convolutional operations are local, and thus, they do not capture the global structure. To enlarge their receptive field, one can increase the size of the filters, or increase the number of convolutional and pooling layers. This, however, would require additional computational and memory costs, and will complicate the network architecture and the training procedure. One way to solve this problem is by using dilated convolutions, \ie convolutions with filters that have holes~\cite{Fu_2018_CVPR}.  This allows  to enlarge the receptive field of the filters without decreasing the spatial resolution or increasing the number of parameters and computation time.  

Using this principle, Fu \etal~\cite{Fu_2018_CVPR} proposed an encoding module that operates in two stages. The first stage extracts a dense feature map using an encoder whose  last few downsampling operators (pooling, strides) are replaced with dilated convolutions  in order to enlarge the receptive field of the filters. The second stage processes the dense feature map using three parallel modules;  a full image encoder, a cross channel leaner, and an atrous spatial pyramid pooling (ASPP).  The  full image encoder maps the dense feature map into a latent representation. It uses an average pooling layer with a small kernel size and stride to reduce the spatial dimension. It is then followed by a fully connected layer to obtain a feature vector, then add a convolutional layer with $1\times1$ kernel and copy the resultant feature vector into a feature map where each entry has the same feature vector. The ASPP module extracts features from multiple large receptive fields via dilated convolutions, with three different dilation rates. The output of the three modules are concatenated to form the latent representation.

These encoding techniques extract absolute features, ignoring the depth constraints of neighboring pixels, \ie relative features. To overcome this limitation, Gan \etal~\cite{Gan_2018_ECCV} explicitly model the relationships of different image locations using   an affinity layer. They also combine absolute and relative features in an end-to-end network.   In this approach, the input image is first processed by a ResNet50 encoder. The produced absolute feature map is fed into a context network, which captures  both neighboring and global context information.  It is composed of an affinity layer, which computes correlations between the features of neighboring pixels,  followed by a fully-connected layer, which combines absolute and relative features. The output is  fed into a depth estimator, which produces a coarse depth map.

\subsubsection{Decoding networks}


Many techniques first compute a latent representation of the input and then use a top network to decode the latent representation into a coarse depth map.  In general, the decoding process can be done with either a series of fully-connected layers~\cite{eigen2014depth,li2015depth,wang2015designing,liu2016learning}, or upconvolutional layers~\cite{eigen2015predicting,mayer2016large,zhou2016learning,laina2016deeper,godard2017unsupervised}.

\vspace{6pt}
\paragraph{Using fully-connected layers}

Eigen \etal~\cite{eigen2014depth},  Li \etal~\cite{li2015depth},   Eigen and Fergus~\cite{eigen2015predicting} and Liu \etal~\cite{liu2016learning} use a top network composed of two fully-connected layers. The main advantage of using fully connected layers is that their receptive field is global. As such, they aggregate information from the entire image in the process of estimating the depth map. By doing so, however, the number of parameters in the network is high, and subsequently the memory requirement and computation time increase substantially.  As such, these methods only estimate low-resolution coarse depth maps, which are then refined using some refinement blocks. For example, Eigen and Fergus~\cite{eigen2015predicting} use  two refinement blocks similar to those used in~\cite{eigen2014depth}. The first one produces predictions at a mid-level resolution, while the last one produces high resolution depth maps, at half the resolution of the output. In this approach, the coarse depth prediction network and the first refinement stage are trained jointly, with 3D supervision.

%

\vspace{6pt}
\paragraph{Using up-convolutional layers}

Dosovitskiy \etal~\cite{dosovitskiy2015flownet}  extended the approach of Eigen \etal~\cite{eigen2014depth} by removing the fully-connected layers. Instead, they pass the feature map, \ie the latent representation (of size $6\times8\times 1024$), directly into a decoder to regress the optical flow in the case of the FlowNetSimple of~\cite{dosovitskiy2015flownet}, and a depth map in the case of~\cite{eigen2015predicting}. 
In general, the decoder  mirrors the encoder. It also includes skip  connections,  \ie connections from some layers of the encoder to their corresponding counterpart in  the decoder. Dosovitskiy \etal~\cite{dosovitskiy2015flownet}   use variational refinement to refine the coarse optical flow. 


Chen \etal~\cite{Chen_2018_ECCV} used a similar approach to produce dense depth maps given an RGB image with known depth at a few pixels.  At training, the approach takes the ground-truth depth map and a binary mask indicating valid ground-truth depth pixels, and generates two other maps: the nearest-neighbor fill of the sparse depth map, and the Euclidean distance transform of the binary mask. These two maps are then concatenated together and with the input image and used as input to an encoder-decoder, which learns the residual that will be added to the sparse depth map. The network follows the same architecture as in~\cite{chen2016single}. Note that the same approach has been also used to infer other properties, \eg the optical flow as in Zhou \etal~\cite{zhou2016learning}.



%


\subsubsection{Combining and stacking multiple networks}
Several previous papers showed that stacking and combining multiple networks can lead to significantly improved performance. For example,  Ummenhofer and Zhou \cite{ummenhofer2017demon} introduced DeMoN,  which takes an image pair as input and predicts the depth map of the left image and the relative pose (egomotion) of the right image  with respect to the left. The network consists of a chain of three blocks  that iterate over optical flow, depth, and relative camera pose estimation. The first block in the chain, called bootstrap net, is composed of two encoder-decoder networks. It gets the image pair as input and then estimates, using the first encoder-decoder, the optical flow and a confidence map. These, along with the original pair of images, are fed to the second encoder-decoder, which outputs the initial depth and egomotion estimates.  The second component, called iterative net, is trained to improve, in a recursive manner, the depth, normal, and motion estimates. Finally, the last component, called refinement net, upsamples,  using and encoder-decoder network, the output of the iterative net to obtain high resolution depth maps.

Note that, unlike other techniques such as FlowNetSimple of~\cite{dosovitskiy2015flownet} which require a calibrated pair of images,  Ummenhofer and Zhou~\cite{ummenhofer2017demon}  estimates jointly the relative camera motion and the depth map.  


FlowNetSimple of~\cite{dosovitskiy2015flownet} has been later extended by Ilg~\etal~\cite{ilg2017flownet} to FlowNet2.0, which achieved results that are competitive with the traditional methods, but with an order of magnitude faster. The idea is to combine multiple FLowNetSimple networks to compute  large displacement optical flow. It (1) stacks multiple FlowNetSimple and FlowNetC networks~\cite{dosovitskiy2015flownet}. The flow estimated by each network is used to warp, using  a warping operator,  the right image onto the left image, and feed the concatenated left image, warped image, estimated flow, and the brightness error, into the next network.  This way, the next network in the stack can focus on learning the remaining increment between the left and right images, (2)  adds another FlowNetSimple network, called FlowNet-SD,  which focuses on small subpixel motion, and 
(3)  uses a learning schedule consisting of multiple datasets. The output of the FlowNet-SD  and the stack of multiple FlowNetSimple modules are merged together and processed using a fusion network, which provides the final flow estimation. 

Roy \etal~\cite{roy2016monocular} observed that  among all the training data sets  currently available, there is limited training data for some depths. As a consequence, deep learning techniques trained with these datasets   will naturally achieve low performance in the depth ranges that are under-represented in the  training data.   Roy \etal~\cite{roy2016monocular}   mitigate the problem by combining CNN with  a Neural Regression Forest.  A patch around a pixel  is processed with an ensemble of binary regression tree, called Convolutional Regression Tree (CRT). At every node of the CRT, the patch is processed with   a shallow  CNN associated to that node, and then passed to the left or right child node with a Bernouli probability for further convolutional processing. The process is repeated until the patch reaches the leaves of the tree.    Depth estimates made by every leaf are weighted with the corresponding path probability. The  regression results of every CRT are then fused into a final depth estimation. 

Chakrabarti~\cite{chakrabarti2016depth} combine global and local methods. The method first maps an input image into a latent representation of size $1\times 1096$, which is then reshaped into a feature map of size $427\times 562 \times 64$. In other words, each pixel is represented with   a global descriptor (of size $1\times 64$) that characterizes the entire scene. A parallel path takes patches of size $97\times 97$ around each pixel and computes a local feature vector of size $1\times 1024$. This one is then concatenated with the global descriptor of that pixel and fed into a top network. The whole network is trained  to predict, at every image location, depth derivatives of different orders, orientations, and scales.  However, instead of a single estimate for each derivative, the network outputs probability distributions that allow it to express confidence about some coefficients, and ambiguity about others.  Scene depth is then estimated by harmonizing this overcomplete set of network predictions, using a globalization procedure that finds a single consistent depth map that best matches all the local derivative distributions.

\subsubsection{Joint task learning}
\label{sec:joint_task_learning}

Depth estimation and many  other visual image understanding problems, such as segmentation,  semantic labelling, and scene parsing, are strongly correlated and mutually beneficial. Leveraging on the  complementarity properties of these tasks, many recent papers proposed to either jointly solve these tasks so that one boosts the performance of another.


To this end, Wang \etal~\cite{wang2015towards}  follow the CNN structure in~\cite{eigen2014depth} but adds additional semantic nodes, in the final layer, to predict the semantic label. Both depth estimation and semantic label prediction are trained jointly using a loss  function that is a weighted sum of the depth error and the semantic loss. The overall network is composed of a joint global CNN, which predicts, from the entire image,  a coarse depth and segmentation maps, and a regional CNN, which operates on image segments (obtained by over segmenting the input image) and predicts a more accurate depth  and segmentation labels within each segment. These two  predictions form unary terms to a hierarchical CRF, which produces the final depth and semantic labels. The CRF includes additional pairwise terms such as  pairwise edges between neighboring pixels, and pairwise edges between neighboring segments. 

Zhou \etal~\cite{Zou_2018_ECCV} follow the same idea to jointly estimate, from two successive images and in a non-supervised manner,  the depth map at each of them, the 6D relative camera pose, and the forward and backward optical  flows.   The approach uses three separate network, but jointly  trained using  a  cross-task consistency loss:  a DepthNet, which estimates depth from two successive frames,  PoseNet, which estimates the relative camera pose, and a FlowNet, which estimates the optical flow between the two frames. To handle non-rigid transformations that cannot be explained by the camera motion, the paper exploits the forward-backward consistency check to identify valid regions, \ie regions that moved in a  rigid manner,  and avoid enforcing the cross-task consistency in the non-valid regions.  

In the approaches of Wang \etal~\cite{wang2015towards} and Zhou \etal~\cite{Zou_2018_ECCV}, the networks (or network components) that estimate each modality do not directly share knowledge. Collaboration between them is only through a joint loss function. To enable information exchange between the different task, Xu \etal~\cite{Xu_2018_CVPR} proposed an approach that first maps the input  image into a latent representation using a CNN. The latent representation is then decoded, using four decoding streams, into a depth map, a normal map, an edge map, and a semantic label map. These multi-modal intermediate information is aggregated using  a multi-model distillation module, and t hen passed into  two decoders, one estimates   the refined depth map and the other one estimates the refined semantic  label map.  For the multi-model distillation module, Xu \etal~\cite{Xu_2018_CVPR} investigated three  architectures:
		\begin{itemize}
			\item Simple concatenation of the four modalities.
			\item Concatenating the four modalities and feeding  them to two different encoders. One encoder learns the features that are appropriate for inferring depth while the second learns features that are appropriate for inferring semantic labels.
			\item Using an attention mechanism to guide the message passing between the multi-modal features, before concatenating them and feeding  them to the encoder that learns the features for depth estimation or the one which learns the features for semantic labels estimation.  
		\end{itemize}

\noi Xu \etal~\cite{Xu_2018_CVPR} showed that the third option obtains remarkably better performance than the others.

Instead of a distillation module, Jiao \etal~\cite{Jiao_2018_ECCV}  proposed a  a synergy network whose backbone is a shared encoder. The network then  splits into two branches, one for depth estimation and another for semantic labelling. These two branches share knowledge through lateral sharing units . The network is trained with a  attention-driven loss, which guides the network to pay more attention to the distant depth regions during training. 

Finally, instead of estimating depth and semantic segmentation in a single iteration,  Zhang \etal~\cite{Zhang_2018_ECCV} performed it recursively, using Task-Recursive learning. It is composed of an encoder and a decoder network, with a series of residual blocks (ResNet), upsampling blocks, and Task-Attention Modules. The input  image is first fed into the encoder and then into the task-recursive decoding  to estimate depth and semantic segmentation.     In the decoder,  the two tasks (depth estimation and segmentation) are alternately processed by adaptively evolving previous experiences of both tasks to benefit each other. A task-attention module is used before each residual block and takes depth and segmentation features from the previous residual block as input.  It is composed of a balance unit, to balance the contribution of the features of the two sources. The balanced output is fed into a series of convolutional-deconvolutional layers designed to get different spatial attentions by using receptive  field variation.  The output is an attention map, which is used to generate the gated depth and segmentation features. These are fused  by concatenation followed by one convolutional layer. 		

While jointly estimating depth and other cues, \eg semantic labels,  significantly improves the performance of both tasks, it requires a large amount of training data annotated with depth and semantic labels. 

%


\section{Training}
\label{sec:training}

The training process aims to find the network parameters $\weights^*$  that minimize a loss function $\loss$, \ie:
\begin{equation}
	\weights^* = \argmin_{\weights} \loss(\estimateddisparitymap, \Theta, \weights).
\end{equation}

\noi Here,  $\Theta$ is the training data, which can be composed of input images, their associated camera parameters, and/or their corresponding ground-truth depth.       We will  review in Section~\ref{sec:datasets}  the different different datasets that have been used for training deep learning-based depth reconstruction algorithms, and for evaluating their performances. We will then review the different loss functions  (Section~\ref{sec:loss_functions}),  the degree of supervision required in various methods (Section~\ref{sec:degree_supervision}), and the domain adaptation and  transfer learning techniques (Section~\ref{sec:domain_adaptation}).




\subsection{Datasets and data augmentation}
\label{sec:datasets}

\begin{table*}[t]
	\caption{\label{tab:3ddatasets_sterep}\label{tab:3ddatasets_mvs}Datasets for depth/disparity estimation. "$\#f$" refers to the number of frames per video. "\#img./scene" refers to the number of images per scenes. We also refer the reader to~\cite{laga20183d} for more 3D datasets. }
	
	\resizebox{\linewidth}{!}{%
		\begin{tabular}{@{} l  l l l  l l l l@{ } l  llll@{} l  l@{}}
		\toprule 
		 \multirow{1}{*}{\textbf{Name}}  &  
		 \multirow{2}{*}{\textbf{Type}} &   
		 \multirow{2}{1.2cm}{\textbf{Frame size}}   & 
		  \multicolumn{4}{c}{\textbf{Number of (pairs of) images}}     & 
		  & 
		  \multicolumn{4}{c}{\textbf{Video (scene) }}   &  
		  &
		  \multirow{2}{1cm}{\textbf{Camera params}}  &
		  \multirow{2}{*}{\textbf{Disp/Depth}}  \\ 
  		  
		  \cline{4-7}  \cline{9-12}  \\ 
		
		& & & Total &  \#train & \#val &\#test & &    $\#f$ & \#train & \#test &\#img./scene &   & \\

		\toprule

		CityScapes~\cite{cordts2016cityscapes}  & real & $2048\times 1024$ &  $5000$ &  $2975$ & $ 500$ & $1525 $ & &  $-$ & $-$ & $-$ & $-$ & &$-$  & $-$ \\

		\hline
		KITTI  2015~\cite{menze2015object}  & real & $1242\times375$ & $-$ &  $-$ & $-$ & $-$   & & $400$ &  $200$&  $200$& $4$ & & int+ext& sparse\\

		\hline
		
		KITTI  2012~\cite{geiger2012we}  & real& $1240\times376$ & $ $389$ $ & $194$ & $-$ & $195$ & & $-$&$-$ &$-$ &$-$ & & int+ext & sparse \\

		\hline
		FlyingThings3D~\cite{mayer2016large} & synth. & $960 \times 540$  &  $26066$ &  $21818$&  $-$ & $4248$ &  & $-$  &  $2247$ &$-$ &$-$ & & int+ext & per-pixel\\
		
		\hline
		Monkaa~\cite{mayer2016large} & synth. & $960 \times 540$  &  $8591$ &  $8591$& $-$  & $-$ & & $-$&  $8$& $-$ & $-$ & &int+ext & per-pixel\\
		\hline
		Driving~\cite{mayer2016large} & synth. & $960 \times 540$  &  $4392$ &  $4392$& $-$ & $-$ &   &$-$ &  $1$ & $-$ & $-$ & & int+ext & per-pixel\\
		
		\hline
		MPI Sintel~\cite{butler2012naturalistic}  & synth.  & $1024 \times 436$ & $1041$ & $1041$ & $-$&   $-$ & & $35$ & $23$ & $12$ & $50$  & & $-$ & per-pixel\\
		\hline
		
		\hline
		SUN3D~\cite{xiao2013sun3d} &  rooms & $640\times 480$  & $-$ & $2.5$M & $-$ & $-$   &   & $-$& $415$ & $-$ &$-$ &   & ext. &  per-pixel\\
		
		\hline
		NYU2~\cite{silberman2012indoor} & indoor  & $640\times 480$  & $-$ & $1449$ & $-$ & $-$   &   & $-$& $464$ & $-$ &$-$ &   & no &  per-pixel\\

		\hline
		RGB-D SLAM~\cite{sturm2012benchmark} &  real & $640\times 480$  & $-$ & $-$ & $-$ & $-$   &   & $-$& $15$ & $4$ & variable &   & int+ext &  per-pixel \\ 

		\hline
		MVS-Synth~\cite{huang2018deepmvs} &   Urban& $1920\times1080$  & $-$ & $-$ & $-$ & $-$   &   & $120$& $-$ & $-$ &$100$ &   & int+ext &  per-pixel\\

		\hline
		ETH3D~\cite{schops2017multi} &  in/outdoor &    $713\times438$ & & $27$ & $-$ & $20$   &   & $ $& $5$ & $5$ &$-$ &   & int+ext & point cloud \\  
		
		
		\midrule
		DTU~\cite{aanaes2016large} & MVS & $1200\times 1600$&  & & & & & $80$&  & & $49-64$ & & int+ext & \\
		
		\hline
		{MVS KITTI2015~\cite{menze2015object}}  & MVS  &  & & & & & & $200 + 200$ &  & &  $20$ & &  $-$ & \\
		\hline
		ETH3D~\cite{schops2017multi} & MVS &  $6048\times4032$ & & & & & & $13+12$  & & &variable & & $-$ &  \\
		
		\hline
		Make3D~\cite{saxena2009make3d}  & Single view& $2272\times 1704$ & $534$  & $400$ & $-$ & $134$ & & $-$& $-$ & $-$ & $-$ &       & $-$  & $55\times305$\\ 	
		
		\hline		
		MegaDepth~\cite{Li_2018_CVPR} & Single, MVS & $-$ & $130$K  & $-$ & $-$ & $-$ & & $196$& $-$ & $-$ & $-$ &       & $-$  &  Eucl., ordinal \\		
		\bottomrule
		
		\end{tabular}
		
		}
\end{table*}
%
%
%
%
%

Unlike traditional 3D reconstruction techniques, training and evaluating deep-learning architectures for depth reconstruction require large amounts of annotated data. This annotated data should be in the form of natural images and their corresponding depth maps, which is very challenging to obtain. Tables~\ref{tab:3ddatasets_sterep}  summarizes some of the datasets that have been used in the literature. Some of them  have been specifically designed to train, test, and benchmark stereo-based depth reconstruction algorithms. They usually contain pairs of stereo images of real or synthesized scenes, captured with calibrated cameras, and their corresponding disparity/depth information as ground truth. The disparity/depth information can be either in the form of  maps at the same or lower esolution as the input images, or in the form of sparse depth values at some locations in the reference image. Some of these datasets contain video sequences and thus are suitable for benchmarking Structure from Motion (SfM) and Simultaneous Localisation and Mapping (SLAM) algorithms.

The  datasets that were particularly designed to train and benchmark multiview stereo  and single view-based reconstruction algorithms (MVS)  are  composed of multiple scenes with  $\nimages \ge 1$ images per scene. Each image is captured from a different viewpoint. 

In general, deep-learning models achieve good results if trained on large datasets. Obtaining the ground-truth depth maps  is, however,  time-consuming and resource intensive.  To overcome this limitation, many papers collect data from some existing datasets and augment them with suitable information and annotations to  make them suitable for training and testing deep learning-based depth reconstruction techniques. In general, they use the four following strategies:
\begin{itemize}
	
	\item \textbf{3D data augmentation. } To introduce more diversity to the training datasets, one can apply to the existing datasets some geometric and photometric  transformations, \eg translation, rotation, and scaling,  as well as additive Gaussian noise and changes in brightness, contrast, gamma, and color. Although some transformations are similarity preserving, they still enrich the datasets.  One advantage of this approach is that it reduces  the network's generalization error.  Also statistical shape analysis techniques~\cite{laga2017numerical,jermyn2017elastic,wang2018shape} can be used to synthesize more 3D shapes from existing ones.
	
	\item \textbf{Using synthesized 3D models and scenes. } One approach to generate image-depth annotations is by synthetically rendering from 3D CAD models 2D and 2.5D views from various (random) viewpoints, poses, and lighting conditions. They can also be overlayed with random textures. 
		
	\item \textbf{Natural image - 3D shape/scene pairs.  } Another approach is to synthesize training data by overlaying images rendered from large 3D model collections on the  top of real images such as those in the SUN~\cite{xiao2010sun}, ShapeNet~\cite{chang2015shapenet}, ModelNet~\cite{wu20153d}, IKEA~\cite{lim2013parsing}, and PASCAL 3D+~\cite{xiang2014beyond} datasets. 
	
\end{itemize}

\noi While the last two techniques allow enriching existing training datasets, they suffer from domain bias and thus require using domain adaptation techniques, see Section~\ref{sec:domain_adaptation}. Finally, some papers overcome the need for ground-truth depth information by training their deep networks without 3D supervision, see Section~\ref{sec:degree_supervision}.

\subsection{Loss functions}
\label{sec:loss_functions}

The role of the loss function is to measure at each iteration how far the estimated  disparity/depth map $\estimateddisparitymap$ is  from the real map $\disparitymap$,  and use it to guide  the update of the network weights.  In general, the loss function is defined as the sum of two terms:
\begin{equation}
	\loss(\estimateddisparitymap, \Theta, \weights) = \loss_1(\estimateddisparitymap, \Theta, \weights)  + \loss_2(\estimateddisparitymap, \Theta, \weights).
	\label{eq:overall_loss} 
\end{equation}

\noi The data term $\loss_1$  measures the error between the ground truth and the estimated depth while the regularization term $\loss_2$  is used to incorporate various constraints, \eg smoothness.  To ensure robustness to spurious outliers,   some techniques, \eg\cite{zhou2017unsupervised}, use a truncated loss, which is defined at each pixel $\pixel$ as $\min (\loss_\pixel, \psi)$. Here, $\loss_\pixel$  denotes the non-truncated loss at pixel $\pixel$, and $\psi$ is a pre-defined threshold.

There are various  loss functions that have been used in the literature. Below, we list the most popular ones. Tables~\ref{tab:performance_stereomatching_ktti15},~\ref{tab:performance_mvs_ktti15}, and~\ref{tab:performance_depthregression} show how these terms have been used to train  depth estimation pipelines. 

\subsubsection{The data term} 
The data term of Equation~\eqref{eq:overall_loss} measures the error between the ground truth and the estimated depth. Such error can be quantified using one or  a weighted sum of two or more of the  error measures described below. The $\ltwo$, the mean absolute difference, the cross-entropy loss, and the Hinge loss  require 3D supervision while  re-projection based losses  can be used without 3D supervision since they do not depend on ground truth depth/disparity.

\vspace{6pt}
\noi \textbf{(1) The $\ltwo$ loss} is defined as 
		\begin{equation}
			\lossltwo= \frac{1}{\npixels} \sum_{\pixel} \|\disparitymap(\pixel) -  \estimateddisparitymap(\pixel)\|^2,
			\label{eq:loss_ltwo}
		\end{equation}
		where $\npixels$ is the number of pixels being considered. 
		
\vspace{6pt}
\noi \textbf{(2) The mean absolute difference (mAD)}  between the ground truth and the predicted disparity/depth maps~\cite{kendall2017end,ummenhofer2017demon,pang2017cascade,cheng2018learning} is defined as follows;
		\begin{equation}
			\lossabsdisparitydiff= \frac{1}{\npixels} \sum_{\pixel} \|\disparitymap(\pixel) -  \estimateddisparitymap(\pixel)\|_1.
			\label{eq:loss_abs_disparity_diff}
		\end{equation}
		Many variants of this loss function have been used. For instance,  Tonioni \etal~\cite{tonioni2017unsupervised} avoid explicit 3D supervision by taking $\disparity_\pixel$  as the disparity/depth at pixel $\pixel$ computed using traditional stereo matching techniques,  $\estimateddisparitymap(\pixel)$ as the estimated disparity,  and $\confidence_\pixel$ as the confidence of the  estimate at $\pixel$. They then define a \emph{confidence-guided loss} as:
		\begin{equation}
		\small{
			\confidenceguidedloss = \frac{1}{\npixels} \sum_\pixel L(\pixel),  \text{  }
			L(\pixel) =  \left\{  \begin{tabular}{@{}l@{}l@{}}
						$\confidence_\pixel |\disparity_x - \estimateddisparity_\pixel|$  &  { if} $\confidence_\pixel \ge \epsilon$,\\
						$0$ & { }otherwise.
			\end{tabular}	
			\right.
		}
		\end{equation}
		
		\noi Here, $\epsilon$ is a user-defined threshold. 
				
		Yao \etal~\cite{yao2018mvsnet}, which first estimate an initial depth map $\estimateddisparitymap_0$ and then the refined one $\estimateddisparitymap$, define the overall loss as the weighted sum of  the mean absolute difference between the ground truth $\disparitymap$  and $\estimateddisparitymap_0$, and the ground truth $\disparitymap$  and $\estimateddisparitymap$:
		\begin{equation}
			\small{\lossweightedmeanabs  = \frac{1}{\npixels} \sum_{\pixel}\left\{  \|\disparity(\pixel) -  \estimateddisparity_0(\pixel)\|_1  +  \lambda\|\disparity(\pixel) -  \estimateddisparity(\pixel)\|_1  \right\}}.
			\label{eq:loss_abs_disparity_diff_weighted}
		\end{equation}
		Here, $\disparity_x = D(x)$, and $\lambda$ is a weight factor, which is set to one in~\cite{yao2018mvsnet}.  Khamis \etal~\cite{khamis2018stereonet}, on the other hand, used the two-parameter robust function $\rho(\cdot)$, proposed in~\cite{barron2017more},   to approximate a smoothed $\lone$ loss. It is defined as follows:
		\begin{equation}
		\small{
			\begin{tabular}{c}
			$\lossapproximatesmoothlone = \frac{1}{\npixels} \sum_{\pixel} \rho(\disparity_{\pixel} -  \estimateddisparity_\pixel, \alpha, c), \text{ where } \alpha = 1, c=2, \text{ and } $\\
			$\rho(x, \alpha, c) = \frac{|2 - \alpha| }{\alpha}\left( \left(  \frac{x^2}{c^2  | 2 - \alpha| }  + 1   \right) ^ {\frac{\alpha}{2}}  - 1 \right).$
			\end{tabular}
		}
			\label{eq:loss_two_params_robust}
		\end{equation}
		
		
\noi Other papers, \eg~\cite{chang2018pyramid}, use the smooth $\lone$ loss, which is widely used in bounding box regression for object detection because of its robustness and low sensitivity to outliers. It is defined as:
		\begin{equation}
			\losssmoothlone = \frac{1}{\npixels} \sum_{\pixel} \text{smooth}_{\lone}(\disparity_\pixel - \estimateddisparity_\pixel), 
			\label{eq:smooth_lone_loss}
		\end{equation}
		where 
		$
			\text{smooth}_{\lone}(x) =
				\left\{  \begin{tabular}{ll}
					$0.5x^2$  & $\text{ if } |x| < 1$, 	 \\
					$|x| - 0.5$  & $\text{ otherwise}$.
				\end{tabular}	
				\right.
		$

Note that some papers restrict the sum to be  over valid pixels in order to avoid outliers, or over regions of interests, \eg foreground or visible pixels~\cite{zhou2016learning}.

\vspace{6pt}
\noi \textbf{(3) The cross-entropy loss~\cite{luo2016efficient,huang2018deepmvs}. } It is defined as:
		\begin{equation}
			\losscrossentropy = - \sum_{\pixel} Q(\disparity_\pixel, \estimateddisparity_\pixel ) \log\left( P(\pixel, \estimateddisparity_\pixel )  \right).
			\label{eq:cross_entropy_loss}
		\end{equation}
		Here, $P(\pixel, \estimateddisparity_\pixel ) $ is the likelihood, as computed by the network,  of pixel $\pixel$ having the disparity/depth $\estimateddisparity_\pixel$.  It is defined in~\cite{luo2016efficient,huang2018deepmvs} as the 3-pixel error:
		\begin{equation}
			Q(\disparity_\pixel, \estimateddisparity_\pixel ) = \left\{ 
					\begin{tabular}{ll}
						$\lambda_1$ & \text{if }  $\disparity_\pixel = \estimateddisparity_\pixel$ \\
						$\lambda_2$ & \text{if }  $|\disparity_\pixel - \estimateddisparity_\pixel  | = 1$\\
						$\lambda_3$ & \text{if }  $|\disparity_\pixel - \estimateddisparity_\pixel  | = 2$\\
						$0$ & \text{otherwise. }  \\
					\end{tabular}
				\right.
		\end{equation}	
		Luo \etal~\cite{luo2016efficient} set $\lambda_1=0.5$, $\lambda_2=0.2$, and $\lambda_3=0.05$.

\vspace{6pt}
\noi \textbf{(4) The sub-pixel cross-entropy loss $\losssubpixelcrossentropy$. }  This loss, introduced by Tulyakov \etal~\cite{tulyakov2017weakly},  enables faster convergence and better accuracy at the sub-pixel level. It is defined using a discretized Laplace distribution centered at the ground-truth disparity:
\begin{equation}
	\small{
	Q(\disparity_\pixel, \estimateddisparity_\pixel ) = \frac{1}{Z} e^{- \frac{1}{b} |\disparity_\pixel - \estimateddisparity_\pixel| }, \text{ } Z = \sum_\disparity e^{- \frac{1}{b} |\disparity_\pixel - \disparity| }.
	}
\end{equation}

\noi Here, $\disparity_\pixel$ is the ground-truth disparity at pixel $\pixel$ and $\estimateddisparity_\pixel$ is the estimated disparity at the same pixel.

\vspace{6pt}
\noi \textbf{(5) The hinge loss criterion~\cite{zbontar2016stereo,shaked2017improved}.} It is computed by considering
pairs of examples centered around the same image position where one example belongs to the positive and one to the negative class. Let $s_+$ be the output of the network for the
positive example, $s_-$  be the output of the network for the negative example, and let $m$, the margin, be a positive real number. The hinge loss for that pair of examples is defined as:
\begin{equation}
	\losshinge = max(0, m + s_- - s_+). 
\end{equation} 

\noi It  is zero when the similarity of the positive example is greater  than the similarity of the  negative example by at least the margin $m$, which is set to $0.2$ in~\cite{zbontar2016stereo}.

\vspace{6pt}
\noi\textbf{(6) The re-projection (inverse warping) loss. }  Obtaining 3D ground truth data is very expensive. To overcome this issue, some techniques measure the loss based on the re-projection error.  The rational is that if the estimated disparity/depth map is as close as possible to the ground truth, then the discrepancy between the reference image and any of the other images but unprojected using the estimated depth map onto the reference image, is also minimized.   It can be defined in terms of the photometric error~\cite{bai2016exploiting,zhou2017unsupervised}, also called per-pixel $\lone$ loss~\cite{flynn2016deepstereo}, or image reconstruction error~\cite{zhong2017self}. It is defined as the $\lone$ norm between the reference image $\referenceimage$ and  $ \tilde{\image}_t$, which is $\image_t$  but unwarped onto $\referenceimage$ using the camera parameters:
    	\begin{equation}
    		\inversewarploss= \frac{1}{\npixels} \sum_{\pixel} \| \referenceimage(\pixel) -  \tilde{\image}_t(\pixel)\|_1.
		\label{eq:photometric_loss}
    	\end{equation}
	
\noi It can also be defined using the distance between the features $\featuremap$ of the reference image and the features $\tilde{\featuremap}_t$ of any of the other images but unwarped onto the view of the reference image using the camera parameters  and the computed depth map~\cite{yang2018segstereo}:
		\begin{equation}
			\inversefeaturewarploss= \frac{1}{\npixels} \sum_{\pixel} \| \featuremap_(\pixel) -  \tilde{\featuremap}_t(\pixel)\|_1. 
		\end{equation}

\noi Other terms can be added to the re-projection loss. Examples include the $\lone$ difference between the gradients of $\referenceimage$ and  the gradient of $ \tilde{\image}_t$~\cite{zhong2017self}:
\begin{equation}
	\lossgradient = \frac{1}{N}\sum_{\pixel} \|\nabla \image_{ref}(\pixel)  - \tilde{\image}_t(\pixel)\|_1, 
\end{equation}

\noi  and the structural dissimilarity between patches in $\referenceimage$ and  in $ \tilde{\image}_t$~\cite{zhong2017self,wang2004image}. We denote this loss by $\lossstructuraldisspatch$.

\vspace{6pt}
\noi\textbf{(7) Matching loss. } Some methods, \eg\cite{zbontar2016stereo,chen2015deep,shaked2017improved}, train the feature matching network separately from the subsequent disparity computation and refinement blocks, using different loss functions. Chen \etal~\cite{chen2015deep} use a loss that measures the $\ltwo$ distance between the predicted matching score and the ground-truth score:
\begin{equation}
	\lossltwomatching = \| \text{predicted\_score}(\pixel, \disparity) -  \text{label}(\pixel, \disparity)\|,
\end{equation}	

\noi where $\text{label}(\pixel, \disparity)  \in \{0, 1\}$ is the ground-truth label indicating whether the pixel $\pixel (i, j)$ on the left image corresponds to the pixel $(i - \disparity, j)$ on the right image, and $\text{predicted\_score}(\pixel, \disparity)$ is the predicted matching score for the same pair of pixels.

\vspace{6pt}
\noi\textbf{(8) The semantic loss. }  Some papers incorporate semantic cues, \eg segmentation~\cite{yang2018segstereo} and edge~\cite{song2018stereo} maps, to guide the depth/disparity estimation.  These can be either provided at the outset, \eg estimated with a separate method as in~\cite{song2018stereo}, or  estimated jointly with the depth/disparity map using the same network trained end-to-end. The latter case requires a  semantic loss. For instance,   Yang \etal~\cite{yang2018segstereo}, which use segmentation as semantics, define the semantic loss $\lossltwomatching $ as the distance between the classified warped maps and ground-truth labels.  Song \etal~\cite{song2018stereo}, which use edge probability map as semantics, define the semantic loss as follows;
			\begin{equation}				
				\small{\lossedgesemantic = \frac{1}{\npixels} \sum_{\pixel}\left\{  |\partial_u \disparity_\pixel | e^{-| \partial_u \xi_\pixel |  } +  |\partial_v \disparity_\pixel | e^{-| \partial_v \xi_\pixel |  }\right\},} 
			\end{equation}	

\noi where $\pixel=(u, v)$ and $\xi$ is the edge probability map.

\subsubsection{The regularization term} 

In general, one can make many assumptions about the disparity/depth map and incorporate them into the regularization term of Equation~\eqref{eq:overall_loss}. Examples of constraints include: smoothness~\cite{zhong2017self}, left-right consistency~\cite{zhong2017self},  maximum depth~\cite{zhong2017self}, and scale-invariant gradient loss~\cite{ummenhofer2017demon}. The regularization term can then be formed using a weighted sum of these losses.

\vspace{6pt}
\noi\textbf{(1) Smoothness. } It can be measured using the magnitude of the first or second-order gradient of the estimated disparity/depth map.  For instance, Yang \etal~\cite{yang2018segstereo} used the $\lone$ norm of the first-order gradient:
		\begin{equation}
			\small{ \losslonegradient = \frac{1}{N} \sum_x \left\{ (\nabla_u \disparity_x ) +  (\nabla_v \disparity_x)\right\}, \pixel = (u, v).}
			\label{eq:loss_gradient_diff}
		\end{equation}	
	
\noi Here  $\nabla$ is the gradient operator. Zhou \etal~\cite{zhou2017unsupervised}   and Vijayanarasimhan \etal~\cite{vijayanarasimhan2017sfm} define smoothness as the $\ltwo$ norm of the second-order gradient:
		\begin{equation}
			\ltwosmoothnessgradient= \frac{1}{N} \sum_x \left\{ (\nabla_u^2 \disparity_x )^2 +  (\nabla_v^2 \disparity_x)^2\right\}.
			\label{eq:smoothness_2ndgradient}
		\end{equation}	
		
\noi Zhong \etal~\cite{zhong2017self} used the second-order gradient   but weighted with the image's second-order gradients:
		\begin{equation}
			\small{\lossweightsecondgradient = \frac{1}{N} \sum \left\{ | \nabla_u^2 \disparity_x | e^{-| \nabla_u^2 \leftimage(x)  |} +  | \nabla_v^2 \disparity_x | e^{-| \nabla_v^2 \leftimage(x)  |} \right\}.}
			\label{eq:smoothness_2ndgradient2}
		\end{equation}

\noi Finally, Tonioni \etal~\cite{tonioni2017unsupervised} define the smoothness at a pixel $\pixel$ as the absolute difference between the disparity predicted at $\pixel$ and those predicted at each pixel $y$ within a certain predefined neighborhood $\mathcal{N}_\pixel$ around the pixel  $\pixel$. This is then averaged over all pixels:
\begin{equation}
	\lossavgbhblone= \frac{1}{\npixels} \sum_{\pixel} \sum_{y \in \mathcal{N}_\pixel} |\disparity_\pixel - \disparity_y |.  
\end{equation}

\vspace{6pt}
\noi\textbf{(2) Consistency. } Zhong \etal~\cite{zhong2017self}  introduced the loop-consistency loss, which is constructed as follows; Consider the left image $\leftimage$ and the  synthesized image $\tilde{\image}_{left}$ obtained by warping the right image to the left image coordinate with the disparity map defined on the right image. A second synthesized left image $\tilde{\tilde{\image}}$   is generated by warping the left image to the right image coordinates by using the disparities at the left and right images, respectively. The loop consistency loss consistency term is then defined as:
		\begin{equation}
			\lossloopconsistency  = | \tilde{\tilde{\image}} - \leftimage|.
			\label{eq:loop_consistency}
		\end{equation}

\noi Godard \etal~\cite{godard2017unsupervised}  introduced the left-right consistency term, which attempts to make the left-view disparity map be equal to the projected right-view disparity map. It can be seen as a linear approximation of the loop consistency and is defined as follows; 
		\begin{equation}
				\lossleftrightconsistency	 = \frac{1}{\npixels} \sum_{\pixel} | \disparity_{\pixel}  - \tilde{\disparity}_{\pixel} |,
				\label{eq:leftrightconsistency}
		\end{equation}
		where $\tilde{\disparity}$ is the disparity at the right image but reprojected onto the coordinates of the left image. 
		
\vspace{6pt}
\noi\textbf{(3) Maximum-depth heuristic. }  There may be multiple warping functions that achieve similar warping loss, especially for textureless areas. To provide strong regularization in these areas, Zhong \etal~\cite{zhong2017self}  use the Maximum-Depth Heuristic (MDH)~\cite{perriollat2011monocular}, which is defined as the sum of all depths/disparities:
		\begin{equation}
			\lossmdh = \frac{1}{\npixels} \sum_{\pixel} | \disparity_{\pixel} |.
			\label{eq:mdh}
		\end{equation}
		
\vspace{6pt}
\noi\textbf{(4) Scale-invariant gradient loss~\cite{ummenhofer2017demon},  }  defined as:
		\begin{equation}
			\lossscaleinvariantgradient  = \sum_{h \in A} \sum_{\pixel} \| g_h[\disparitymap](\pixel)   - g_h[\estimateddisparitymap](\pixel) \|_2,
		\end{equation}
\noi where $A = \{1,2, 4, 8, 16 \}$,  $\pixel = (i, j)$, $f_{i,j} \equiv f(i, j)$, and
		\begin{equation}
			\small{g_h[f](i, j)  = \left( \frac{f_{i+h, j} - f_{i, j} }{| f_{i+h, j} - f_{i, j} |},  \frac{f_{i, j+h} - f_{i, j} }{| f_{i,  j+h} - f_{i, j} |} \right)^\top.}
		\end{equation}






\subsection{Degree of supervision}
\label{sec:degree_supervision}

Supervised methods for depth estimation, which have achieved promising results, rely on large quantities of ground truth depth data. However, obtaining ground-truth depth data, either manually or using  traditional stereo matching algorithms or 3D scanning devices, \eg Kinect,  is  extremely difficult and expensive, and is prune to noise and inaccuracies. Several mechanisms have been recently  proposed in the literature to make the 3D supervision as light as possible. Below, we discuss the most important ones. 

\subsubsection{Supervision with stereo images}

Godard \etal~\cite{godard2017unsupervised} exploit the left-right consistency to perform unsupervised depth estimation from a monocular image. The approach is trained, without 3D supervision, using stereo pairs. At runtime, it only requires one input image and returns the disparity map from the same view as the input image. For this, the approach uses the left-right consistency loss of Equation~\eqref{eq:leftrightconsistency}, which attempts to make the left-view disparity map be equal to the projected right-view disparity map.  

Tonioni \etal \cite{tonioni2017unsupervised} fine-tune pre-trained networks without any 3D supervision by using stereo pairs. The idea is leverage on traditional stereo algorithms and state-of-the-art confidence measures in order to fine-tune a deep stereo model based on disparities provided by standard stereo algorithms that are deemed as highly reliable by the confidence measure. This is done by minimizing a  loss function  made out of two terms: a confidence-guided loss  and a smoothing
term.

Note that while stereo-based supervision does not require ground-truth 3D labels, these techniques usually rely on the availability of  calibrated stereo pairs during training.  
 
\subsubsection{Supervision with camera's aperture}

Srinivasan \etal~\cite{Srinivasan_2018_CVPR}'s approach uses  as supervision the information provided by a camera's aperture.   It introduces two differentiable aperture rendering functions that use the input image and the predicted depths to simulate  depth-of-field effects caused by real camera apertures. The  depth estimation network is trained end-to-end to predict the scene depths that best explain these finite aperture images as defocus-blurred renderings of the input all-in-focus image.

\subsubsection{Training with relative/ordinal depth annotation}

People, in general, are better at judging relative depth~\cite{todd2003visual}, \ie assessing whether a point $A$ is closer than point $B$.  
Chen \etal~\cite{chen2016single} introduced an algorithm for learning to estimate metric depth using only annotations of relative depths. In this approach each image is annotated with only the ordinal relation between a pair of pixels, \ie point $A$ is closer to point $B$, further than $B$, or it is hard to tell.   To train the network, a ConvNet in this case, using such ordinal relations,  Chen \etal~\cite{chen2016single}  introduced an improved ranking loss, which encourages the predicted depth  to agree with the ground-truth ordinal relations. It is defined as:
\begin{equation}
	\lossimprovedranking(\estimateddisparitymap, \Theta, \weights) = \sum_{i=1}{N} \omega_k\loss_k(\image, x_k, y_k, l_k, \estimateddisparity),
 \end{equation}
where $\omega_k$, $l_k$, and $\loss_k$ are, respectively, the weight, the label,  and loss of the $k-$th  pair $(x_k, y_i)$ defined as:
\begin{equation}
	\small{
	\loss_k =   \left\{  \begin{tabular}{ll}
					$\log(1 + \text{exp}( (-\estimateddisparity_{xk} + \estimateddisparity_{yk}) l_k)$, & if $l_k \neq 0$, \\ 
					$(\estimateddisparity_{xk}  -\estimateddisparity_{yk} )^2$, & otherwise. 
				\end{tabular}
			\right.
		}
\end{equation}

\noi Xian \etal~\cite{Xian_2018_CVPR} showed that training with only one pair  of ordinal relation  for each image is not sufficient to get satisfactory results. They then extended this approach by annotating each image with 3K pairs and showed that they can achieve substantial improvement  accuracy.  The challenge, however, is how to cheaply get such large number of relative ordinal annotations. They propose to use optical flow  maps from web stereo images. 
Note that at runtime, both methods take a single image of size  $384\times 384$  and output a dense depth map of the same size.

\subsubsection{Domain adaptation and transfer learning}
\label{sec:domain_adaptation}

Supervised deep learning often suffers from the lack of sufficient training data. Also, when using range sensors, noise is often present and the measurements can be very sparse. Kuznietsov \etal~\cite{kuznietsov2017semi} propose an approach to depth map prediction from monocular images that learns in a semi-supervised way. The ida is to use  sparse ground-truth depth for supervised learning, while enforcing the deep network to produce photoconsistent dense depth maps in a stereo setup using a direct image alignment / reprojection loss.

While obtaining ground-truth depth annotations of real images is challenging and time consuming, synthetic images with their corresponding depth maps can be easily generated using computer graphics techniques. However,  the domain of real images is different from the domain of graphics-generated images.  Recently, several domain adaptation strategies have been  proposed to solve this domain bias issue. These allow training on synthetic data and transfer what has been learned to the domain of real images.

Domain adaptation methods for depth estimation can be classified into two  categories. Methods in the first category transform the data of one domain to look similar in style to the data in the other domain. For example, Atapour-Abarghoue \etal~\cite{Atapour-Abarghouei_2018_CVPR} proposed a two-staged approach.   The first stage includes training a depth estimation model using synthetic data. The second stage is  trained to transfer the style of synthetic  images to real-world images. 	By doing so, the style of real images is first transformed to match the style of synthetic data and then fed into the depth estimation network, which has been trained on synthetic data.  Zheng \etal~\cite{Zheng_2018_ECCV} performed  the opposite;  it transforms the synthetic images to become more realistic and use them to train the depth estimation network.  Guo \etal~\cite{Guo_2018_ECCV}, on the other hand,  trains a stereo matching network using synthetic data to predict occlusion maps and disparity maps of stereo image pairs.  In a second step, a monocular depth estimation network is trained on real data by distilling the knowledge of the stereo network~\cite{hinton2015distilling}. These methods use adversarial learning.

Methods in the second class operate on the network architecture and the loss functions used for their training.  Kundu \etal~\cite{Kundu_2018_CVPR} introduced AdaDepth, an unsupervised mechanism for domain adaptation. The approach uses an encoder-decoder architecture of the form $\estimateddepthmap = \decodingfunc\left( \encodingfunc(\images_s)\right)$. It is first trained, in a supervised manner, using synthetic data $\images_s$.  Let $\decodingfunc_s$ and $\encodingfunc_s$ be the decoding and encoding functions learned from synthetic data. Let $\decodingfunc_t$ and $\encodingfunc_t$ be the decoding and encoding functions that correspond to real data $\images_t$. Kundu \etal~\cite{Kundu_2018_CVPR} assume that $\decodingfunc_t = \decodingfunc_s = \decodingfunc$. Its goal  is to match the distributions of the latent representations generated by $\encodingfunc_s$ and $\encodingfunc_t$. This is done by initializing the network with the weights that have been learned using synthetic data.   It then uses adversarial learning to minimize an objective function that discriminates between $\decodingfunc_t(\images_t)$ and  $\decodingfunc_s(\images_s)$, and another objective function that discriminates between $\estimateddepth_s$ and $\decodingfunc (\encodingfunc_t(\images_t))$. The former ensures that real data, when fed to the encoder, are mapped to the same latent space as the one learned during training. The latter ensures that inferences through the corresponding transformation functions $\decodingfunc(\encodingfunc_s(\cdot))$ and $\decodingfunc(\encodingfunc_t(\cdot))$ are directed towards the same output density function.

\section{Discussion and comparison}
\label{sec:discussion_and_comparison}
%
\label{sec:performance_metrics}

This section discusses some state-of-the-art techniques using  quantitative and qualitative performance criteria. 

\subsection{Evaluation metrics and criteria}

The most commonly used quantitative metrics for evaluating the performance of a depth estimation algorithm include (the lower these metrics are the better); 
\begin{itemize}
	\item \textit{Computation time}  at training and runtime. While one can afford large computation time during training, some applications may require realtime performance at runtime.

	\item \textit{Memory footprint.}  In general deep neural networks have a large number of parameters. Some of them operate on volumes  using 3D convolutions. This would require large memory storage, which can affect their performance at runtime. 
	
	\item \textit{The End-Point Error (EPE).}  Called also geometric error, it is defined as the  distance between the ground truth $\depthmap$ and the predicted disparity/depth $\estimateddepthmap$, \ie $EPE(\depthmap, \estimateddepthmap) = \|  \depthmap - \estimateddepthmap\|$. This metric has two variants: Avg-Noc and Avg-ll. The former is measured in non-occluded areas while the latter is measured over the entire image.

	\item \textit{Percentage of Erroneous pixels (PE). } It is defined as the percentage of pixels where the true  and  predicted disparity/depth differ with more than a predefined threshold $\threshold$. Similar to the EPE, this error can be measured on non-occluded ares (Out-Noc) and over the entire image (Out-All).

	\item \textit{The bad pixel error (D1). }  It is defined as the percentage of disparity/depth errors below a threshold. This metric is computed in non-occluded (Noc) and in all pixels (All), in background (bg) and in foreground (fg) pixels. 
		
	
	\item \textit{Absolute relative difference. } It is defined as the average over all the image pixels of the $\lone$ distance between the groud-truth and the estimate depth/disparity, but scaled by the estimated depth/disparity:
		\begin{equation}
			\text{Abs rel. diff} = \frac{1}{\npixels}\sum_{\npixels} \frac{| \depth_i- \estimateddepth_i |}{\estimateddepth_i}.
		\end{equation}
	
	\item \textit{Squared relative difference. } It is defined as the average over all the image pixels of the $\ltwo$ distance between the groud-truth and the estimate depth / disparity, but scaled by the estimated depth/disparity:
		\begin{equation}
			\text{Abs rel. diff} = \frac{1}{\npixels}\sum_{\npixels} \frac{| \depth_i- \estimateddepth_i |^2}{\estimateddepth_i}.
		\end{equation}
		
	\item \textit{The linear Root Mean Square Error. } It is defined as follows:
		\begin{equation}
			\text{RMSE(linear)} = \sqrt{ \frac{1}{\npixels} \sum_{\npixels} {| \depth_i- \estimateddepth_i |}^2 }.
		\end{equation}
	
	\item \textit{The log Root Mean Square Error. } It is defined as follows:
		\begin{equation}
			\text{RMSE(log)} = \sqrt{ \frac{1}{\npixels} \sum_{\npixels} {| \log \depth_i- \log \estimateddepth_i |}^2 }.
		\end{equation}
\end{itemize}

\noi The accuracy is generally evaluated using the following metrics (the higher these metrics are the better);
\begin{itemize}

	\item \textit{Maximum relative error. } It is defined as the percentage of pixels $i$ such that
		\begin{equation}
			\max \left(\frac{\depth_i}{\estimateddepth_i}, \frac{\estimateddepth_i}{\depth_i} \right) < \epsilon,
		\end{equation}
		
		\noi where $\epsilon$ is a user-defined threshold. It is generally set to $1.25$, $1.25^2$, and $1.25^3$.
	
	\item \textit{Density. }  It is defined as the percentage of pixels for which depth has been estimated.
	
\end{itemize}

\noi In addition to these quantitative  metrics, there are several qualitative aspects to consider. Examples include;
\begin{itemize}
	\item \textit{Degree of 3D supervision. }  One important aspect of deep learning-based depth reconstruction methods is the degree of 3D supervision they require during training. In fact, while obtaining multiview stereo images is easy, obtaining their corresponding ground-truth depth maps and/or pixel-wise correspondences is quite challenging. As such, techniques that require minimal or no 3D supervision are usually preferred  over those that require ground-truth depth maps during training. 
	
	\item \textit{End-to-end training. } In general, the depth estimation pipeline is composed of multiple blocks. In methods, these  blocks are trained separately. Others train them jointly in an end-to-end fashion. Some  these techniques include a deep learning-based  refinement module. Others directly regress  the final high resolution map without additional post-processing or regularization. 

	\item \textit{Sub-pixel accuracy. }   In general, its is desirable to achieve  sub-pixel accuracy without any additional post-processing or regularization. 

	\item \textit{Change in  disparity range. } This may require changing the network structure as well as re-training. 
\end{itemize}


\noi We will use these metrics and criteria to compare and discuss existing methods.

\subsection{Pairwise stereo matching techniques}

Table~\ref{tab:performance_stereomatching_ktti15} compares the properties and performance of  deep learning-based depth estimation methods from stereo images. Below, we discuss some of them.

%
   
\begin{table*}

\caption{\label{tab:performance_stereomatching_ktti15}Performance comparison of deep learning-based stereo matching algorithms on the test set of KITTI 2015 benchmark (as of 2019/01/05). PE: percentage of erroneous pixels. EPE: End-Point Error. S1: the bad pixel error. Non-Occ: non-occluded pixels. All: all pixels. fg: foreground pixels. bg: background pixels. Noc: non-occluded pixels only. The bad pixel metric (D1) considers the disparity/depth at a pixel to be correctly estimated if the error is less than $3$ pixels  or less than $5\%$ of its value.}

\resizebox{\textwidth}{!}
{%

	\begin{tabular}{|@{ } >{\raggedright}p{1.8cm}  @{ }|
		@{ }  >{\raggedright}p{3.5cm} @{ } |	
		@{ } c@{ } | @{ }  >{\raggedright}p{3.2cm}   @{ } | @{ } c@{ } | 
		@{ }c@{ }|@{ }c@{ }|@{ }c@{ }|@{ } c@{ }|@{ }c@{ }|@{ }c@{ } |
		@{ }c@{ }|@{ }c@{ }|@{ }c@{ }|@{ } c@{ }|@{ }c@{ }|@{ }c@{ } |		
		@{ } c @{ }|
		@{ }  >{\raggedright}p{2.5cm} @{ }|}
		
		\hline
		\multirow{2}{*}{\textbf{Method}} &   
		
		\textbf{Description} &
		\multicolumn{3}{c|}{\textbf{Training}} & 
		
		\multicolumn{3}{c|}{\textbf{Avg D1 Non-Occ / Est}}  &
		\multicolumn{3}{c|}{\textbf{Avg D1 All / Est}}  & 
				
		\multicolumn{3}{c|}{\textbf{Avg D1 Non-Occ / All}}  &
		\multicolumn{3}{c|}{\textbf{Avg D1 All / All}}  & 
		\multirow{2}{*}{\textbf{Time (s)}} & \multirow{2}{*}{\textbf{Environment}}
	\\
		
		\cline{3-17}
			&
			&
			\textbf{3D Sup} & \textbf{Loss} & \textbf{End-to-end} &
			
			\textbf{D1-fg} & \textbf{D1-bg} & \textbf{D1-all} &
			\textbf{D1-fg} & \textbf{D1-bg} & \textbf{D1-all} &

			\textbf{D1-fg} & \textbf{D1-bg} & \textbf{D1-all} &
			\textbf{D1-fg} & \textbf{D1-bg} & \textbf{D1-all} & 
			&			
			\\
		\hline
%

		
		MC-CNN Accr~\cite{zbontar2015computing}&  
			Raw disparity $+$ classic refinement  &
			\checkmark  &  $\lossltwomatching$ &   \xmark& 	
			$7.64$ & $2.48$ & $3.33$ &
			$8.88$ & $2.89$ & $3.89$ &   
			$7.64$ & $2.48$ & $3.33$ &
			$8.88$ & $2.89$ & $3.89$ &  
			$67$&
			Nvidia GTX Titan X (CUDA, Lua/Torch7)\\
			
		\hline	
		Luo \etal~\cite{luo2016efficient} & 
			Raw disparity $+$ classic refinement  &
			\checkmark & $\losscrossentropy$ & \xmark&
			$7.44$ & $3.32$ & $4.00$ &
			$8.58$ & $3.73$ & $4.54$ &    
			$7.44$ & $3.32$ & $4.00$ &
			$8.58$ & $3.73$ & $4.54$ &  
			$1$&
			Nvidia GTX Titan X (Torch)\\
			
		\hline	
		Chen \etal~\cite{chen2015deep} &  
			Raw disparity $+$ classic refinement  &
			\checkmark  &  $\lossltwomatching $ &   \xmark&
			$-$ & $-$ & $-$ &
			$-$ & $-$ & $-$ &  
			$-$ & $-$ & $-$ &
			$-$ & $-$ & $-$ &  
			$-$&
			\\
	
		
		\hline	
		L-ResMatch~\cite{shaked2017improved}  &  
			Raw disparity $+$ confidence score $+$ classic refinement &
			\checkmark  &   & $\circ$ &		%
			$5.74$ & $2.35$ & $2.91$ &
			$6.95$ & $2.72$ & $3.42$ &  
			$5.74$ & $2.35$ & $2.91$ &
			$6.95$ & $2.72$ & $3.42$ &  
			$48$&
			Nvidia Titan-X\\
		
		
		\hline	
		Han \etal \cite{han2015matchnet}  &  
			Matching network&
			\checkmark & $\losscrossentropy $ & \xmark &
			$-$ & $-$ & $-$ &
			$-$ & $-$ & $-$ &  
			$-$ & $-$ & $-$ &
			$-$ & $-$ & $-$ &  
			$-$&
			Nvidia GTX Titan Xp\\		
		\hline
		Tulyakov \etal \cite{tulyakov2017weakly}&  
			 MC-CNN fast~\cite{zbontar2015computing} + weakly-supervised learning &
			 & $-$   & \xmark & 
			$9.42$ & $3.06$ & $4.11$ &
			$10.93$ & $3.78$ & $4.93$ &   
			$9.42$ & $3.06$ & $4.11$ &
			$10.93$ & $3.78$ & $4.93$ &  
			$1.35$&
			1 core 2.5 Ghz + K40 NVIDIA, Lua-Torch\\

		\hline	
		FlowNetCorr \cite{dosovitskiy2015flownet} &  
			Refined disparity&
			\checkmark &  $ \lossabsdisparitydiff$ & \checkmark &
			$-$ & $-$ & $-$ &
			$-$ & $-$ & $-$ &  
			$-$ & $-$ & $-$ &
			$-$ & $-$ & $-$ &  
			$1.12$&
			Nvidia GTX Titan\\
			
		\hline	
		DispNetCorr \cite{mayer2016large} &  
			Raw disparity &
			\checkmark&  & $\circ$&
			$3.72$ & $4.11$ & $4.05$ &
			$4.41$ & $4.32$ & $4.34$ &   
			$3.72$ & $4.11$ & $4.05$ &
			$4.41$ & $4.32$ & $4.34$ &  
			$0.06$&
			Nvidia Titan-X\\
			
		\hline	
		Pang \etal \cite{pang2017cascade} &  
			Raw disparity&
			\checkmark&  $\lossabsdisparitydiff$& $\circ$&
			$3.12$ & $2.32$ & $2.45$ &
			$3.59$ & $2.48$ & $2.67$ &   
			$3.12$ & $2.32$ & $2.45$ &
			$3.59$ & $2.48$ & $2.67$ &  
			$0.47$&
			$-$\\	
		
		\hline
		Yu \etal \cite{yu2018deep}& 
			Raw disparity map&
			\checkmark & $\lossabsdisparitydiff$ & $\circ$&
			$-$ & $-$ & $-$ &
			$-$ & $-$ & $-$ &  
			$5.32$ & $2.06$ & $2.32$ &
			$5.46$ & $2.17$ & $2.79$ &  
			$1.13$&
			Nvidia 1080Ti\\
			
		\hline	
		Yang \etal~\cite{yang2018segstereo} - supp.&  
			Raw disparity&
			\checkmark & $\lossabsdisparitydiff + \lossltwomatching + \losslonegradient $ & \checkmark &
			$3.70$ & $1.76$ & $2.08$ &
			$4.07$ & $1.88$ & $2.25$ &   
			$3.70$ & $1.76$ & $2.08$ &
			$4.07$ & $1.88$ & $2.25$ &  
			$0.6$ &
			\\
		
		\hline	
		Yang \etal~\cite{yang2018segstereo} - unsup. &  
			Raw disparity &
			\xmark&  $\inversewarploss + \lossltwomatching +  \losslonegradient $  & \checkmark &
			$-$ & $-$ & $-$ &
			$-$ & $-$ & $-$ &  
			$-$ & $-$ & $7.70$ &
			$-$ & $-$ & $8.79$ &  
			$0.6$ &
			\\		
			
		\hline	
		Liang \etal~\cite{liang2018learning}  &  
			Refined disparity&
			\checkmark & $\lossabsdisparitydiff$ & \checkmark &
			$-$ & $-$ & $-$ &
			$-$ & $-$ & $-$ &  
			$2.76$ & $2.07$ & $2.19$ &
			$3.40$ & $2.25$ & $2.44$ &  
			$0.12$&
			Nvidia Titan-X\\
			
		\hline	
		Khamis \etal~\cite{khamis2018stereonet}  & 
			Raw disparity $+$ hierarchical refinement&
			\checkmark &$\lossapproximatesmoothlone$ with $\alpha =1, c = 2$  & \checkmark&
			$-$ & $-$ & $-$ &
			$-$ & $-$ & $-$ &  
			$-$ & $-$ & $-$ &
			$7.45$ & $4.30$ & $4.83$ &  
			$0.015$&
			Nvidia Titan-X\\
			

		\hline		
		Gidaris \&  Komodakis \cite{gidaris2017detect}&  
			Refinement only&
			\checkmark & $\lossabsdisparitydiff $ & $\circ$&
			$4.87$ & $2.34$ & $2.76$ &
			$6.04$ & $2.58$ & $3.16$ &   
			$4.87$ & $2.34$ & $2.76$ &
			$6.04$ & $2.58$ & $3.16$ &  
			$0.4$&
			Nvidia Titan-X\\
			
		\hline	
		Chang \& Chen \cite{chang2018pyramid}   &  
			Raw disparity&
			\checkmark &  $ \losssmoothlone$& $\circ$&
			$4.31$ & $1.71$ & $2.14$ &
			$4.62$ & $1.86$ & $2.32$ &  
			$4.31$ & $1.71$ & $2.14$ &
			$4.62$ & $1.86$ & $2.32$ &  
			$0.41$&
			Nvidia GTX Titan Xp\\
			
%
%

		\hline	
		Zhong \etal \cite{zhong2017self} &  
			Raw disparity map&
			\xmark & $\alpha_1\loss_{1}^{12} + \alpha_2\inversewarploss + \alpha_3\losssubpixelcrossentropy+ \alpha_4\lossweightsecondgradient + \alpha_5\lossloopconsistency + \alpha_6 \lossmdh$ &  $\circ$ &
			$6.13$ & $2.46$ & $3.06$ &
			$7.12$ & $2.86$ & $3.57$ &   
			$6.13$ & $2.46$ & $3.06$ &
			$7.12$ & $2.86$ & $3.57$ &  
			$0.8$&
			P100\\
		\hline		
		
		Kendall \etal \cite{kendall2017end}  &  
			Refiend disparity map without refinement module&
			\checkmark & $\lossabsdisparitydiff$ &  \checkmark &
			$5.58$ & $2.02$ & $2.61$ &
			$6.16$ & $2.21$ & $2.87$ &  
			$5.58$ & $2.02$ & $2.61$ &
			$6.16$ & $2.21$ & $2.87$ &  
			$0.9$&
			Nvidia GTX Titan X\\
		
		\hline
			
		Standard SGM-Net \cite{seki2017sgm}  &  
			Refinement with CNN-based SGM&
			\checkmark & Weighted sum of path cost and neighbor cost  &  $\circ$ &
			$-$ & $-$ & $-$ &
			$-$ & $-$ & $-$ &  
			$7.44$ & $2.23$ & $3.09$ &
			$-$ & $-$ & $-$ &  
			$67$&
			Nvidia Titan-X\\
			
		\hline
		Signed SGM-Net \cite{seki2017sgm}  &  
			Refinement with CNN-based SGM&
			\checkmark& Weighted sum of path cost and neighbor cost & $\circ$&
			$7.43$ & $2.23$ & $3.09$ &	
			$8.64$ & $2.66$ & $3.66$ &   
			$7.43$ & $2.23$ & $3.09$ &	
			$8.64$ & $2.66$ & $3.66$ &    %
			$67$&
			Nvidia Titan-X\\
		
		
%

%
%
%
%
%
%
%
		
		\hline
		
		Cheng \etal \cite{cheng2018learning}&  
			Refinement&
			\checkmark & $\lossabsdisparitydiff$ & $\circ$ &
			$2.67$ & $1.40$ & $1.61$ &
			$2.88$ & $1.51$ & $1.74$ &  
			$2.67$ & $1.40$ & $1.61$ &
			$2.88$ & $1.51$ & $1.74$ &  
			$0.5$&
			GPU @ 2.5 Ghz (C/C++)\\
			
		\hline
		EdgeStereo \cite{song2018stereo}&  
			Raw disparity &
			\checkmark& $\displaystyle\sum_{sc=1}^{\text{nscales}} \left(\lossabsdisparitydiff + \alpha \lossedgesemantic\right)_{sc}$ & \xmark&
			$3.04$ & $1.70$ & $1.92$ &
			$3.39$ & $1.85$ & $2.10$ &  
			$3.04$ & $1.70$ & $1.92$ &
			$3.39$ & $1.85$ & $2.10$ &  
			$0.32$&
			Nvidia GTX Titan Xp\\
		\hline
		Tulyakov \etal \cite{tulyakov2018practical}&  
			Disparity with sub-pixel accuracy&
			\checkmark & $\losssubpixelcrossentropy$   & $\circ$ & 
			$3.63$ & $2.09$ & $2.36$ &
			$4.05$ & $2.25$ & $2.58$ &  
			$3.63$ & $2.09$ & $2.36$ &
			$4.05$ & $2.25$ & $2.58$ &  
			$0.5$&
			1 core @ 2.5 Ghz (Python)\\


		\hline
		Jie \etal \cite{jie2018left}&  
			Refined disparity with DL&
			\checkmark & $\lossabsdisparitydiff$ & \checkmark&
			$4.19$ & $2.23$ & $2.55$ &
			$5.42$ & $2.55$ & $3.03$ &  
			$4.19$ & $2.23$ & $2.55$ &
			$5.42$ & $2.55$ & $3.03$ &  
			$49.2$&
			Nvidia GTX Titan X\\



		\hline
		Seki \etal \cite{seki2016patch}&  
			Raw disparity, confidence map, SGM-based refinement &
			\checkmark & $\losscrossentropy$ & $\circ$ &
			$7.71$ & $2.27$ & $3.17$ &  
			$8.74$ & $2.58$ & $3.61$ & 
			$7.71$ & $2.27$ & $3.17$ &  
			$8.74$ & $2.58$ & $3.61$ &
			$68$&
			Nvidia GTX Titan X\\

		\hline
		Kuzmin \etal \cite{kuzmin2017end}&  
			Only aggregated cost volume&
			\checkmark&  $\losscrossentropy $  & $\circ$ &
			$10.11$ & $4.81$ & $5.68$ &
			$11.35$ & $5.32$ & $6.32$ &  
			$10.11$ & $4.82$ & $5.69$ &
			$11.35$ & $5.34$ & $6.34$ &  
			$0.03$&
			GPU @ 2.5 Ghz (C/C++)\\

		\hline
		Tonioni \etal \cite{tonioni2017unsupervised} &  
			Unsupervised adaptation - DispNetCorr1D~\cite{mayer2016large} + CENSUS~\cite{zabih1994non}&
			\checkmark&  $ \confidenceguidedloss + \alpha \lossavgbhblone$ & NA &
			$-$ & $-$ & $-$ &
			$-$ & $-$ & $-$ &  
			$-$ & $-$ & $-$ &
			$-$ & $-$ & $0.76$ &  
			$-$&
			GPU @ 2.5 Ghz (Python)\\
			

		\hline                     
	\end{tabular}
}
\end{table*}

\subsubsection{Degree of supervision}

Most of the state-of-the-art  methods require ground-truth depth maps to train their deep learning models.  This is reflected in the loss functions they use to train the networks. For instance,  Flynn \etal~\cite{flynn2016deepstereo}, 
Kendall \etal~\cite{kendall2017end},  Pang \etal~\cite{pang2017cascade}, Cheng \etal~\cite{cheng2018learning},  and Liang \etal~\cite{liang2018learning} minimize the $\lone$ distance between the estimated disparity/depth and the ground truth (Equation~\eqref{eq:loss_abs_disparity_diff}), while Luo \etal~\cite{luo2016efficient} minimize the cross-entropy loss of Equation~\eqref{eq:cross_entropy_loss}.  Khamis \etal~\cite{khamis2018stereonet} used the same approach but by using the two-parameter robust function (Equation~\eqref{eq:loss_two_params_robust}).  Chen \etal~\cite{chen2015deep}, which formulated the stereo matching problem as a classification problem, trained  their network  to classify whether pixel $\pixel$ on the left image and pixel $\pixel -\disparity$ on the right image are in correspondence (positive class) or not (negative class). The loss is then defined as the $\ltwo$ distance between the output of the network for the pixel pair $(\pixel , \pixel - \disparity)$ and the ground-truth label (0 or 1) of this pair of pixels.

In general, obtaining ground-truth disparity/depth maps is very challenging. As such, techniques that do not require 3D supervision are more attractive.  The key to training without 3D supervision is the use of loss functions that are based on the reprojection error, \eg Equation~\eqref{eq:photometric_loss}. 
  This approach has been adopted in recent techniques, \eg Zhou \etal~\cite{zhou2017unsupervised} and Yang \etal~\cite{yang2018segstereo}. One limitation of these techniques is that they assume that the camera parameters are known so that the unwarping or re-projection onto the coordinates of the other image can be calculated. Some techniques, \eg\cite{ummenhofer2017demon}, assume that the camera parameters are unknown and regress them at the same time as depth/disparity in the same spirit as Structure from Motion (SfM) or visual SLAM.

As shown in Table~\ref{tab:performance_stereomatching_ktti15}, methods that are trained with 3D supervision achieve a better performance at runtime than those without 3D supervision. For example, Yang \etal~\cite{yang2018segstereo}  evaluated their networks in both modes and showed that the supervised one achieved and average D1-all (All/All) of $2.25\%$ compared to $8.79\%$ for the unsupervised one.

\subsubsection{Accuracy and disparity range} 
 
Based on the metrics of Section~\ref{sec:performance_metrics}, the unsupervised adaptation method of Tonioni \etal~\cite{tonioni2017unsupervised}  boosted significantly the performance  of DispNetCorr1D~\cite{mayer2016large} (from $4.34$ on Avg D1 All/All (D1-all) to $0.76$). This suggests that such adaptation module can be used to boost the performance of the other methods.   

Note that only a few methods could achieve sub-pixel accuracy. Examples include the approach of Tulyakov \etal~\cite{tulyakov2018practical}, which uses the sub-pixel MAP approximation instead of the softargmin. Also,  Tulyakov \etal~\cite{tulyakov2018practical}'s approach    allows changing the disparity range at runtime without retraining the network. 

\subsubsection{Computation time and memory footprint}

Computation time and memory footprint, which in general are interrelated, are very important especially at runtime. Based on Table~\ref{tab:performance_stereomatching_ktti15}, we can distinguish three types of methods; Slow methods, average-speed methods, which produce a depth map in around one seconds, and fast methods, which require less than $0.1$ seconds to estimate a single depth map. 

Slow methods require  more than $40$ seconds to estimate one single depth map. There are multiple design aspects that make these method slow. For instance, some of them perform multiple forward passes as in~\cite{seki2017sgm}. Others either deepen the network by using a large number of layers including many fully connected layers, especially in the similarity computation block as in the MC-CNN Acc of~\cite{zbontar2015computing} and the L-ResMatch of~\cite{shaked2017improved}),  or use multiple subnetworks. Other methods estimate the depth map in a recurrent fashion, \eg by using multiple convLSTM blocks as in Jie \etal~\cite{jie2018left}. 

According to Table~\ref{tab:performance_stereomatching_ktti15}, the approach of Khamis \etal~\cite{khamis2018stereonet} is the fastest one as it produces, at run time, a disparity map in $15$ms, with a subpixel accuracy of $0.03$, which corresponds to an error of less than $3$cm at $3$m distance from the camera. In fact, Khamis \etal~\cite{khamis2018stereonet} observed that most of the time and compute is spent matching features at higher resolutions, while most of the performance gain comes from matching at lower resolutions. Thus, they compute the cost volume  by matching features at low resolution. An efficient refinement module is then used to  upsample the low resolution depth map to the input resolution.

Finally, t Mayer \etal~\cite{mayer2016large} and Kendall \etal~\cite{kendall2017end}   can run very fast, with $0.06$s and $0.9$s consumed on a single Nvidia GTX Titan X GPU, respectively. However, disparity refinement is not included in these networks, which limits their performance.

\subsection{Multiview  stereo  techniques}

Table~\ref{tab:performance_mvs_ktti15} compares the properties and performance of five deep learning-based  multiview stereo reconstruction algorithms. Note that the related papers have reported performance results using different datasets.

\subsubsection{Degree of supervision}

Most of the methods described in Table~\ref{tab:performance_mvs_ktti15} are trained using 3D supervision. The only exception is the approach of Flynn \etal~\cite{flynn2016deepstereo}, which is trained using a posed set of images, \ie images with known camera parameters. At training, the approach takes a set of images, leaves one image out, and learns how to predict it from the remaining ones. The rational is that providing a set of posed images is much simpler than providing depth values at each pixel in every reference image. 

\subsubsection{Accuracy and depth range}
In terms of accuracy, the approach of Huang \etal~\cite{huang2018deepmvs} seems to outperform the state-of-the art, see Table~\ref{tab:performance_mvs_ktti15}. However,  since these methods have been evaluated on different datasets, it is not clear whether they would achieve the same level of accuracy on other datasets. As such, the accuracy results reported in Table~\ref{tab:performance_mvs_ktti15} are just indicative not conclusive.

Finally, since most of the MVS methods rely on Plane Sweep Volumes or image/feature umprojection onto depth planes, the depth range needs to be set in advance. Changing the depth range and its discretization at runtime would require re-training the methods.

\begin{table*}[t]
\caption{\label{tab:performance_mvs_ktti15} Performance comparison of deep learning-based  multiview stereo matching algorithms on the KITTI 2015 benchmark. Accuracy and completeness refer, respectively, to the mean accuracy and mean completeness (the lower the better).  }
\resizebox{\textwidth}{!}
{%
	\begin{tabular}{
	       |@{ }  >{\raggedright}p{1cm} @{ }|
		@{ }  >{\raggedright}p{3.5cm} @{ } |	
		@{ }  >{\raggedright}p{1cm} @{ } |   
		@{ } c@{ } | @{ }  c  @{ } | @{ } >{\raggedright}p{2cm}  @{ } | c@{ } | c@{ } | c@{ } | 
		@{ } c@{ } | @{ } >{\raggedright}p{2.2cm} @{ } | @{ } c@{ } |
		@{ }c@{ }|@{ }c@{ }|@{ }c@{ }|
		}
		\hline
		\multirow{2}{*}{\textbf{Method}} &   
		\textbf{Description} &
		\multirow{2}{1cm}{\textbf{Depth range}} &  
		\multicolumn{6}{c|}{\textbf{Training}} & 		
		\multicolumn{3}{c|}{\textbf{Testing}} & 	
		\multicolumn{3}{c|}{\textbf{Performance}}  \\

		\cline{4-15}
			&
			&
			&
			\textbf{\#views}  & \textbf{3D Sup} & \textbf{Loss} & \textbf{End-to-end}  & \textbf{Time (s)} & \textbf{Memory} & 
			\textbf{\#views} & \textbf{Time (s) } & \textbf{Memory } & 							      				
			\textbf{Dataset} & \textbf{Accuracy} & \textbf{Completeness} \\										
			
		\hline
		
			\cite{flynn2016deepstereo}& Feature projection and unprojection. 
			&
			$96$&	
			$4+1$&  \xmark &  $\lone$ color loss &  \checkmark &  $-$ & $-$ &	
			$4$ & $12$ min & $-$ &			
			KITTI 2012 & $-$ & $-$ \\
			\hline
			
			\cite{hartmann2017learned}&  
			Multi-patch similarity&
			$256$&	
			variable&  \checkmark &  softmax loss &  $\circ$ &  $-$ & $-$ &	
			variable & $0.07$  & $-$ &									
			DTU & $1.336$ & $2.126$ 	\\
			\hline
			
			\cite{kar2017learning}&  
			Feature unprojection. Volumetric reconstruction followed by projection to generate depth &
			$300$&	
			$5$ or $10$& \checkmark  &  $\lone$ loss&  \checkmark &  $-$ & $-$ &	
			$5$ or $10$ & $0.033$  & $-$ &									
			ShapeNet & $-$ & $-$ \\
			\hline
			
			\cite{huang2018deepmvs}&  
			Plane Sweep Volumes &
			$100$&	
			$6$&  \checkmark &  class-balanced cross entropy &  \checkmark &  $5$ days & $-$ &	
			variable & $0.05$ for $32^3$ grid,  $0.40$  for $64^3$ grid & $-$ &									
			ETH3D & $0.036$ & $-$ \\
			\hline
			
			\cite{yao2018mvsnet}&  
			Feature unprojection&
			$256$&	
			$3$&  \checkmark &  $\lone$ & \checkmark  &  $-$ & $-$ &	
			$5$ & $4.7$ per view & $-$ &									
			DTU & $0.396$ & $0.527$ \\
		\hline                              
	\end{tabular}
}
\end{table*}

\subsection{Depth regression techniques}

\begin{table*}[t]
\caption{\label{tab:performance_depthregression} Comparison of some deep learning-based depth regression techniques.  }
\resizebox{\textwidth}{!}
{%
	\begin{tabular}{|@{ } >{\raggedright}p{2cm} @{ }|@{ }>{\raggedright}p{4cm} @{ }| @{ }>{\raggedright}p{1.2cm} @{ }|@{ }>{\raggedright}p{1.2cm} @{ } | @{ }>{\raggedright}p{1.2cm} @{ }|@{ }c@{ } |@{ }c@{ } | @{ }c @{ }|@{ }c@{ } | @{ }c @{ }| @{ }c@{ } |@{ }c @{ }|@{ }c@{ } |@{ }c @{ }|@{ }c @{ } |@{ }c @{ }|@{ }c @{ }|@{ }c @{ }|@{ }c@{ } |@{ }c @{ }|@{ }c @{ }|@{ }c @{ }|@{ }c @{ }|@{ }c@{ }|}
		\hline
		\multirow{3}{*}{\textbf{Method}} &   
		\multirow{3}{*}{\textbf{Description}	}		&
		\multicolumn{2}{c|}{\textbf{Training}	}		&
		\multicolumn{2}{c|}{\textbf{Runtime}	}		&
		\multicolumn{7}{c|}{\textbf{KITTI 2012}}				&
		\multicolumn{4}{c|}{\textbf{Make3D}}				&
		\multicolumn{7}{c|}{\textbf{NYUDv2}}				\\
		
		\cline{3-24}
		& & 
		\textbf{\#views} & \textbf{3D sup} &	
		\textbf{\#views}  & \textbf{time}   &		
		\multicolumn{4}{c|}{\textbf{Performance ($\downarrow$ the better)}}	&  \multicolumn{3}{c|}{\textbf{Accuracy ($\uparrow$ the better) }}		& 
		\multicolumn{4}{c|}{\textbf{Performance ($\downarrow$ the better)}} 	&  
		\multicolumn{4}{c|}{\textbf{Performance ($\downarrow$ the better)}}	&  \multicolumn{3}{c|}{\textbf{Accuracy ($\uparrow$  the better)}}		\\
		\cline{7 -24}
		& & 
		& & 
		& & 
		abs rel. & sqr rel. & RMSE& RMSE  &  $< 1.25$  & $< 1.25^2$ & $<1.25^ 3$	 &
		abs rel. & sqr rel. & RMSE & RMSE  &  
		abs rel. & sqr rel. & RMSE & RMSE&   $< 1.25$  & $< 1.25^2$ & $< 1.25^ 3$	 \\
		
		& &   & & &   & & &   (lin) &  (log) &    &   &  	 &
		     &   &  (lin) &  (log) &    
		     &   &  (lin) &  (log) &       &   &  	 \\
			 
		\hline
		Chakrabarti \etal~\cite{chakrabarti2016depth} &	Probability that model confidence and ambiguities&
							$1$&  \checkmark  & 
							$1$ & $24$ & 
							 $-$ & $-$ &$-$ &$-$ & $-$ & $-$ & $-$     &
							 $-$ & $-$ & $-$ & $-$ &
							$0.149$ & $0.118$   &  $0.620$ &  $0.205$     & $0.806$ & $0.958$ & $0.987$  \\
		
		\hline 
		Kuznietsov \etal~\cite{kuznietsov2017semi} & Supervised followed by unsupervised 	&
							$2$ & 3D + Stereo & 
							$1$ &  $-$ & 
							$0.113$ & $0.741$ & $4.621$ & $0.189$  &  0.862 & 0.960&  0.986 & 
							$0.157$ & $-$ & $3.97$ & $0.062$ & 
							$-$ & $-$ &$-$ &$-$ & $-$ &  $-$ & $-$  \\
							
		\hline
		Zhan \etal~\cite{Zhan_2018_CVPR}  &  visual odometry	&
							Stereo seq. + cam. motion&\xmark & 
							$1$ &  & 
							$0.144$ & 1.391 &5.869 &0.241 &  0.803 &0.928 & 0.969 &
							 $-$ & $-$ & $-$ & $-$ &
							$-$ & $-$ &$-$ &$-$ & $-$ & $-$ & $-$   \\
		
		\hline			
		Eigen~\cite{eigen2014depth} 	&Multi-scale &
							$1$& \checkmark& 
							$1$ & $-$& 
							$0.1904$ & $1.515$&  $7.156$&  $0.270$    & 0.702 & 0.890 &0.958& 
						  	$-$ & $-$ & $-$ & $-$ &
							$0.214$ & $0.204$ &$0.877$ &$0.283$ & $0.614$ & $0.888$ & $0.972$  \\
		\hline
		Eigen \etal~\cite{eigen2015predicting}  &  VGG - multi-sclae CNN for depth, normals, and  labeling&
							$1$& \checkmark & 
							$1$ & $0.033$& 
							$ $ &  $ $  & $ $ & $ $  & $-$ & $ $   & $ $ &
							 $-$ & $-$ & $-$ & $-$ &
							$0.158$ & $0.121$ &$0.641$ &$0.214$   & $0.769$ & $0.950$& $0.988$ \\

		\hline 
		Fu \etal~\cite{Fu_2018_CVPR}   & ResNet - Ordinal regression&
							$1$ & \checkmark& 
							$1$ & $-$ & 
							$0.072 $ &  $0.307$  & $2.727$ & $0.120$    & $ 0.932 $   & $0.984$ &$ 0.994$   &
							 $0.157$ & $-$ & $3.97$ & $0.062$ & 
							$0.115$ & $-$ &$0.509$ &$0.051$ &  $-$ & $-$ & $-$  \\
		\hline  
		 Gan \etal~\cite{Gan_2018_ECCV} &Uses Affinity, Vertical Pooling, and Label Enhancement&
							$1$& \checkmark& 
							$1$ & $0.07$ & 
							 $0.098$ &  $0.666$  & $ 3.933$ & $0.173$  &   $0.890$   & $ 0.964 $ &$ 0.985$   &
							 $-$ & $-$ & $-$ & $-$ &
							$0.158$ & $-$ &$0.631$ &$0.066 $ &   $0.756$ & $ 0.934$ & $0.980$  \\
							
		\hline 
		 Garg  \cite{garg2016unsupervised} &  variable-size input &
							$2$& \xmark  & 
							$1$ & $-$& 
							$0.177$ &  $1.169 $  & $5.285$ & $0.282$      & $0.727$ & $ 0.896 $   & $ 0.958 $ &
							 $-$ & $-$ & $-$ & $-$ &
							$-$ & $-$ &$-$ &$-$ & $-$ & $-$ & $-$   \\
		\hline					
		Godard \cite{godard2017unsupervised} & Training with a stereo pair&
							$2$ (calibrated)& \xmark & 
							$1$ & $-$ & 
							$0.114$ & $0.898$ & $4.935$ & $0.206$ &  0.830 &0.936 & 0.970& 
							$0.535$ & $11.990$ & $11.513$ & $0.156$ & 
							$-$ & $-$ &$-$ &$-$ & $-$ & $-$ & $-$    \\

		\hline 
		 Jiao \cite{Jiao_2018_ECCV} & $40$ categories - pay more attention to distant regions &
							$1$&  \checkmark& 
							$1$ & $-$ & 
							$-$ & $-$ &$-$ &$-$ & $-$   & $-$ & $-$ &
							 $-$ & $-$ & $-$ & $-$ & 
							 $0.098 $ &  $ $  & $0.329$ & $ 0.125$  &   $ 0.917 $   & $ 0.983 $ &$0.996$    \\

		\hline
		 Laina \cite{laina2016deeper}  & VGG - feature map up-sampling &
							$1$ & \checkmark  & 
							$1$ & $0.055$ & 
							$ $ &  $ $  & $ $ & $ $  & $-$ & $ $   & $ $  &
							 $0.176$ & $-$ & $4.6$ & $0.072$ &  
							$0.194$ & $-$ &$0.79$ &$0.083$ &   $-$ & $-$ & $-$  \\							
		\hline
		 Laina \cite{laina2016deeper}  & ResNet - feature map up-sampling &
							$1$ & \checkmark   & 
							$1$ & $0.055$&  
		 					 $ $ &  $ $  & $ $ & $ $  & $-$     & $ $ &$ $   &
							 $ $ & $-$ & $ $ & $ $ & 
							$0.127$ & $-$ &$0.573$ &$0.055$ &   $0.811$ & $ 0.953 $ & $0.988$  \\
		\hline  
		Lee  \cite{Lee_2018_CVPR} &Split and merge &
							$1$& \checkmark & 
							$1$ & $-$ & 
							 $ $ &  $ $  & $ $ & $ $  & $-$ & $ $   & $ $    &
							 $-$ & $-$ & $-$ & $-$ & 
							$0.139$ & $ 0.096$ &$0.572$ &$ 0.193 $  & $0.815$ & $ 0.963$ & $0.991$  \\
							
		\hline
		Li \cite{li2015depth} &  Multiscale patches, refinement with CRF&
							$1$&  \checkmark & 
							$1$& $-$  & 
							$ $ &  $ $  & $ $ & $ $  & $-$ & $ $   & $ $ & 
							 $0.278$ & $-$ & $7.188$ & $-$ &  
							$0.232$ & $-$ &$0.821$ &$-$ &   $0.6395$ & $0.9003$ & $0.9741$ \\ 

		\hline  
		 Qi \cite{Qi_2018_CVPR} & Joint depth and normal maps &
							$1$&  \checkmark& 
							$1$ & $0.87$& 
							$-$ & $-$ &$-$ &$-$ & $-$ & $-$ & $-$ &  
							 $-$ & $-$ & $-$ & $-$ & 
							$ 0.128$ &  $ $  & $0.569$ & $ $  &   $0.834 $   & $ 0.960$ &$0.990$     \\
		
		\hline  
		Roy  \cite{roy2016monocular} &  CNN + Random Forests&
							$1$&  \checkmark& 
							$1$ & $-$& 
							$ $ &  $ $  & $ $ & $ $  & $-$ & $ $   & $ $ & 
							 $0.260$ & $-$ & $12.40$ & $0.119$ & 
							$0.187$ & $-$ &$0.74$ &$-$ &   $-$ & $-$ & $-$  \\
		

		\hline 
		 Xian \cite{Xian_2018_CVPR} & Training with $3$K ordinal relations per image &
							$1$& $3$K ordinal  & 
							$1$ & $-$  & 
							$ $ &  $ $  & $ $ & $ $  & $-$ & $ $   & $-$     &
							  $-$ & $-$ & $-$ & $-$ & 
							$0.155$ & $-$ &$0.660$ &$0.066$ &  $0.781$ & $0.950$ & $0.987$  \\
							
		\hline
		Xu \cite{xu2017multi} &  Integration with continuous CRF&
							$1$& \checkmark& 
							$1$& & 
							$-$ & $-$ &$-$ &$-$ & $-$ & $-$ & $-$     &
							 $0.184$ & $-$ & $4.386$ & $0.065$ &  
							$0.121$ &  $ - $  & $0.586$ & $0.052$  &  $0.706$   & $0.925$ &$0.981$  \\
							
		\hline 
		 Xu \cite{Xu_2018_CVPR} &  Joint depth estimation and scene parsing &
							$1$& \checkmark  & 
							$1$ & $-$& 
							$ $ &  $ $  & $ $ & $ $  & $-$ & $ $   & $ $ & 
							 $-$ & $-$ & $-$ & $-$ & 
							$0.214$ & $-$ &$0.792$ &$-$ &   $0.643$ & $ 0.902 $ & $0.977$  \\

		\hline
		 Wang \cite{wang2015designing} & Depth and semantic prediction &
							$1$&  \checkmark & 
							$1$ & $-$ & 
							$ $ &  $ $  & $ $ & $ $  & $-$ & $ $   & $ $     &
							 $-$ & $-$ & $-$ & $-$ & 
							$0.220$ & $-$ &$0.745$ &$0.262$ &   $0.605$ & $0.890$ & $0.970$  \\
		\hline
		Zhang \etal~\cite{zhang2018deep} & Hierarchical guidance strategy for depth refinement & 
							$1$ &   \checkmark &
							$1$ &  $0.2$  &
							$0.136$ &  $-$  & $4.310$ & $-$  &   $0.833$   & $0.957$ &$ 0.987$     &
							$0.181$ & $-$ & $4.360$ & $-$ & 
							$0.134$ & $-$ &$0.540$ &$-$ & $0.830$ & $0.964$ & $0.992$ \\					
		\hline 
		 Zhang \etal~\cite{Zhang_2018_ECCV} & ResNet50 - Joint segmentation and depth estimation&
							$1$&  \checkmark& 
							$1$ & $0.2$ & 
							 $-$ &  $-$  & $-$ & $-$  & $-$ & $-$   & $-$ &
							 $0.156$ & $-$ & $0.510$ & $0.187$ &  
							$0.140$ & $-$ &$0.468$ &$-$ &   $0.815$ & $0.962$ & $0.992$  \\	
		
		\hline  
		Zou \cite{Zou_2018_ECCV} &  Joint depth and flow&
							$2$& \xmark   & 
							$2$ & $-$& 
							$0.150$ & $1.124$ & $5.507$ & $0.223$ &    $ 0.806 $ & $0.933$&  $ 0.973$ & 
							$0.331$ & $2.698-$ &$0.416$ &$6.89$ &
							$-$ & $-$ &$-$ &$-$ & $-$ & $-$ & $-$    \\	
		\hline  
		Zhou  \cite{zhou2017unsupervised} & Depth + pose& 
							$\ge 2$ & \xmark & 
							$1$ & $-$& 
							 $ 0.183  $ &  $1.595 $  & $ 6.709  $ & $ 0.270$  &  $ 0.734 $   & $  0.902 $ &$0.959 $   &
							 $0.383$ & $5.321$& $10.47$& $0.478$ & 
							 $-$ & $-$ &$-$ &$-$ & $-$ & $-$ & $-$    \\
		\hline
		Zhou \cite{zhou2016learning}& 
			3D-guided cycle consistency&
			$2$&  \xmark& 
			$2$ & $-$  & 
			$-$ &  $-$  & $-$ & $-$  & $-$ & $-$   & $-$ &
			$-$ & $-$ & $-$ & $-$ & 
			$-$ & $-$ &$-$ &$-$ & $-$ & $-$ & $-$ \\
			
		\hline
		Dosovitski \cite{dosovitskiy2015flownet}&  
			Regression from calibrated stereo images&
			$2$ (calibrated) &  \checkmark & 
			$2$ (calibrated) & $1.05$ &
			$-$ &  $-$  & $-$ & $-$  & $-$ & $-$   & $-$ &
			$-$ & $-$ & $-$ & $-$ & 
			$-$ & $-$ &$-$ &$-$ & $-$ & $-$ & $-$ \\
		
		\hline
		\cite{ummenhofer2017demon}&  
			Regression from a pair of images &
			$2$&  \checkmark&  
			$2$ & $0.11$ & 
			$-$ &  $-$  & $-$ & $-$  & $-$ & $-$   & $-$ &
			$-$ & $-$ & $-$ & $-$ & 
			$-$ & $-$ &$-$ &$-$ & $-$ & $-$ & $-$ \\
		
		\hline
		Pang \cite{pang2017cascade}&
			Cascade residual learning &
			$2$ (calibrated)&  \checkmark&
			$2$ & $-$ & 
			$-$ &  $-$  & $-$ & $-$  & $-$ & $-$   & $-$ &
			$-$ & $-$ & $-$ & $-$ & 
			$-$ & $-$ &$-$ &$-$ & $-$ & $-$ & $-$ \\
		\hline
		Ilg \cite{ilg2017flownet}  &  
			Extension of FlowNet~\cite{dosovitskiy2015flownet} &
			$2$&  \checkmark &  
			$2$ & $-$ &
			$-$ &  $-$  & $-$ & $-$  & $-$ & $-$   & $-$ &
			$-$ & $-$ & $-$ & $-$ & 
			$-$ & $-$ &$-$ &$-$ & $-$ & $-$ & $-$ \\
		\hline
		Li \cite{li2015depth} &  
			Depth from multiscale patches, refinement with CRF&
			$1$&  \checkmark & 
			$1$& $-$  & 
			$-$ &  $-$  & $-$ & $-$  & $-$ & $-$   & $-$ &
			$0.278$ & $-$ & $7.188$ & $-$ & 
			$0.232$ & $-$ &$0.821$ &$-$ & $-$ & $-$ & $-$ \\
		
		\hline
		Liu \cite{liu2016learning}  &  
			Refinement with continuous CRF&
			$1$&  \checkmark & 
			$1$& $-$ &
			$-$ &  $-$  & $-$ & $-$  & $-$ & $-$   & $-$ &
			$0.314$ & $-$ & $0.314$ & $-$ & 
			$0.230$ & $-$ &$0.824$  & $-$  &$ 0.614 $ & $0.883 $ & $0.971$\\	
			
		\hline
		Xie \cite{xie2016deep3d} &   
			Predict one view from another using estimated depth &
			$2$ stereo&  \xmark &
			$1$ & $-$ & 
			$-$ &  $-$  & $-$ & $-$  & $-$ & $-$   & $-$ &
			$-$ & $-$ & $-$ & $-$ & 
			$-$ & $-$ &$-$ &$-$ & $-$ & $-$ & $-$ \\	
		
		\hline
		Chen \cite{chen2016single}  &  
			Training with one ordinal relation per image&
			$1$& 1 ordinal &
			$1$ & $-$ &
			$-$ &  $-$  & $-$ & $-$  & $-$ & $-$   & $-$ &
			$-$ & $-$ & $-$ & $-$ & 
			$ 0.34 $ & $0.42$ &$ 1.10 $ &$0.38$ & $-$ & $-$ & $-$ \\
		
		\hline
		Mayer \cite{mayer2016large}&  
			A dataset for training&
			$1$ &  \checkmark  &
			$0.06$ &  $-$  &
			$-$ &  $-$  & $-$ & $-$  & $-$ & $-$   & $-$ &
			$-$ & $-$ & $-$ & $-$ & 
			$-$ & $-$ &$-$ &$-$ & $-$ & $-$ & $-$ \\
			
							
%
							
		\hline 
	\end{tabular}
}		
\end{table*}

Table~\ref{tab:performance_depthregression}  summarizes the properties and compares the performance of some of the state-of-the-art methods for deep learning-based depth regression  on KITTI2012, Make3D, and NYUDv2 datasets.  

As shown in this table, the methods of Jiao \etal~\cite{Jiao_2018_ECCV} and Fu \etal~\cite{Fu_2018_CVPR} seem to achieve the best accuracy on NYUDv2 dataset. Their common property is the way they handle different depth ranges. In fact, previous methods treat near and far depth values equally. As such, most of them achieve good accuracy in near depth values but their accuracy drops for distant depth values. Jiao \etal~\cite{Jiao_2018_ECCV} proposed a new loss function that pays more attention to distant depths. Fu \etal~\cite{Fu_2018_CVPR}, on the other hand, formulated depth estimation as a classification problem. The depth range is first discretized into intervals. The network then learn to classify each image pixel into one of the depth intervals. However, instead of using uniform discretization, Fu \etal~\cite{Fu_2018_CVPR}   used a spacing increasing discretization. 

Another observation from Table~\ref{tab:performance_depthregerssion} is that recent  techniques trained without 3D supervision, \eg by using stereo images  and re-projection loss, are becoming very competitive since their performances are close to techniques that use 3D supervision. Kuznietsov \etal~\cite{kuznietsov2017semi}  showed that the performance can be even further improved using semi-supervised techniques where the network is first trained with 3D supervision and then fine-tuned with stereo supervision.  Also, training with ordinal relations seems to improve the performance of depth estimation. In fact, while the early work of Chen \etal~\cite{chen2016single}, which used one ordinal relation per image, achieved relatively low performance, the recent work of Xian \etal~\cite{Xian_2018_CVPR}, which used $3$K ordinal relations per image, achieved an accuracy that is very close to supervised techniques. Their main benefit is that, in general metric depths obtained by stereo matching or depth sensor  are noisy. However, their corresponding ordinal depths are accurate. Thus, obtaining reliable ordinal depths for training is significantly easier. 

 Finally, methods  which jointly estimate depth and normals maps~\cite{Qi_2018_CVPR} or depth and semantic segmentation~\cite{Zhang_2018_ECCV} outperform many methods that estimate depth alone. Their performance can be further improved by using the loss function of~\cite{Jiao_2018_ECCV}, which pays more attention to distant depths, or by using the spacing increasing depth range discretization of~\cite{Fu_2018_CVPR}.

%
%
%
%
		
\section{Future research directions}
\label{sec:future}

Despite the extensive research undertaken in the past five years, deep learning-based depth reconstruction achieved promising results. The topic, however, is still in its infancy and further developments are yet to be expected. In this section, we present some of the current issues and highlight  directions for future research.    

\begin{itemize}
	\item \textit{Input: } Most of the current techniques do not handle high resolution input, require calibrated images, and cannot vary the number of input images at training and testing without re-training.  The former is mainly due to the computation and memory requirements of most of the deep learning techniques. Developments in high performance computing can address this issue in the future. However, developing lighter deep architectures remains desirable especially if to be deployed in mobile and portable platforms.

	\item \textit{Accuracy: } Although refinement modules can improve the resolution of the estimated depth maps, it is still small compared to the resolution of the images that can be recovered.  As such, deep learning techniques find it difficult to recover  small details, \eg vegetation and hair.   Also, most of the techniques discretize the depth range. Although some methods can achieve sub-pixel accuracy, changing the depth range, and the discretization frequency, requires retraining the networks. Another issue is  the accuracy, which, in general, varies for different depth ranges. Some of the recent works, \eg~\cite{Jiao_2018_ECCV}, tried to address this problem,  but it still remains an  open and  challenging problem since it is highly related to the data bias issue and the type of loss functions used to train the network. Accuracy of existing methods is also affected by complex scenarios, \eg occlusions and highly cluttered scenes, and objects with complex material properties. 

	\item \textit{Performance: } Complex deep networks are very expensive in terms of memory requirements. Memory footprint is even a major issue when dealing with high resolution images and when aiming to reconstruct high resolution depth maps.  While this can be mitigated by using multi-scale and part-based reconstruction techniques, it can result in high computation time.   

	\item \textit{Training: } Deep learning techniques rely heavily on the availability of training datasets annotated with ground-truth labels. Obtaining ground-truth labels for depth reconstruction is very expensive. Existing techniques mitigate this problem by either designing loss functions that do not require 3D annotations, or use domain adaptation and transfer learning strategies. The former, however, requires calibrated cameras.  Domain adaptation techniques are recently attracting more attention since, with these techniques, one can train with synthetic data, which are easy to obtain, and real-world data. 

	\item \textit{Data bias and generalization: } Most of the recent deep learning-based depth reconstruction techniques have been trained and tested on publicly available benchmarks. While this gives an indication on their performances, it is not clear yet how do they generalize and perform on completely unseen images, from a completely different category.  Thus, we expect in the future to see the emergence of large datasets, similar to ImageNet but for 3D reconstruction. Developing self-adaptation techniques, \ie techniques that can adapt themselves to new scenarios in real time  or with minimum supervision, is one promising direction for future research.

\end{itemize}

\section{Conclusion}
\label{sec:conclusion}

This paper provided a comprehensive survey of the recent developments in depth reconstruction using deep learning techniques.  Despite its infancy, these techniques are achieving acceptable results, and some recent developments are even competing, in terms of accuracy of the results, with traditional techniques.  We have seen that, since 2014, more than $100$ papers on the topic have been published in major computer vision and machine learning conferences and journals, and more new papers are being published even during the final stage of this submission. We believe that since 2014, we entered a new era where data-driven and machine learning techniques play a central role in image-based depth reconstruction. 

Finally, there are several related topics that have not been covered in this survey. Examples include:
\begin{itemize}
	\item Synthesis of novel 2D views from one or multiple images, which use similar formulations as depth estimation, see for example \cite{kulkarni2015deep,yang2015weakly,tatarchenko2016multi,zhou2016view,park2017transformation}.
	
	\item  Structure-from-Motion (SfM) and Simultaneous Localization and Mapping (SLAM), which aim to recover at the same time depth maps and  (relative) camera pose (or camera motion). Interested readers can refer to recent papers such as~\cite{mayer2016large,vijayanarasimhan2017sfm,wang2018recurrent}.

	\item Image-based object reconstruction algorithms, which aim to recover the entire 3D geometry of objects from one or multiple images. 
\end{itemize}

\noi While these topics are strongly related to depth estimation, they require a separate survey given the large amount of work that has been dedicated to them in the past 4 to 5 years.

%
%
%
%
%
%


\bibliographystyle{IEEEtran}
\bibliography{reconstruction}

\end{document}